\documentclass{article}

\usepackage{microtype}
\usepackage{graphicx}
\usepackage{subcaption}
\usepackage{booktabs} % for professional tables

\usepackage{hyperref}

\usepackage[preprint]{icml2026}

\usepackage{hyperref}
\usepackage{url}
\usepackage{amsmath}
\usepackage{amssymb}
\usepackage{mathtools}
\usepackage{amsthm}
\usepackage{algorithm}
\usepackage{algorithmic}
\usepackage{bm}
\usepackage{graphicx}
\usepackage{fancybox}
\usepackage{float}
\usepackage{caption}
\usepackage{multicol}
\usepackage{relsize}
\usepackage{enumitem}
\usepackage{wrapfig}
\usepackage{booktabs}
\usepackage{multirow}
\usepackage{colortbl}
\usepackage{subcaption}

\usepackage[capitalize,noabbrev]{cleveref}

\theoremstyle{plain}
\newtheorem{theorem}{Theorem}[section]
\newtheorem{proposition}[theorem]{Proposition}
\newtheorem{lemma}[theorem]{Lemma}
\newtheorem{corollary}[theorem]{Corollary}
\theoremstyle{definition}

\theoremstyle{remark}

\usepackage[textsize=tiny]{todonotes}

\icmltitlerunning{}

\begin{document}

\twocolumn[
  \icmltitle{Improving Fine-Grained Control via Aggregation of Multiple Diffusion Models}



  \icmlsetsymbol{equal}{*}

  \begin{icmlauthorlist}
\icmlauthor{Conghan Yue}{sysu,ten}
\icmlauthor{Zhengwei Peng}{sysu}
\icmlauthor{Shiyan Du}{sysu}
\icmlauthor{Zhi Ji}{ut,ten}
\icmlauthor{Chuangjian Cai}{ten}
\icmlauthor{Le Wan}{ten}
\icmlauthor{Dongyu Zhang}{sysu}
\end{icmlauthorlist}

\icmlaffiliation{sysu}{Sun Yat-Sen University}
\icmlaffiliation{ten}{Game AI Center, Tencent}
\icmlaffiliation{ut}{University of Toronto}
\icmlcorrespondingauthor{Dongyu Zhang}{zhangdy27@mail.sysu.edu.cn}

  \icmlkeywords{Machine Learning, ICML}
  \vskip 0.3in
]



\printAffiliationsAndNotice{}  % no special notice (required even if empty)

\begin{abstract}
While many diffusion models perform well when controlling particular aspects such as style, character, and interaction, they struggle with fine-grained control due to dataset limitations and intricate model architecture design. This paper introduces a novel training-free algorithm for fine-grained generation, called Aggregation of Multiple Diffusion Models (AMDM). The algorithm integrates features in the latent data space from multiple diffusion models within the same ecosystem into a specified model, thereby activating particular features and enabling fine-grained control. Experimental results demonstrate that AMDM significantly improves fine-grained control without training, validating its effectiveness. Additionally, it reveals that diffusion models initially focus on features such as position, attributes, and style, with later stages improving generation quality and consistency. AMDM offers a new perspective for tackling the challenges of fine-grained conditional generation in diffusion models. Specifically, it allows us to fully utilize existing or develop new conditional diffusion models that control specific aspects, and then aggregate them using the AMDM algorithm. This eliminates the need for constructing complex datasets, designing intricate model architectures, and incurring high training costs. 
% Code is available at: \url{https://github.com/Hammour-steak/AMDM}.
\end{abstract}

\section{Introduction}
% Diffusion models \citep{sohl2015deep, ho2020denoising, songdenoising, song2020score, karras2022elucidating} are designed to establish a relationship between data and noise, utilizing neural networks to learn the reverse process \citep{anderson1982reverse}. This enables the generation of data from random noise, showcasing exceptional performance in generative tasks. In practical applications like Text-to-Image (T2I) \citep{nichol2022glide, chenre, lee2024holistic, xu2024imagereward} and Image-to-Image (I2I) \citep{zhang2023inversion, mou2024diffeditor} generation, conditional diffusion models \citep{rombach2022high, chungdiffusion, esser2024scaling} are widely used. 

Diffusion models \cite{sohl2015deep, ho2020denoising, songdenoising, song2020score, karras2022elucidating} have achieved excellent performance in generative tasks. In particular, conditional diffusion models \cite{rombach2022high, chungdiffusion, esser2024scaling} not only deliver advanced results in practical applications such as Text-to-Image (T2I) \cite{nichol2022glide, chenre, lee2024holistic, xu2024imagereward} and Image-to-Image (I2I) generation \cite{zhang2023inversion, mou2024diffeditor}, but also offer highly flexible conditional control mechanisms.
% 扩散模型在生成任务上取得了优秀的表现，其条件扩散模型在实际应用如文生图和图生图中不但取得了最先进的效果，也拥有非常灵活的条件控制方式，进而使得扩散模型的研究成为主流。
% 扩散模型旨在建立数据与噪声之间的扩散关系，利用神经网络学习逆向过程，进而能从随机噪声中生成数据，在生成任务上取得了优秀的表现。在实际应用中，例如文生图和图生图，则会使用条件扩散模型，不但取得了最先进的效果，也拥有非常灵活的条件控制方式，进而使得扩散模型的研究成为主流。

Recent research on conditional diffusion models has focused on achieving fine-grained control, including object attributes \citep{wu2023uncovering, wang2024compositional}, interactions \citep{hoe2024interactdiffusion, jia2024customizing}, layouts \citep{zheng2023layoutdiffusion, chai2023layoutdm, chen2024textdiffuser}, and style \citep{wang2023stylediffusion, huang2024diffstyler, qi2024deadiff}. However, maintaining consistency across diverse nuanced control remains a significant challenge.
Generating multiple objects with overlapping bounding boxes can lead to attribute leakage, where the description of one object inappropriately influences others, causing inconsistencies between objects and the background. Fine-grained interaction details may be illogical, and style integration may compromise object attributes. 

Existing generation approaches only partially address these issues due to the inherent complexity and diversity of fine-grained control, coupled with limitations in datasets and model architectures. Some works \cite{li2023gligen, zhou2024migc, wang2024instancediffusion} may perform well in preventing attribute leakage among multiple instances during layout generation but perform poorly in managing object interactions, while others \cite{ye2023ip, huang2024diffstyler} may excel in style transfer but exhibit limited control over layout. 

Interestingly, most of these methods are based on Stable Diffusion \cite{rombach2022high}, which is theoretically grounded in DDPM \cite{ho2020denoising} and classifier-free guidance \cite{dhariwal2021diffusion} for conditional control. In this context, a distinct line of research has recently emerged that leverages the shared diffusion equation across different models to combine multiple diffusion models at inference time, thereby integrating their respective information. The predominant strategy is to apply linear weighting in the score space \cite{ garipov2023compositional, bradley2025mechanisms, thornton2025composition, skreta2025feynman,he2025rne,skretasuperposition}. However, these methods were originally designed to synthesize new data distributions under mutually independent (orthogonal) conditioning variables \cite{bradley2025mechanisms}. Although fine-grained generation can be pursued by enforcing identical conditions, this practice violates the orthogonality assumption embedded in the original design, leading to suboptimal generation quality. To mitigate the distributional shift caused by linear weighting, several works have proposed more sophisticated weighting schemes, which, however, introduce additional inference overhead \cite{skretasuperposition,he2025rne}. Another line of methods performs linear weighting in the parameter space of the networks \cite{biggs2024diffusion, ohdawin, wangensembling}, but this requires the combined models to share exactly the same architecture, thereby substantially limiting their applicability.
% 因此，目前还出现了一类工作，利用这些模型共享同一扩散方程的性质，将他们进行组合，从而获得不同模型的信息。一些组合方法的主体方案为对得分空间的线性加权，但他们的设计初衷都是给定不同条件从而生成新的分布，虽然可以通过设置相同条件实现细粒度生成，但对于细粒度任务来说，这违反设计时条件正交的假设，从而使得生成效果不理想。此外，为了对抗线性加权所带来的偏离，许多组合方法采用了更复杂的加权方式，这也引入了一定的推理成本。还有一类组合方法对网络参数空间进行线性加权，但这要求组合模型之间具有完全相同的网络架构，限制了应用范围。

This paper proposes a training-free AMDM algorithm that operates directly in the latent data space guided by a geometric perspective. The algorithm is both conceptually simple and practically effective, relying solely on closed-form computations with negligible computational overhead. As a result, it offers an efficient and viable solution for fine-grained control tasks that are highly coupled and non-orthogonal in nature.
In summary, this work makes the following contributions: \textbf{(1)} We propose a novel diffusion model aggregation algorithm, AMDM, that can aggregate intermediate variables in latent data space from multiple conditional diffusion models within the same ecosystem enabling fine-grained generation; \textbf{(2)} We conduct extensive experiments, with both qualitative and quantitative results demonstrating significant improvements, particularly in regions where previous models exhibited limited controllability; \textbf{(3)} We reveal that diffusion models initially focus on generating coarse-grained features such as position, attributes, and style, while later stages emphasize quality and consistency.

\section{Preliminaries: Stable Diffusion}
Stable Diffusion \cite{rombach2022high} is a class of diffusion models defined in the latent space. The original data $\mathbf{x}_0$ is mapped through the encoder (VAE \cite{kingma2013auto}) to obtain the latent variable $\mathbf{z}_0 = E(\mathbf{x}_0)$, which evolves according to the DDPM \cite{ho2020denoising} diffusion paradigm:
\begin{equation}\label{forward_t}
    p(\mathbf{z}_t | \mathbf{z}_{t-1}) = \mathcal{N}(\sqrt{\alpha_t}\mathbf{z}_{t-1}, (1-\alpha_t)\bm I),
\end{equation}
where \(\mathbf{z}_t\) represents the noisy latent data at timestep \(t\in [0,T]\), \(\alpha_t\) is the coefficient drift schedule generally satisfying $\lim_{t\rightarrow T}\alpha_t=0$. From (\ref{forward_t}), the forward marginal distribution is:
\begin{equation}\label{forward}
    p(\mathbf{z}_t | \mathbf{z}_{0}) = \mathcal{N}(\sqrt{\bar\alpha_t}\mathbf{z}_0, (1-\bar\alpha_t)\bm I),
\end{equation}
where $\bar\alpha_t=\prod_{i=1}^{t}\alpha_i$. Assuming the denoising neural network is \(\bm\epsilon_{\theta}\) and the condition is $y$, the loss function is the variational lower bound of its likelihood, i.e., the KL divergence of the joint probability:
\begin{equation}
\begin{aligned}
\mathcal{L}&=KL(q(\mathbf{z}_{0:T})\|p_{\theta}(\mathbf{z}_{0:T}))\\
&\propto \mathbb{E}_{t,\mathbf{z}_t,\epsilon_t}\left[\|\bm\epsilon_t - \bm\epsilon_{\theta}(\mathbf{z}_{t},t,y)\|^2 \right],
\end{aligned}
\end{equation}
where $\bm\epsilon_t \sim \mathcal{N}(\mathbf{0},\bm I)$. More generally,  the reverse sampling process is as follows:
\begin{equation}\label{cfg}
\begin{aligned}
 p_{\theta}(\mathbf{z}_{t-1}\mid \mathbf{z}_{t}) =&\mathcal N(\bm\mu_{\theta}(\mathbf{z}_t, t, y),\sigma_{t}^2 \bm I)\\
 &\mathcal{N}\left(\sqrt\frac{\bar\alpha_{t-1}}{\bar\alpha_t}\mathbf{z}_{t}+\left(\sqrt{1-\bar\alpha_{t-1}-\sigma_{t}^2}\right. \right.  \\
 & \left. \left. -\sqrt\frac{\bar\alpha_{t-1}(1-\bar\alpha_t)}{\bar\alpha_t}\right)\bm\epsilon_{\theta}(\mathbf{z}_{t},t,y),\sigma_{t}^2 \bm I\right),
\end{aligned}
\end{equation}
where $\sigma_t$ is a free variable.

\section{Method}
% In this section, we first analyze the challenges and limitations of current fine-grained control research，随后分析聚合运算的存在性，然后根据细粒度任务条件耦合和扩散模型生态特点提出假设，发现推理过程中几何性质，从而指导出AMDM的最终设计。
In this section, we first analyze the challenges and limitations faced by current research on fine-grained control. We then discuss the existence and applicability of aggregation operations. 
Subsequently, we present Proposition \ref{p_normupper}, which reveals the geometric properties of the inference process, 
and Proposition \ref{p1}, which guides the final design of AMDM.

\subsection{Analysis}\label{analysis}
\begin{figure}[!t]
  \centering
  \includegraphics[width=1.0\columnwidth]{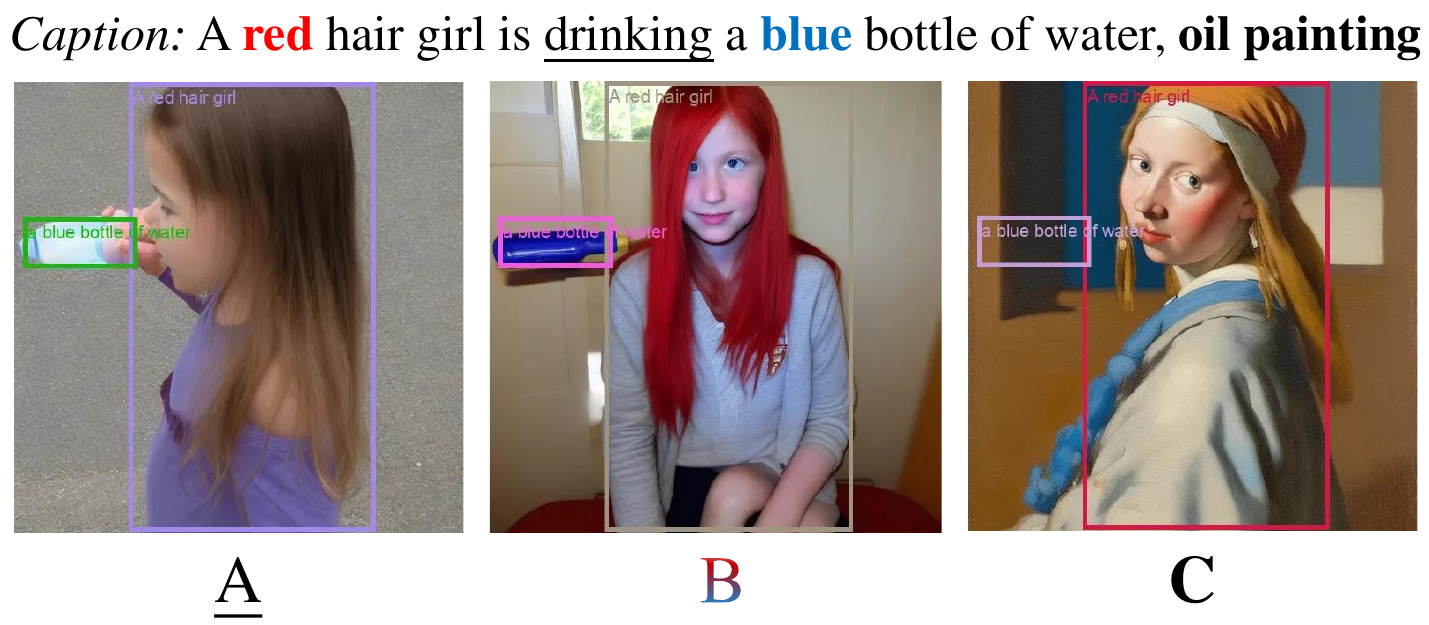}
  \caption{Examples of fine-grained conditional control of the same caption by different models.}
  \label{anal}
  % \vskip -0.15 in
\end{figure}

Current fine-grained conditional control models offer limited controllability and face numerous issues. For example, given the caption "A red hair girl is drinking from a blue bottle of water, oil painting" and corresponding bounding boxes for positioning control, different models are likely to show varying performance, as illustrated in Figure \ref{anal}. Model A, which receives additional inputs for position information and actions, excels at accurate positioning and interactions, achieving high-quality results. However, it struggles with attribute control and maintaining the oil painting style. Conversely, Model B incorporates extra input for position and attribute information, managing both but not accurately capturing interactions and stylistic elements. Model C references the style of an image, enabling precise management of style characteristics but lacking adequate control over location and attribute details. The fundamental reason for these issues lies in the complexity and flexibility of fine-grained control tasks, which makes it challenging for limited datasets and specific model architectures to account for all the intricate features. It is noteworthy that these models share a common foundation, as they are all based on the same diffusion process. Recognizing this shared basis, our objective is to develop an aggregation algorithm that leverages these commonalities to integrate the distinctive characteristics of multiple models, achieving fine-grained conditional control in a more direct and efficient manner.

\paragraph{Do aggregation operations exist in different diffusion models?}
For a latent diffusion model $p_{\theta}$, the true latent data manifold along the diffusion process is $\mathcal{M}_t(\xi_{t}) = \{ \mathbf{z}_t \in \mathbb{R}^n \mid p(\mathbf{z}_t) \geq \xi_{t} \}$, where an appropriate choice of $\xi_{t}$ 
ensures that $\mathcal{M}_t$ is a set of meaningful data. 
We define the generation domain at time $t$ under condition $y$ as $D^{\theta}_{t,y}(\tau_{t,y}^{\theta}) = \{ \mathbf{z}_t \in \mathbb{R}^n \mid p_\theta(\mathbf{z}_t \mid y) \geq \tau_{t,y}^{\theta} \}$.
Since conditional samples are a selection of the unconditional population, there exists a largest threshold $\tau_{t,y}^{\theta}$ such that 
$D^{\theta}_{t,y}(\tau_{t,y}^{\theta}) \subset \mathcal{M}_t(\xi_{t})$.
This implies that, given $\xi_{t}$, the $D^{\theta}_{t,y}(\tau_{t,y}^{\theta})$ is uniquely determined, and we thus omit the explicit dependence on the threshold for simplicity.
Mathematically, the most fundamental requirement for performing aggregation operations is that all elements must reside in a shared space. However, in the case of aggregation between two models, it is evident that data obtained through conditional sampling from \(p_{\theta_1}\) should lie within \(D^{\theta_1}_{0,y_1}\), whereas data obtained from \(p_{\theta_2}\) should lie within \(D^{\theta_2}_{0,y_2}\). According to the definition, $D^{\theta_1}_{0,y_1} \subset \mathcal M_0,  D^{\theta_2}_{0,y_2} \subset \mathcal M_0$, which implies that the data from different models can be embedded into a shared representation space \(\mathcal M_0\). This observation directly leads to the \textbf{first requirement}: \(\mathcal M_0\) must coincide across models, implying that the latent space encoders are consistent, and the aggregation operations at \(t=0\) exist.
For \(t > 0\), this naturally leads to the \textbf{second requirement}: \(\mathcal M_t\) must coincide across models. In the context of diffusion models, a sufficient condition for this requirement is that the underlying SDEs are identical. Consequently, we obtain $D^{\theta_1}_{t,y_1} , D^{\theta_2}_{t,y_2} \subset \mathcal M_t$, which ensures the existence of aggregation operations for all time steps $t$. 

\paragraph{Which models can achieve fine-grained generation through aggregation operations?}
In practice, we expect the result of the operation to lie within the smaller space $D^{\theta_1}_{t,y_1} \cap D^{\theta_2}_{t,y_2}$, since only in this space is it possible to faithfully incorporate information from both models and achieve fine-grained generation. In order to achieve aggregation operation at time $t$, it is further required that the domains $D^{\theta_1}_{t,y_1}$ and $D^{\theta_2}_{t,y_2}$ of different models possess a non-empty intersection. 
First, we observe that application models developed on top of SD, in addition to satisfying the aforementioned two theoretical requirements, share several common characteristics in their denoising networks:  
\textbf{1. Additive Architectures:} Typically freeze most or all of the weights of the original base model and then integrate a new, lightweight, and trainable network module in parallel.  
\textbf{2. Modified Architectures:} Replace or alter certain layers within the U-Net while retaining its overall structure.  
\textbf{3. Preserved Architectures:} Fine-tuning is performed with additional training samples from specific domains while keeping the overall architecture intact.  
This indicates that although the three approaches differ in their denoising network architectures, they preserve most of the original functionality and only enhance specific control aspects. Therefore, we consider these models that satisfy the above properties of the denoising network and the two requirements as belonging to the same \textbf{diffusion model ecosystem}, indicating that when given the same input, different models exhibit similar tendencies. Accordingly, we can believe that there exists \textit{functional proximity}: $\|\bm\mu_{\theta_1}(\mathbf{z}_{t},t,y)-\bm\mu_{\theta_2}(\mathbf{z}_{t},t,y)\| \le L_{t}$.
When it comes to fine-grained generation, the conditions $y_1$ and $y_2$ describe the same task $\mathcal{T}$, which means that they are strongly coupled and semantically consistent, as their global prompts are at least aligned. This implies that we can assume \textit{conditional proximity}: $\|y_1-y_2\|\le L_{\mathcal T}$. Based on this, it is natural to infer that for two models \(p_{\theta_1}\) and \(p_{\theta_2}\) within the same diffusion model ecosystem, given two distinct condition descriptions \(y_1\) and \(y_2\) of a fine-grained task, they are inclined to generate the same samples and pursue a unified objective. Specifically, we have $D^{\theta_1}_{0,y_1} \cap D^{\theta_2}_{0,y_2} \neq \varnothing$.
Moreover, for the noised state at any time \(t\), we likewise have $D^{\theta_1}_{t,y_1} \cap D^{\theta_2}_{t,y_2} \neq \varnothing$. Therefore, models within the same diffusion ecosystem are capable of achieving fine-grained generation.
% 因此，处于同一扩散生态下的模型能够实现细粒度生成。

The above analysis suggests that an aggregation operation exists only when the models are within the same diffusion model ecosystem. This finding also provides theoretical support for the validity of other compositional methods and thus establishes a solid foundation for the design of the subsequent algorithm.

% Moreover, belonging to the same diffusion ecosystem is an essential condition underlying all compositional methods, which in turn provide theoretical support for their validity. Beyond the case of two models, it can also naturally extend to multiple models, thereby providing a solid foundation for the design of the subsequent algorithm.

% 首先我们注意到，这些基于SD开发的应用模型除了满足了以上两个理论要求，他们的去噪网络还都具有以下特征：1.Additive Architectures，这类方法通常会冻结原始基础模型的全部或大部分权重，然后并联地加入一个新的、小型的、可训练的网络模块。2.Modified Architectures，一些研究可能会替换或修改U-Net内部的某些层，但保留整体结构。3.保持Architectures，即用一些额外的域的训练样本进行微调。这说明，即使这三种方式的去噪网络架构存在差异，但它们保持大部分原有功能不变，只在特定控制方面进行提升，因此我们称这些模型处于同一扩散模型生态，具有功能临近性。此外，对于细粒度任务来说，不同的模型目标是一致的，所以他们接受不同的条件\(y_1\) and \(y_2\)是高度耦合的，互相兼容的，这表明不同条件的语义信息是高度一致的，具有条件临近性。基于以上，很自然的对于同一扩散模型生态的不同模型$p_{\theta_1}$ and $p_{\theta_2}$，对于一个细粒度任务的两个不同描述条件\(y_1\) and \(y_2\)，他们会倾向于生成同样的样本，拥有一致的目标。这意味者$D^{\theta_1}_{0,y_1} \cap D^{\theta_2}_{0,y_2} \neq \varnothing$, and for the noised state at any time \(t\), we similarly have $D^{\theta_1}_{t,y_1} \cap D^{\theta_2}_{t,y_2} \neq \varnothing$.

% Fortunately, in fine-grained generation tasks, the intersection naturally exists for \textbf{almost all}. 
% This is because both models share the same generation target in fine-grained tasks, meaning the conditions \(y_1\) and \(y_2\) are compatible, and the data generated by both models 
% tend to be similar, i.e., $D^{\theta_1}_{0,y_1} \cap D^{\theta_2}_{0,y_2} \neq \varnothing$, and for the noised state at any time \(t\), we similarly have $D^{\theta_1}_{t,y_1} \cap D^{\theta_2}_{t,y_2} \neq \varnothing$.

\subsection{Algorithm: AMDM}

% \begin{figure}[t]
% \begin{center}
% \centerline{\includegraphics[width=0.6\columnwidth]{images/p1.pdf}}
% \caption{Geometry of deviation optimization between two models. The green points represent the original sampled points, the red points indicate the results of spherical aggregation, and the gold points denote the final results after deviation optimization.}
% \vskip -0.3in
% \label{manop}
% \end{center}
% \end{figure}

\paragraph{Spherical Aggregation.} To identify a feasible aggregation algorithm, we need to examine the properties of 
\(\mathcal{M}_t\). At the initial stage \(t = T\), since \(p(\mathbf{z}_T)\) follows a standard Gaussian prior, it follows from the spherical concentration property of high-dimensional Gaussian distributions that \(\mathcal{M}_T\) is concentrated on an \((n-1)\)-dimensional manifold, namely an \(n\)-dimensional hypersphere. The proof is provided in Appendix \ref{concentration}. For \(t < T\), \citet{chung2022improving} demonstrated that \(\mathcal{M}_t\) is concentrated on an \((n-1)\)-dimensional manifold, which approximates an \(n\)-dimensional hypersphere as \(t\) becomes large. Furthermore, we have the following proposition:
\begin{proposition}\label{p_normupper}
Let \( \mathbf z'_t \) denote the aggregated variable at time step \( t \). For the sampling step from \( t \) to \( t-1 \), two diffusion models \(p_{\theta_1}\) and \(p_{\theta_2}\) sample \( \mathbf{z}^{\theta_1}_{t-1} \) and \( \mathbf{z}^{\theta_2}_{t-1} \) respectively from (\ref{cfg}). Then, with probability at least $1-\gamma$, the absolute difference in norms and the angles are bounded by:
\begin{equation}\label{e_normupper}
\Big|\|\mathbf z^{\theta_1}_{t-1}\|-\|\mathbf z^{\theta_2}_{t-1}\|\Big|\leq L_t+L_{\theta_2}L_\tau+2\sigma_t \sqrt{2 \ln \frac{4}{\gamma}},
\end{equation}
\begin{equation}\label{cosin}
\begin{gathered}
\cos \varphi \geq 1-\frac{\delta^2}{\|\mathbf z^{\theta_1}_{t-1}\|^2+\|\mathbf z^{\theta_2}_{t-1}\|^2},\\
\delta \leq L_t+ L_{\theta_2}L_\tau +  \sigma_t \sqrt{2 \left( n + 2\sqrt{n \ln \frac{1}{\gamma}} + 2\ln \frac{1}{\gamma} \right)}.
\end{gathered}
\end{equation}
where $\varphi$ is the angle between $\mathbf z^{\theta_1}_{t-1}$ and $\mathbf z^{\theta_2}_{t-1}$, and $L_{\theta_2}$ is the Lipschitz constant.
\end{proposition}
% 命题 \ref{p_normupper} 表明，当同一扩散生态下的不同模型在细粒度任务中对同一点进行采样时，所生成的新变量具有近似相等的模长，且彼此间的夹角极小。综上所述，我们可以断言：当 $t<T$ 时，采样变量分布于一个局部球面流形上。
The proof is provided in Appendix \ref{proof_normupper}.
% and the numerical experiments are presented in Appendix \ref{assumpnorm}.
Proposition \ref{p_normupper} indicates that when different models from the same diffusion ecosystem perform sampling on the same point in fine-grained tasks, the resulting new variables have approximately equal norms and small angles between them. 
Taken together, we may assert that when $t<T$, the sampled variables lie on a local spherical manifold.
Motivated by the geometric properties of the global sphere at time $T$ and the local spheres at $t<T$, we propose spherical interpolation for aggregation, which maximizes the retention of aggregated data on the original manifold while minimizing deviations:
\begin{equation}\label{agg_x}
\begin{aligned}
\mathbf z'_{t-1} &=Slerp(\mathbf z^{\theta_1}_{t-1}, \mathbf z^{\theta_2}_{t-1},w)\\
&=\frac{\sin((1 - w) \varphi)}{\sin(\varphi)} \mathbf z^{\theta_1}_{t-1} + \frac{\sin(w \varphi)}{\sin(\varphi)} \mathbf z^{\theta_2}_{t-1},
\end{aligned}
\end{equation}
where \(w\in [0,1]\) is the weighting factor that balances the contribution of each model. Spherical aggregation integrates the conditional control information of \(p_{\theta_2}\) and \(p_{\theta_1}\), while keeping the new variables stable near the manifold.

\begin{figure}[t]
\begin{center}
\centerline{\includegraphics[width=1.0\columnwidth]{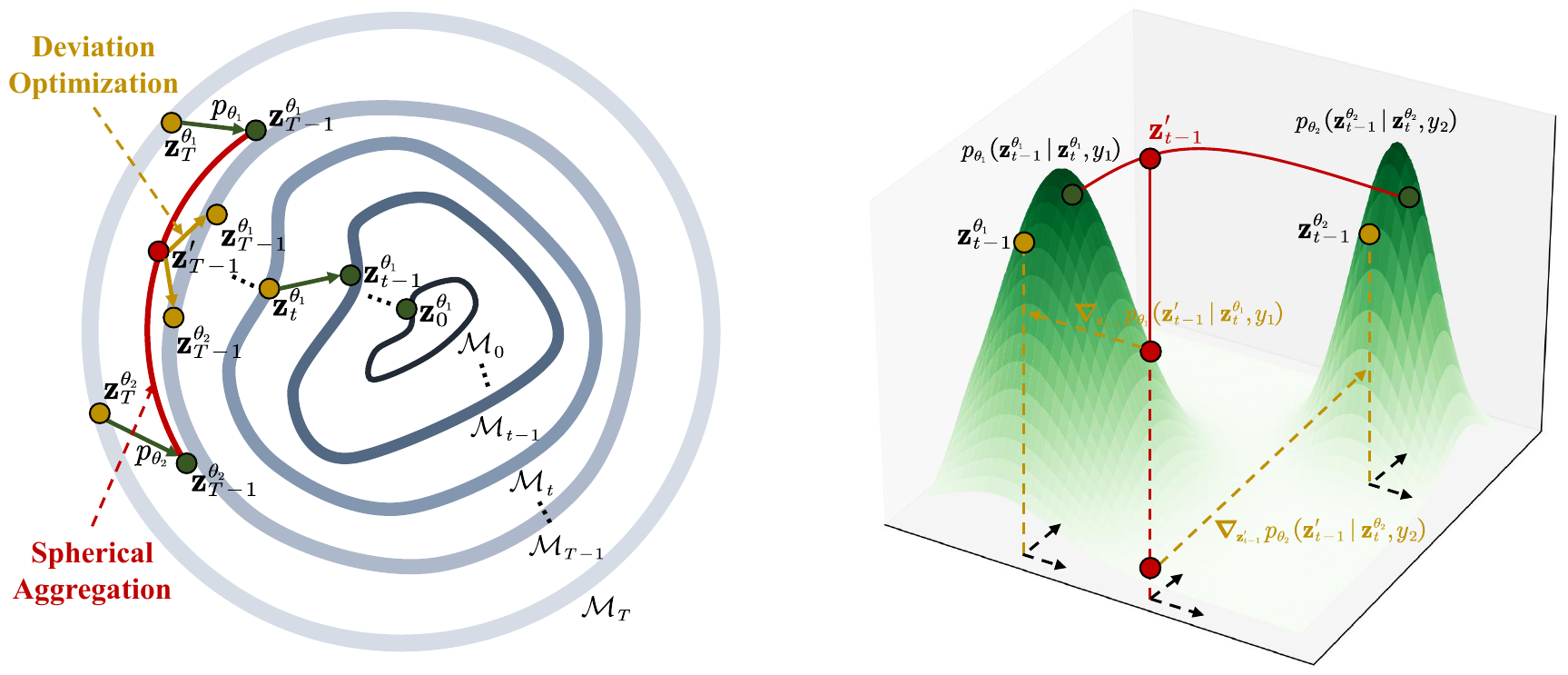}}
\caption{Geometry of AMDM (Left) and Deviation Optimization (Right). The algorithm employs spherical aggregation and deviation optimization to incorporate conditional information during the initial steps. Subsequently, direct sampling is applied to expedite the process and generate high-quality images.}
\label{gamdm}
\end{center}
\vskip -0.3in
\end{figure}

\paragraph{Deviation Optimization.} 
Ideally, we expect \(\mathbf z'_{t-1}\in D^{\theta_1}_{t-1,y_1} \bigcap D^{\theta_2}_{t-1,y_2}\), but deviations are likely to occur in practice. Considering the step from $t$ to $t-1$, since $p_{\theta_1}(\mathbf z^{\theta_1}_{t-1}\mid \mathbf z^{\theta_1}_t, y_1)$ 
is high-dimensional Gaussian, its samples concentrate in a thin spherical shell around the mean. Motivated by this, we perform a radial adjustment of 
$\mathbf z'_{t-1}$ toward that shell to obtain 
$\mathbf{\tilde z}_{t-1}\in D^{\theta_1}_{t-1,y_1}$. 
This adjustment alone does not guarantee 
$\mathbf{\tilde z}_{t-1}\in D^{\theta_2}_{t-1,y_2}$. Nevertheless, we show that $\mathbf{\tilde z}_{t-1}$ belongs to $D^{\theta_2}_{t-1,y_2}$ with high probability, which leads to the following proposition:
\begin{proposition}\label{p1}
For the diffusion model \( p_{\theta_1} \) defined by (\ref{cfg}) and any new intermediate variable \(\mathbf{z}'_{t-1}\) from (\ref{agg_x}), let:
    \begin{equation}\label{op}
        \mathbf{\tilde z}_{t-1} = \mathbf{z}'_{t-1} - \eta^{\theta_1}_{t-1} \frac{\mathbf{z}'_{t-1} - \bm\mu_{\theta_1}(\mathbf{z}^{\theta_1}_t, t, y_1)}{\|\mathbf{z}'_{t-1} - \bm\mu_{\theta_1}(\mathbf{z}^{\theta_1}_t, t, y_1)\|},
    \end{equation}
where $\eta^{\theta_1}_{t-1}$ is a small optimization step size. There exists $\eta^{\theta_1}_{t-1}$ such that $\mathbf{\tilde z}_{t-1}\in D^{\theta_1}_{t-1,y_1}$. Moreover, an approximate lower bound on the probability that $\mathbf{\tilde z}_{t-1}\in D^{\theta_2}_{t-1,y_2}$ is given by:
\begin{equation}\label{lowbound}
\small
P\left(\mathbf{\tilde z}_{t-1}\in D^{\theta_2}_{t-1,y_2}\right)\geq 1 - 2 \exp\left(
   -\frac{n\left(\epsilon_{t-1}^{\theta_2} - \tfrac{d}{\sigma_t\sqrt{n}}\right)^{2}}
          {1 + 2\left(\epsilon_{t-1}^{\theta_2} - \tfrac{d}{\sigma_t\sqrt{n}}\right)}
\right),
\end{equation}
where $d = \phi_w(\varphi) \delta + \eta^{\theta_1}_{t-1}$ and $\phi_w(\varphi)
= \sin\!\bigl((1-w)\varphi/2\bigr)/\sin(\varphi/2)$.
\end{proposition}

The proof is given in Appendix \ref{proof_p1}; a detailed analysis of $d$ is presented in Appendix \ref{maxdistance}; and the geometry of deviation optimization is illustrated in Figure \ref{gamdm} (Right). Proposition \ref{p1} shows that $\mathbf{\tilde z}_{t-1}$ obtained after spherical aggregation can be corrected into $D^{\theta_1}_{t-1,y_1}$ through deviation optimization, and with high probability can also be corrected into $D^{\theta_2}_{t-1,y_2}$. 

\begin{corollary}
The Deviation Optimization \eqref{op} can be naturally extended to the deterministic limit $\sigma_t \to 0$, which can be interpreted as aligning the variable onto the manifold induced by the deterministic flow.
% While Proposition 3.2 establishes a probabilistic bound in the stochastic regime where $\sigma_t > 0$, its geometric intuition extends naturally to the deterministic limit $\sigma_t \to 0$. In the stochastic case, Deviation Optimization acts as steering the state towards the high-density region (manifold), effectively pulling $z_{t-1}$ back to the vicinity of this region in the conditional distribution $p(z_{t-1}|z_t)$. As $\sigma_t \to 0$, this conditional distribution degenerates from the transition kernel of the noisy SDE into a Dirac measure centered on the characteristic trajectory induced by the corresponding ODE vector field, i.e.,$p(z_{t-1}|z_t) \Rightarrow \delta(z_{t-1} - \Phi_{t\to t-1}(z_t))$, where $\Phi_{t\to t-1}$ denotes the ODE flow map. Consequently, the "concentration" property in SDEs seamlessly transitions into a "manifold projection" property in the ODE limit: in this context, Deviation Optimization is equivalent to explicitly aligning the numerical state with the deterministic flow to correct drifts and deviations arising from discrete integration.
\end{corollary}

\textit{Proof sketch.} As $\sigma_t \to 0$, this conditional distribution $p(z_{t-1}|z_t)$ degenerates from the transition kernel of the noisy SDE into a Dirac distribution centered on the characteristic trajectory induced by the corresponding ODE vector field, i.e.,$p(z_{t-1}|z_t) \Rightarrow \delta(z_{t-1} - \Phi_{t\to t-1}(z_t))$, where $\Phi_{t\to t-1}$ denotes the ODE flow map. Consequently, the concentration property in SDEs seamlessly transitions into a manifold projection property in the ODE. 

In this context, deviation optimization pulls the current intermediate state back onto the manifold, thereby mitigating path deviations induced by the aggregation step. It ensures that the aggregated variable effectively synthesizes information from both models while preserving high sampling fidelity.

Motivated by Propositions \ref{p_normupper} and \ref{p1}, we combine \eqref{agg_x} and \eqref{op} to derive an aggregation algorithm for two diffusion models, comprising two key components: spherical aggregation and deviation optimization. Spherical aggregation aggregates the conditional control information from different models and ensures that the new intermediate variables remain stable near the manifold, while deviation optimization ensures more precise retention on the corresponding data manifold, enhancing sample quality. The algorithm iteratively performs spherical aggregation and deviation optimization for each model during the first \(s\) steps, followed by direct sampling from \(p_{\theta_1}\). This algorithm can be readily extended to multiple models, resulting in the final Aggregation of Multiple Diffusion Models (AMDM) algorithm, as shown in Algorithm \ref{alg2} and Figure \ref{gamdm} (Left), where the spherical aggregation of multiple elements is defined through successive pairwise aggregation in a fixed order $(1,…,N)$, i.e., a left-fold recursion.
\begin{algorithm}[t]
\caption{AMDM}
\begin{algorithmic}\label{alg2}
\STATE {\bfseries Input:} models from the same diffusion ecosystem $p_{\theta_1}, p_{\theta_2}, ..., p_{\theta_N}$, conditions $y_1,y_2,...,y_N$, aggregation step $s$, weighting factors $w_1,w_2,...,w_{N-1}$ and optimization steps $\eta^{\theta_1}_{t}, \eta^{\theta_2}_{t},...,\eta^{\theta_N}_{t}$
\STATE $\mathbf{z}_T^{\theta_1},\mathbf{z}_T^{\theta_2},...,\mathbf{z}_T^{\theta_N} \sim N(\bm 0,\bm I)$
    \FOR{$t=T,...,1$}
        \STATE $\mathbf z^{\theta_1}_{t-1}\sim p_{\theta_1}(\mathbf{z}_{t-1}^{\theta_1} | \mathbf{z}_t^{\theta_1}, y_1)$
        \IF{$t>T-s$}
            \STATE $\mathbf z^{\theta_i}_{t-1}\sim p_{\theta_i}(\mathbf{z}_{t-1}^{\theta_i} | \mathbf{z}_t^{\theta_i}, y_i)$, $i\in [2,N]$
            \STATE $\mathbf z'_{t-1} = Slerp(\mathbf z^{\theta_1}_{t-1},..., \mathbf z^{\theta_N}_{t-1}, w_1,..., w_{N-1})$
            \STATE $\mathbf{z}_{t-1}^{\theta_i} = \mathbf{z}'_{t-1} - \eta_{t-1}^{\theta_i} \frac{\mathbf{z}'_{t-1} - \mu_{\theta_i}(\mathbf{z}_t^{\theta_i}, t, y_i)}{\|\mathbf{z}'_{t-1} - \mu_{\theta_i}(\mathbf{z}_t^{\theta_i}, t, y_i)\|}$, $i\in [1,N]$
        \ENDIF
    \ENDFOR
\STATE $\mathbf x_0^{\theta_1} = \text{Decoder}(\mathbf z_0^{\theta_1})$
\STATE {\bfseries Output:} $\mathbf x^{\theta_1}_{0}$
\end{algorithmic}
\end{algorithm}

We perform aggregation only in the first $s$ steps, and subsequently apply a single model $p_{\theta_1}$ for inference. This procedure can be regarded as aggregating the features of other models into $p_{\theta_1}$, striking a balance between efficiency and effectiveness. Moreover, note that since $\mu_{\theta_i}(\mathbf{z}_t^{\theta_i}, t, y_i)$ can reuse $\epsilon_{\theta_i}(\mathbf{z}_t^{\theta_i}, t, y_i)$ from the previous sampling step, the deviation optimization introduces only a single mathematical operation with negligible computational overhead, further reducing inference time.

As for the selection of $p_{\theta_1}$, in order to ensure high-quality generation in the later stages, it is natural to choose the model with stronger generative capability as $p_{\theta_1}$. This choice can be evaluated using a variety of criteria, such as quantitative metrics, model size, application universality, or direct experimental validation.

It is worth noting that although Proposition \ref{p_normupper} shows that samples lie on a local spherical manifold under the same aggregation point, motivating our spherical aggregation, this geometric property persists after the deviation optimization in Proposition \ref{p1}, further demonstrating the correctness of AMDM algorithm. The corresponding theoretical analysis and proof are provided in Proposition \ref{amdm_prop} and Appendix \ref{geo_amdm}, while numerical validation can be found in Appendix \ref{assumpnorm}.
\section{Experiments}\label{ex}
In this section, we aggregate several classic conditional diffusion models, all of which belong to the same Stable Diffusion ecosystem, followed by a series of ablation experiments to demonstrate its optimality.
The core idea of our experiment is to demonstrate the effectiveness of the AMDM algorithm: Given a set of models with varying control capabilities, we only need to focus on whether the target model, 
after aggregating features from other models, can significantly enhance its performance in areas of weak control, while simultaneously introducing only minor performance trade-offs 
in its areas of strength.
All experiments were conducted using a single RTX 3090 GPU and experimental details and additional experiments are provided in Appendix \ref{ed}.

\subsection{Aggregation Experiments}
\begin{table}
    \caption{Quantitative results on the HOI Detection Score, FID and CLIP Score across different models.}
    \centering
    \resizebox{1.0\columnwidth}{!}{
    \begin{tabular}{c  c c  c c  c  c }
        \toprule
        \multirow{2}{*}{\textbf{Method}} & \multicolumn{2}{c}{\textbf{Default $\uparrow$}} & \multicolumn{2}{c}{\textbf{Known Object $\uparrow$}} & \multirow{2}{*}{\textbf{FID $\downarrow$}} & \multirow{2}{*}{\textbf{CLIP Score $\uparrow$}} \\ 
        \cmidrule(lr){2-3} \cmidrule(lr){4-5}
        & Full & Rare & Full & Rare &  & \\
        \midrule
         MIGC & 16.87 & 18.05 & 17.84 & 19.02 & 30.53 & 27.36 \\
         MIGC(+InteractDiffusion) & \textbf{26.04} & \textbf{21.73} & \textbf{27.02} & \textbf{22.89} & \textbf{22.32} & \textbf{27.55} \\
         \midrule
         \midrule
         \textcolor{gray}{InteractDiffusion} & \textcolor{gray}{29.53} & \textcolor{gray}{23.02} & \textcolor{gray}{30.99} & \textcolor{gray}{24.93} & \textcolor{gray}{18.69} & \textcolor{gray}{26.91} \\
         \textcolor{gray}{InteractDiffusion(+MIGC)} & \textcolor{gray}{31.40} & \textcolor{gray}{24.52} & \textcolor{gray}{32.76} & \textcolor{gray}{26.32} & \textcolor{gray}{18.35} & \textcolor{gray}{27.18} \\
        \bottomrule
    \end{tabular}
    }
    \label{migcinteracttable}
    \vspace{-0.2in}
\end{table}
\paragraph{InteractDiffusion and MIGC.} InteractDiffusion \citep{hoe2024interactdiffusion} is a T2I model that combines a pretrained Stable Diffusion (SD) model with a locally controlled interaction mechanism, enabling fine-grained control over the generated images and demonstrating effective interactivity. MIGC \citep{zhou2024migc} is a T2I model that employs a divide-and-conquer strategy, achieving excellent performance in both attribute representation and isolation of generated instances.
InteractDiffusion primarily focuses on controlling subject-object interactions. However, due to the lack of explicit constraints on object attributes within the model architecture and dataset design, it exhibits suboptimal performance in attribute control. To address this, we attempt to aggregate the features of MIGC \( p_{\theta_2} \) into InteractDiffusion \( p_{\theta_1} \) by applying the AMDM algorithm, introducing attribute control information, and denoting this as InteractDiffusion(+MIGC).  

\begin{figure}[!t]
    \centering
    \includegraphics[width=1.0\columnwidth]{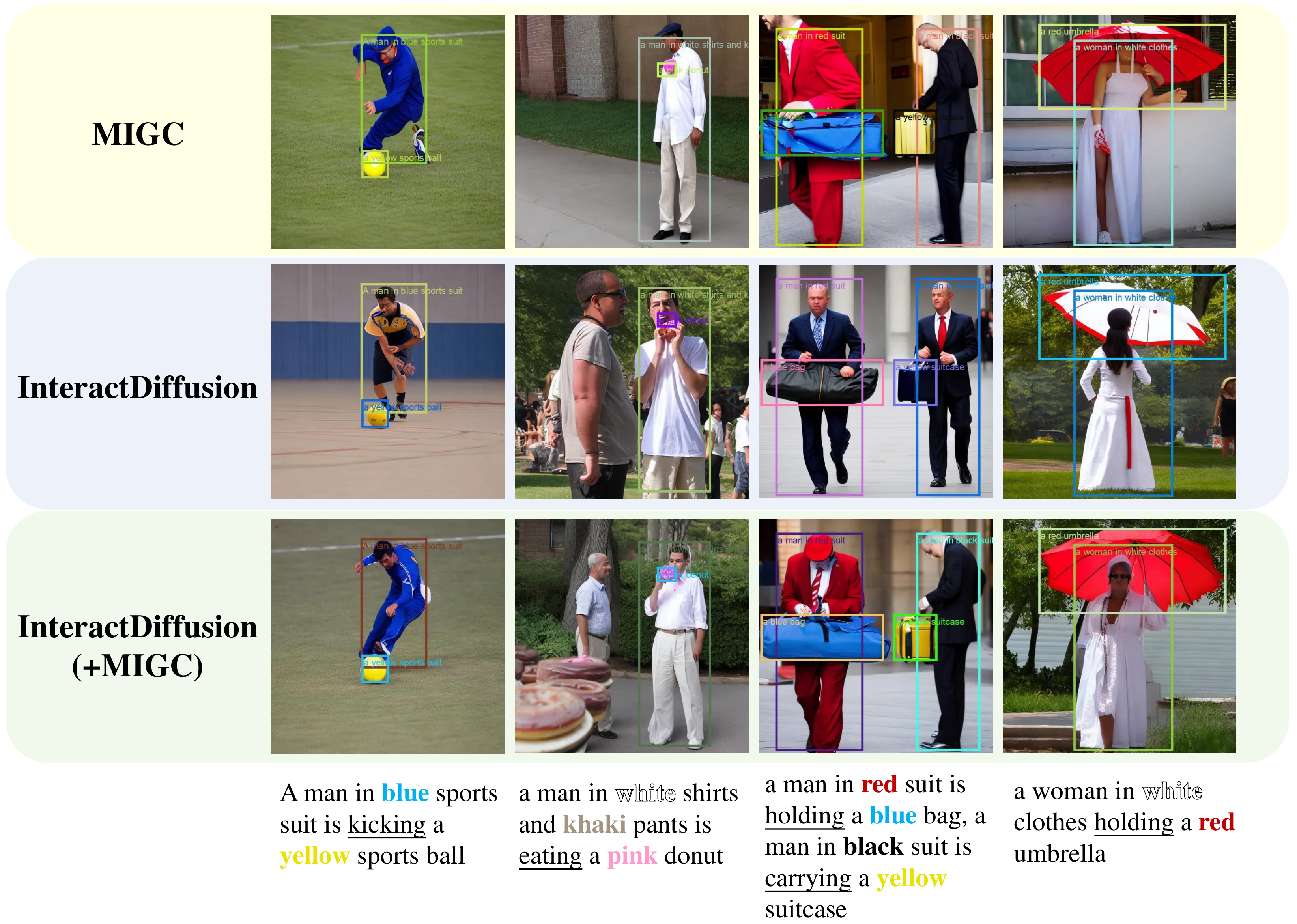}
    \caption{Visual results of aggregating MIGC into InteractDiffusion.}
    \label{interactmigc}
\end{figure}

Visual results are shown in Figure \ref{interactmigc}. It is evident that aggregating the MIGC model into InteractDiffusion using our proposed AMDM algorithm significantly enhances its learned representations, leading to a notable improvement in instance attribute control, and confirming the algorithm's effectiveness.

MIGC demonstrates strong attribute control on the COCO-MIG benchmark \citep{zhou2024migc}, while InteractDiffusion mainly uses the FGAHOI \citep{ma2023fgahoi} for Human-Object Interaction (HOI) detection to show its control over interaction. To further validate the effectiveness of AMDM, we evaluate whether MIGC(+InteractDiffusion) can significantly enhance HOI interaction, ideally reaching the performance level of InteractDiffusion while introducing only minor trade-offs in attribute control on COCO-MIG. 
Conversely, we also examine whether InteractDiffusion(+MIGC) meets the experimental objectives.

\begin{figure}[!t]
    \centering
    \includegraphics[width=1.0\columnwidth]{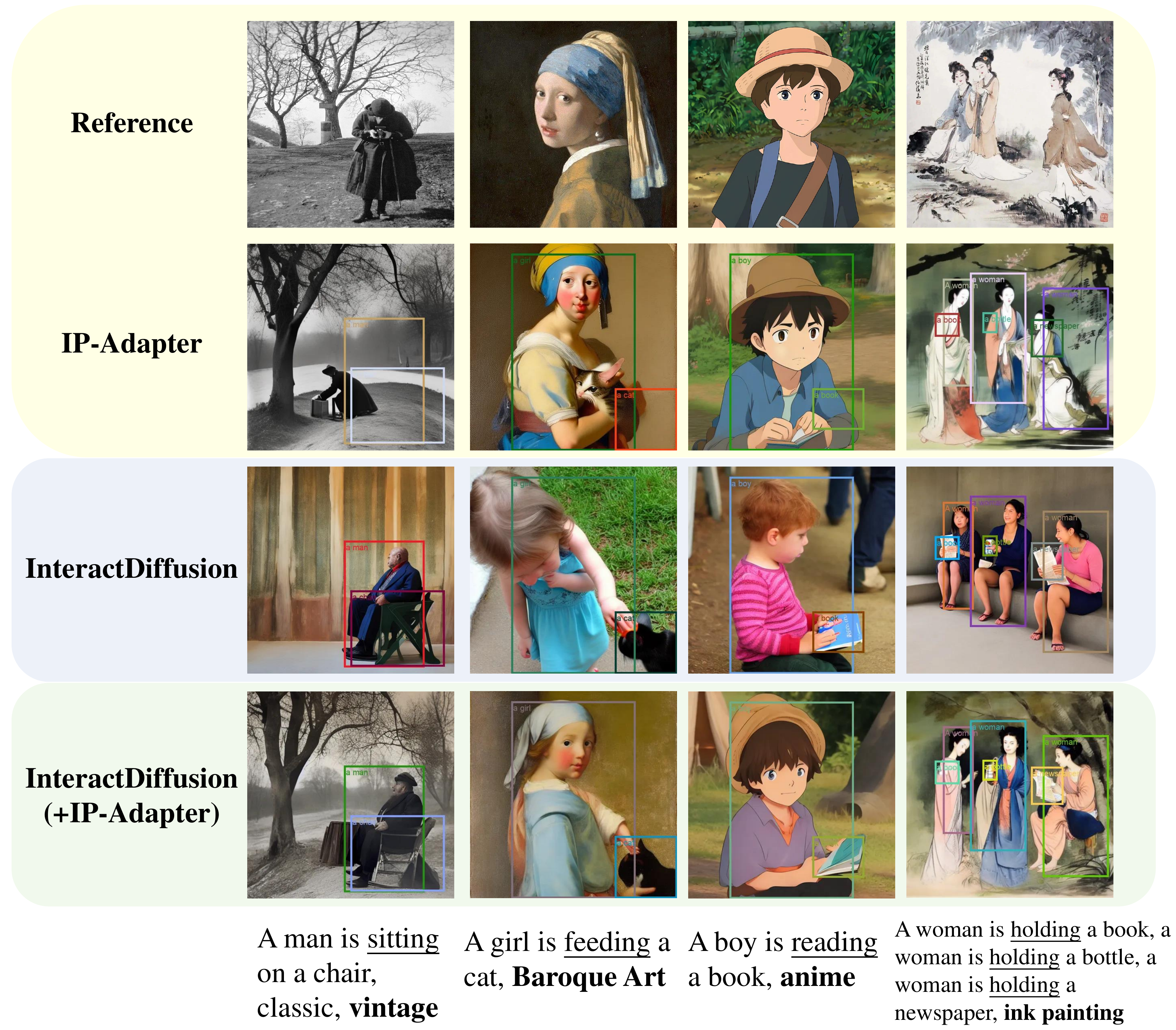}
    \caption{Visual results of aggregating IP-Adapter into InteractDiffusion.}
    \label{interactip}
\end{figure}

\begin{figure}[!t]
\centering
\includegraphics[width=1.0\columnwidth]{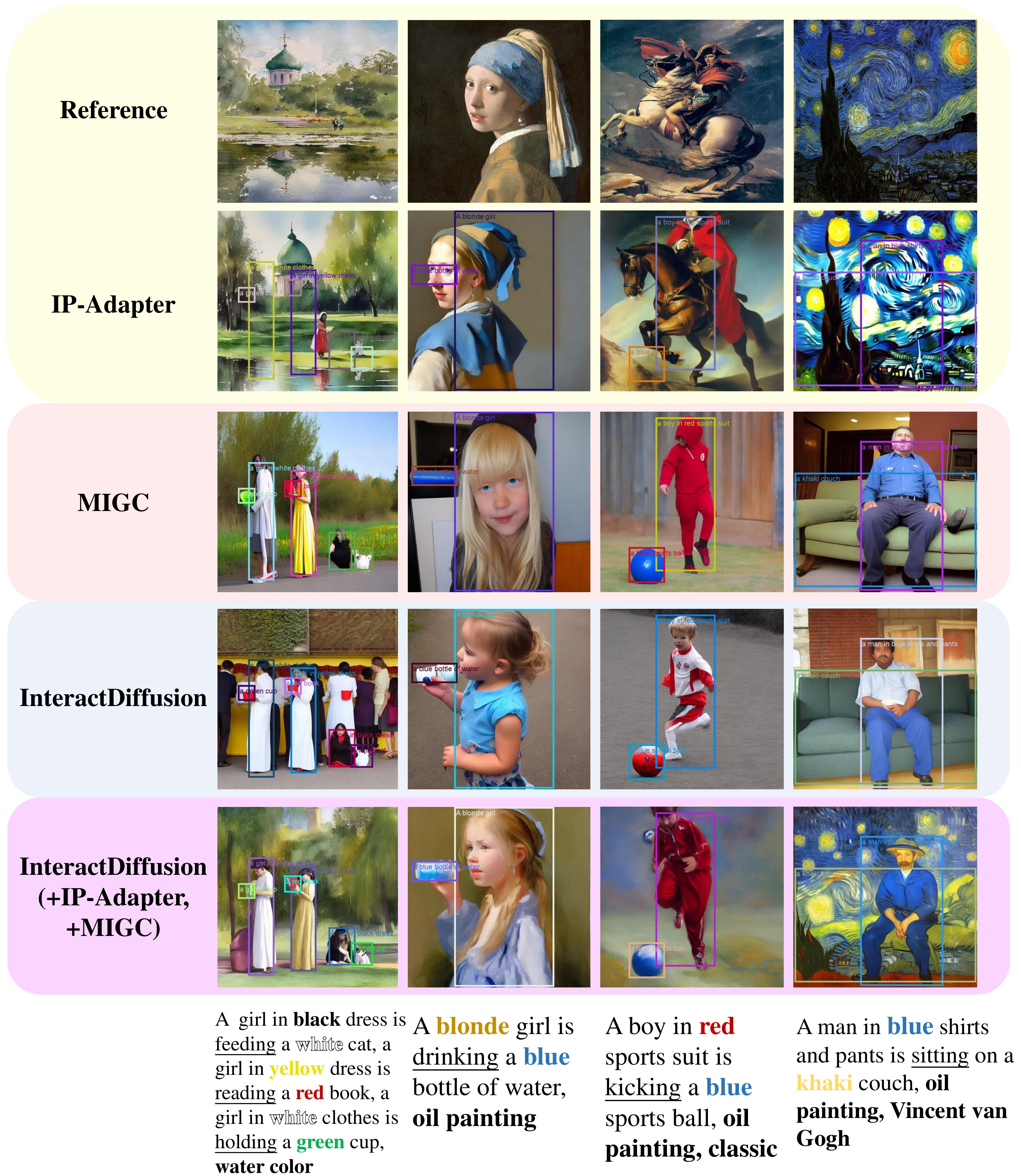}
\caption{Visual results of aggregating MIGC and IP-Adapter into InteractDiffusion.}
\label{interactmigcip}
\end{figure}

The quantitative results of InteractDiffusion(+MIGC) are shown in the first two rows of Table \ref{interactmigctable}. It can be observed that all metrics have significantly improved in InteractDiffusion(+MIGC) in attribute control. Interestingly, as shown in the last two rows of Table \ref{migcinteracttable}, it even surpasses the original InteractDiffusion model in terms of interaction. Similarly, the first two rows of Table \ref{migcinteracttable} and the last two rows of Table \ref{interactmigctable} show that MIGC(+InteractDiffusion) achieves significant improvements across all metrics in interaction capability control, while incurring only minor trade-offs in attribute control, thereby validating the effectiveness of the algorithm.

% Similarly, the results of MIGC(+InteractDiffusion) are shown in the first two rows of Table \ref{migcinteracttable}. 
% It can be seen that MIGC(+InteractDiffusion) also shows significant improvements across all metrics in the interaction capability control. Additionally, as observed in the last two rows of Table \ref{interactmigctable}, the MIGC(+InteractDiffusion) maintains the original performance of MIGC in attribute control, validating the effectiveness of the algorithm.

\begin{table*}[!t]
    \caption{Quantitative results on the COCO-MIG benchmark and CLIP Score across different models.}
    \centering
    \resizebox{\textwidth}{!}{
    \begin{tabular}{c|c c c c c c |c c c c c c | c c}
        \toprule
        \multirow{2}{*}{\textbf{Method}} & \multicolumn{6}{c|}{\textbf{Instance Success Rate (\%) $\uparrow$ }} & \multicolumn{6}{c|}{\textbf{mIoU Score (\%) $\uparrow$}} & \multicolumn{2}{c}{\textbf{CLIP Score $\uparrow$}} \\ 
         \cmidrule(lr){2-7} \cmidrule(lr){8-13} \cmidrule(lr){14-15}
         & $L_2$ & $L_3$ & $L_4$ & $L_5$ & $L_6$ & Avg & $L_2$ & $L_3$ & $L_4$ & $L_5$ & $L_6$ & Avg & Global & Local \\
        \midrule
        InteractDiffusion  & 37.50 & 35.62 & 35.31 & 30.62 & 34.16 & \cellcolor[gray]{0.9}34.06 & 32.98 & 31.63 & 30.82 & 28.29 & 30.40 & \cellcolor[gray]{0.9}30.40 & 31.09 & 27.56 \\
        InteractDiffusion(+MIGC)  & \textbf{62.29} & \textbf{54.33} & \textbf{56.31} & \textbf{53.87} & \textbf{52.89} & \cellcolor[gray]{0.9}\textbf{54.78} & \textbf{54.30} & \textbf{46.37} & \textbf{48.44} & \textbf{47.65} & \textbf{46.64} & \cellcolor[gray]{0.9}\textbf{47.74} & \textbf{32.81} & \textbf{28.96} \\
        \midrule
        \midrule
        \textcolor{gray}{MIGC} & \textcolor{gray}{64.06} & \textcolor{gray}{56.04} & \textcolor{gray}{58.43} & \textcolor{gray}{56.00} & \textcolor{gray}{49.89} & \textcolor{gray}{55.46} & \textcolor{gray}{54.43} & \textcolor{gray}{49.33} & \textcolor{gray}{50.48} & \textcolor{gray}{48.67} & \textcolor{gray}{44.74} & \textcolor{gray}{48.53} & \textcolor{gray}{32.78} & \textcolor{gray}{28.61} \\
        \textcolor{gray}{MIGC(+InteractDiffusion)} & \textcolor{gray}{60.00} & \textcolor{gray}{50.20} & \textcolor{gray}{50.46} & \textcolor{gray}{48.25} & \textcolor{gray}{46.97} & \textcolor{gray}{49.78} & \textcolor{gray}{52.58} & \textcolor{gray}{44.27} & \textcolor{gray}{43.39} & \textcolor{gray}{42.67} & \textcolor{gray}{41.50} & \textcolor{gray}{43.69} & \textcolor{gray}{32.41} & \textcolor{gray}{28.55} \\
        \bottomrule
    \end{tabular}
    }
    \vspace{8pt}
    \label{interactmigctable}
\end{table*}

\paragraph{InteractDiffusion and IP-Adapter.}
IP-Adapter \citep{ye2023ip} is a lightweight I2I model that employs a decoupled cross-attention mechanism to separately process text and image features, enabling multimodal image generation. Due to its superior performance in preserving the style of the reference image, we propose integrating the style information from IP-Adapter \( p_{\theta_3} \) into InteractDiffusion \( p_{\theta_1} \), denoted as InteractDiffusion(+IP-Adapter). The experimental results are shown in Figure \ref{interactip}. It can be observed that IP-Adapter enhances the style representations of InteractDiffusion, fully activating its style controllability, which further validates the effectiveness of the algorithm.

\paragraph{InteractDiffusion, MIGC and IP-Adapter.}
Furthermore, we attempt to aggregate the attribute features from MIGC \(p_{\theta_2}\) and the style features from IP-Adapter \(p_{\theta_3}\) into InteractDiffusion \(p_{\theta_1}\) to evaluate the effectiveness of the AMDM algorithm. The experimental results are presented in Figure \ref{interactmigcip}.

% Furthermore, the fact that fine-grained generation can be achieved by aggregating only the first $s$ steps indicates that diffusion models initially focus on features such as position, attributes, and style, while the later stages emphasize generation quality and consistency. Vice versa, this allows aggregation algorithms to be used only in the first $s$ steps, significantly reducing computational overhead.

\subsection{Ablation Studies}
\begin{figure}[!t]
    \centering
    \includegraphics[width=1.0\columnwidth]{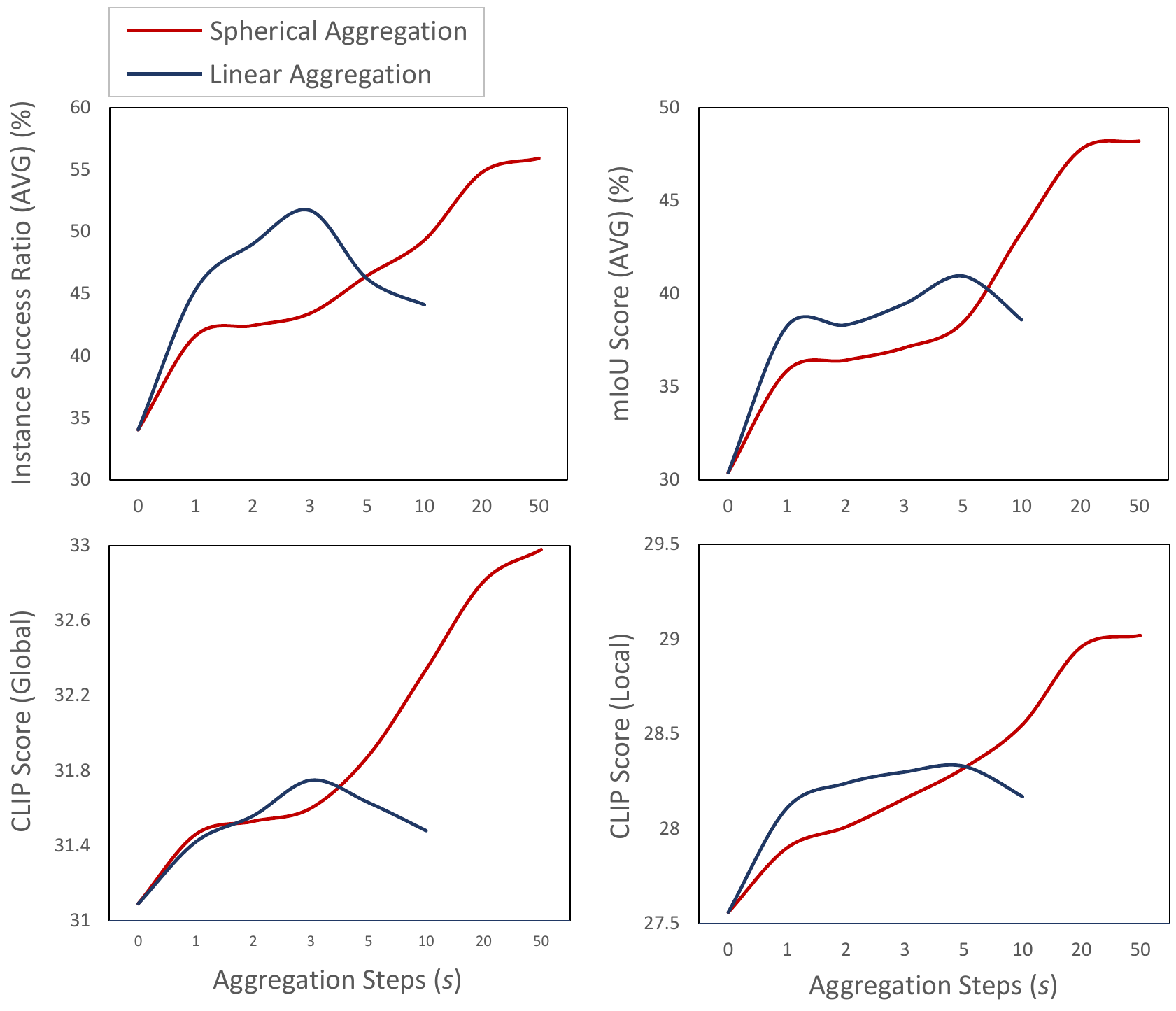}
    \caption{Metrics of spherical aggregation and linear aggregation under different aggregation steps.}
    \label{ablation1}
\end{figure}
\begin{figure}[!t]
    \centering
    \includegraphics[width=1.0\columnwidth]{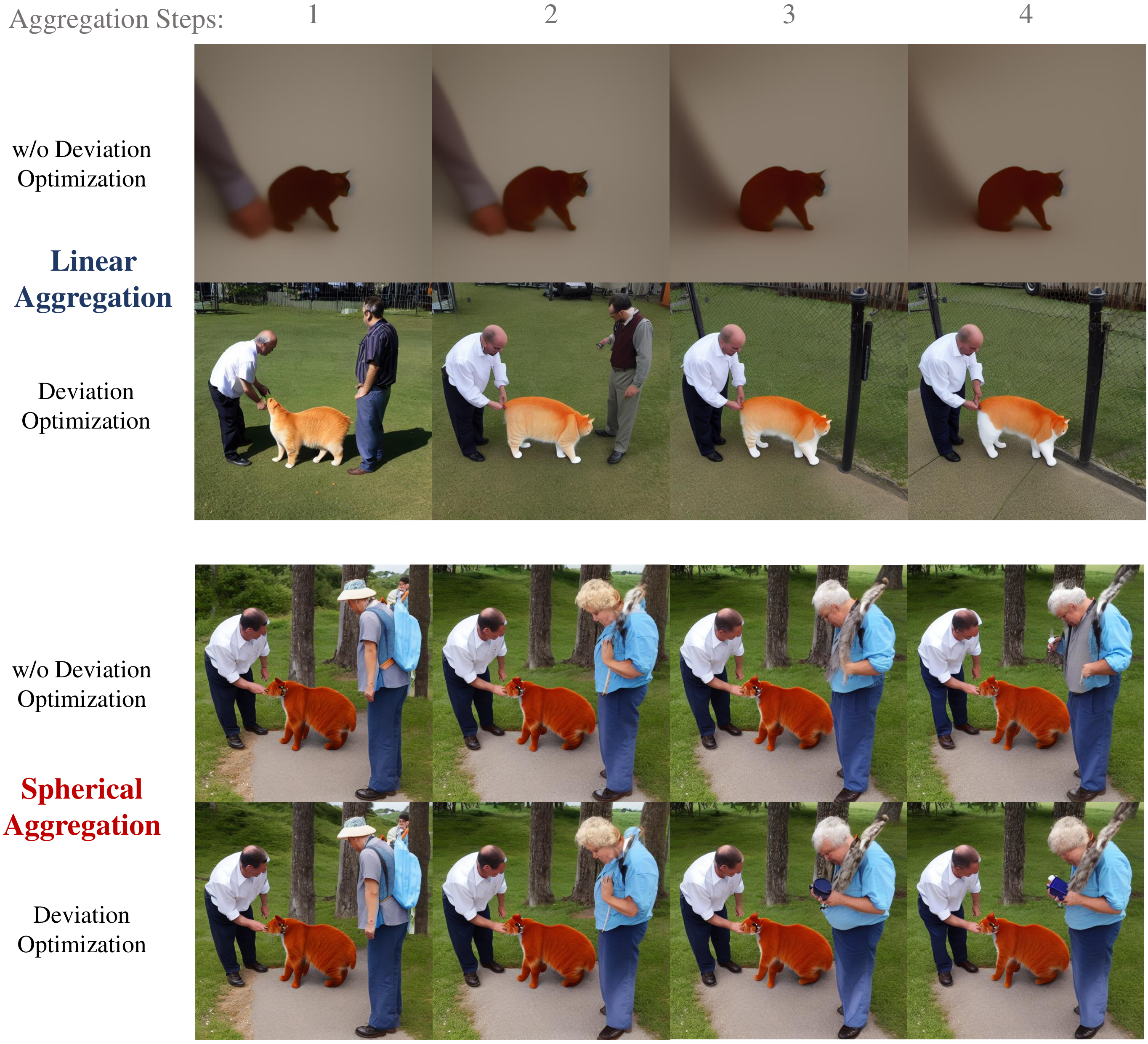}
    \caption{Visual results of deviation optimization for spherical aggregation and linear aggregation under different aggregation steps.}
    \label{ablation2}
\end{figure}

We first attempt a comparison with linear aggregation of InteractDiffusion(+MIGC) which has been of broad interest in compositional generation and the results are shown in Figure \ref{ablation1}. It can be observed that when the number of aggregation steps is small $(s < 5)$, linear aggregation performs slightly better than spherical aggregation. However, as the number of aggregation steps increases, spherical aggregation significantly outperforms linear aggregation across various metrics. This is because, with more aggregation steps, the cumulative error due to the deviation in the data manifold becomes increasingly larger in linear aggregation, which negatively impacts the image quality. 
As shown in Figure \ref{ablation2}, the characteristic of manifold deviation in linear aggregation is evident. In contrast, spherical aggregation consistently minimizes the deviation of the aggregated variables from the spherical manifold, preserving the quality of the final image as much as possible. The deviation optimization further enhances the image quality, demonstrating its effectiveness. A more detailed discussion of linear compositional methods can be found in the Related Work and Appendix \ref{comp}.

We conduct experimental analyses on the absolute value of the norm difference, the angle, and $\delta$, thereby validating the correctness of Proposition \ref{p_normupper} and Proposition \ref{amdm_prop}. 
The results are provided in Appendix \ref{assumpnorm} and Appendix \ref{maxdistance}. 
Moreover, as shown in Figure \ref{ablation1}, the effectiveness of spherical aggregation consistently increases over time and remains almost entirely on the manifold, further validating our inference regarding the local spherical manifold in spherical aggregation.

We also investigated the optimization step of deviation optimization, with experimental results shown in Table \ref{ablation3}. Obviously, when the optimization step is set to 0.3, the best performance is achieved across all metrics. Moreover, the results outperform those achieved with 50 aggregation steps without optimization.

\begin{table*}[!ht]
    \caption{Ablation results of different deviation optimization steps on COCO-MIG benchmark in InteractDiffusion(+MIGC).}
    \centering
    \resizebox{\textwidth}{!}{
    \begin{tabular}{c|c c c c c c |c c c c c c | c c }
        \toprule
        \multirow{2}{*}{\textbf{Deviation Optimization Step}} & \multicolumn{6}{c|}{\textbf{Instance Success Rate (\%) $\uparrow$ }} & \multicolumn{6}{c|}{\textbf{mIoU Score (\%) $\uparrow$}} & \multicolumn{2}{c}{\textbf{CLIP Score $\uparrow$}} \\ 
         \cmidrule(lr){2-7} \cmidrule(lr){8-13} \cmidrule(lr){14-15}
          & $L_2$ & $L_3$ & $L_4$ & $L_5$ & $L_6$ & Avg & $L_2$ & $L_3$ & $L_4$ & $L_5$ & $L_6$ & Avg & Global & Local \\
        \midrule
        0  & 61.56 & 52.08 & 53.75 & 51.62 & 51.87 & \cellcolor[gray]{0.9}53.18 & \textbf{54.76} & 45.09 & 46.36 & 45.48 & 45.79 & \cellcolor[gray]{0.9}46.62 & 31.09 & 27.56 \\
        0.1  & 57.60 & 53.26 & 53.89 & 52.40 & 52.08 & \cellcolor[gray]{0.9}53.26 & 53.27 & 45.53 & 46.86 & 46.31 & 45.80 & \cellcolor[gray]{0.9}46.83 & 31.43 & 27.87 \\
        0.2  & 59.62 & 54.27 & 54.06 & 51.25 & 52.29 & \cellcolor[gray]{0.9}53.40 & 52.71 & 46.32 & 47.17 & 44.68 & 45.74 & \cellcolor[gray]{0.9}46.59 & 31.92 & 28.45 \\
        0.3  & \textbf{62.29} & \textbf{54.33} & \textbf{56.31} & \textbf{53.87} & \textbf{52.89} & \cellcolor[gray]{0.9}\textbf{54.78} & 54.30 & \textbf{46.37} & \textbf{48.44} & \textbf{47.65} & \textbf{46.64} & \cellcolor[gray]{0.9}\textbf{47.74} & \textbf{32.81} & \textbf{28.96} \\
        0.4  & 61.34 & 53.62 & 55.32 & 52.22 & 52.50 & \cellcolor[gray]{0.9}53.96 & 53.46 & 46.21 & 46.82 & 45.26 & 46.23 & \cellcolor[gray]{0.9}46.78 & 32.23 & 28.55 \\
        0.5  & 58.30 & 53.43 & 55.14 & 51.76 & 51.76 & \cellcolor[gray]{0.9}53.27 & 52.88 & 45.03 & 46.52 & 44.95 & 45.51 & \cellcolor[gray]{0.9}46.14 & 31.15 & 27.78 \\
        \bottomrule
    \end{tabular}
    }
    \label{ablation3}
\end{table*}

\begin{figure}[!t]
    \centering
    \includegraphics[width=1.0\columnwidth]{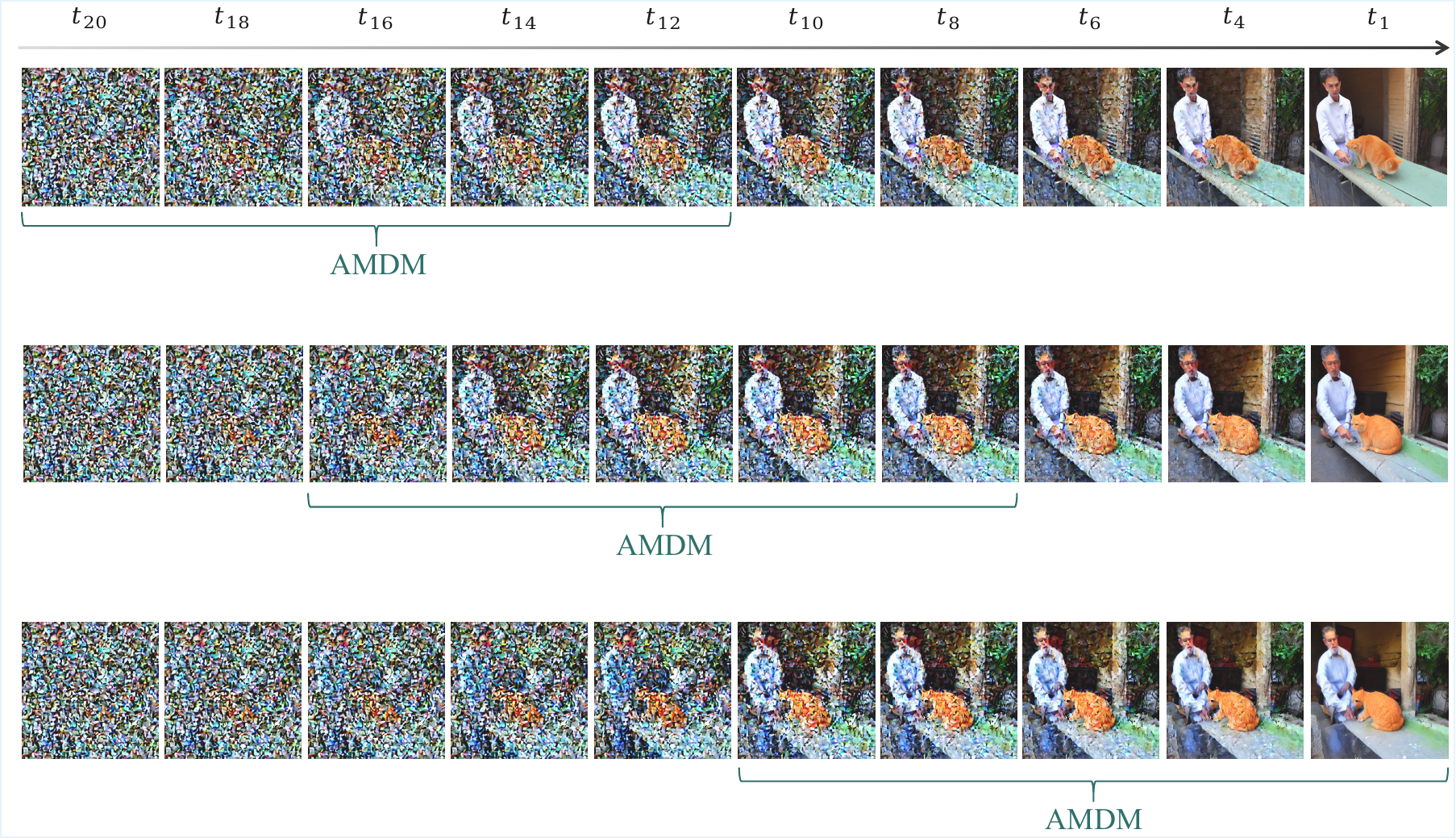}
    \caption{Visual Results of Table \ref{ab_aggregation_stages}.}
    \label{vab}
\end{figure}

Finally, we conduct ablation studies on the aggregation at different stages of the AMDM algorithm. As shown in Table \ref{ab_aggregation_stages} and Figure \ref{vab}, the best results are achieved when aggregation occurs during the initial stages of sampling. At other stages, especially the final ones, this leads to a noticeable drop in generation quality, with weaker control over attributes and interactions. More experimental results are provided in Appendix \ref{visualab}. This supports the claim that diffusion models initially focus on generating features such as position, attributes, and style, while later stages emphasize quality and consistency. 

% Next, We visualize the Table \ref{ab_aggregation_stages} in Figure \ref{vab}, diffusion models generally begin by focusing on generating position and shape, and only produce fine details in the final stages. In the first row, the second sampling step already reveals the subject’s shape and location. When aggregation occurs in the latter half, image quality degrades and the picture becomes blurry. Notably, in the third row there is a clear jump between the 5th and 6th images, which severely affects the final result. It is anticipated that if aggregation occurs only in the final few small steps, the generation quality will further deteriorate; 

\begin{table}[!t]
\caption{InteractDiffusion(+MIGC) with total 20 steps sampling and $s{=}10$.}
\centering
\resizebox{1.0\columnwidth}{!}{
\begin{tabular}{c c c c}
\toprule
\textbf{Aggregation Stage} & \textbf{ISR (avg) $\uparrow$} & \textbf{Default (Full) $\uparrow$} & \textbf{FID $\downarrow$} \\
\midrule
$t_{20} \rightarrow t_{10}$ & \textbf{54.78} & \textbf{31.40} & \textbf{18.35} \\
$t_{15} \rightarrow t_{5}$  & 50.14 & 27.53 & 22.96 \\
$t_{10} \rightarrow t_{0}$  & 47.82 & 24.17 & 41.13 \\
\bottomrule
\end{tabular}
}
\label{ab_aggregation_stages}
\end{table}

\section{Related Work}
\paragraph{Fine-Grained Generation.} 
Beyond text-driven models \cite{nichol2022glide, ramesh2022hierarchical, li2024dreamscene, podellsdxl}, research is increasingly moving toward finer-grained conditional control. A classical route is personalization-controlled generation—covering style \cite{sohn2023styledrop, hertz2024style, chen2024artadapter}, subject \cite{li2024blip, shi2024instantbooth, jiang2024videobooth, ye2023ip}, person \cite{xiao2024fastcomposer, giambi2023conditioning, peng2024portraitbooth}, and interactive generation \cite{huang2023reversion, guoanimatediff, hoe2024interactdiffusion}. Another focus is spatial control \cite{li2023gligen, cheng2023layoutdiffuse, niecompositional, zhou2024migc}, which uses bounding boxes or regions as additional conditions to enforce spatial constraints. More recently, several studies have pursued direct fine-grained multimodal control \cite{huang2023composer, smith2023continual, gu2024mix, kumari2023multi}, but these approaches typically require extensive multi-condition datasets and increasingly complex model architectures.

\paragraph{Compositional Generation.} 
% Some recent works \cite{hinton2002training, vedantam2018generative, du2020compositional, liu2022compositional, du2023reduce, duposition, garipov2023compositional} have attempted to combine multiple models with the aim of generating new distributions to enhance model generalization. Some of these studies \cite{du2020compositional, du2023reduce, garipov2023compositional} focus on the combination of different objects, while others are primarily validated on toy examples \cite{hinton2002training, vedantam2018generative, garipov2023compositional} , lacking practical relevance. 
Some recent works \cite{du2020compositional, liu2022compositional, du2023reduce, duposition, garipov2023compositional, bradley2025mechanisms, thornton2025composition, skreta2025feynman,he2025rne,skretasuperposition} aim to create new distributions by combining different distributions, commonly using techniques such as linear weighted score, followed by simulated annealing or other optimization methods.
Other works \cite{biggs2024diffusion, ohdawin, wangensembling} attempt to combine models by applying weighted interpolation to their parameters. However, this approach requires that the models share an identical architecture, which significantly restricts its applicability.
A more in-depth comparison of our AMDM and its relation to compositional generation methods can be found in the Appendix \ref{comp}.

\section{Conclusion}
This paper proposes a novel AMDM algorithm designed for fine-grained generation, which consists of two main components: spherical aggregation and deviation optimization. Experimental results demonstrate the effectiveness of the AMDM algorithm, revealing that diffusion models initially prioritize image feature generation, shifting their focus to image quality and consistency in later stages. The algorithm provides a new perspective for addressing fine-grained generation challenges: We can leverage existing or develop new conditional diffusion models that control specific aspects, and then aggregate them using the AMDM algorithm. This eliminates the need for constructing complex datasets, designing intricate model architectures, and incurring high training costs.
% This paper proposes the AMDM algorithm, a novel approach comprising spherical aggregation and manifold optimization. Experimental results demonstrate the effectiveness of the AMDM algorithm, revealing that diffusion models initially prioritize image feature generation, shifting their focus to image quality and consistency in later stages. The algorithm presents a novel approach to fine-grained conditional controlled generation. By leveraging existing conditional diffusion models or developing new ones to manage specific aspects, fine-grained control is achieved through the AMDM algorithm. This method eliminates the need for constructing complex datasets, designing intricate model architectures, and incurring high training costs.
% 本文提出了一种新的AMDM算法，其主要有两部分组成，球面聚合和流形优化。实验结果不仅证明了其有效性，还表明了扩散模型在前期关注生成图像特征，后期关注生成图像质量和一致性特点。该算法为我们处理复杂细粒度条件控制生成提供了一种全新的视角：我们可以将相同隐空间和加噪方式的,对不同条件控制的扩散模型利用AMDM其聚合起来，实现细粒度控制，避免了构造复杂数据集，设计复杂的模型架构和昂贵的训练成本。

\section*{Impact Statement}
This paper presents work whose goal is to advance the field of Machine
Learning. There are many potential societal consequences of our work, none
which we feel must be specifically highlighted here.

\bibliography{refs}

@inproceedings{sohl2015deep,
  title={Deep unsupervised learning using nonequilibrium thermodynamics},
  author={Sohl-Dickstein, Jascha and Weiss, Eric and Maheswaranathan, Niru and Ganguli, Surya},
  booktitle={International conference on machine learning},
  pages={2256--2265},
  year={2015},
  organization={PMLR}
}

@article{ho2020denoising,
  title={Denoising diffusion probabilistic models},
  author={Ho, Jonathan and Jain, Ajay and Abbeel, Pieter},
  journal={Advances in neural information processing systems},
  volume={33},
  pages={6840--6851},
  year={2020}
}

@inproceedings{songdenoising,
  title={Denoising Diffusion Implicit Models},
  author={Song, Jiaming and Meng, Chenlin and Ermon, Stefano},
  booktitle={International Conference on Learning Representations},
  year={2021}
}

@article{song2020score,
  title={Score-based generative modeling through stochastic differential equations},
  author={Song, Yang and Sohl-Dickstein, Jascha and Kingma, Diederik P and Kumar, Abhishek and Ermon, Stefano and Poole, Ben},
  journal={In Proceedings of International Conference on Learning Representations (ICLR)},
  year={2021}
}

@article{karras2022elucidating,
  title={Elucidating the design space of diffusion-based generative models},
  author={Karras, Tero and Aittala, Miika and Aila, Timo and Laine, Samuli},
  journal={Advances in Neural Information Processing Systems},
  volume={35},
  pages={26565--26577},
  year={2022}
}

@article{dhariwal2021diffusion,
  title={Diffusion models beat gans on image synthesis},
  author={Dhariwal, Prafulla and Nichol, Alexander},
  journal={Advances in neural information processing systems},
  volume={34},
  pages={8780--8794},
  year={2021}
}

@inproceedings{chungdiffusion,
  title={Diffusion Posterior Sampling for General Noisy Inverse Problems},
  author={Chung, Hyungjin and Kim, Jeongsol and Mccann, Michael Thompson and Klasky, Marc Louis and Ye, Jong Chul},
  booktitle={The Eleventh International Conference on Learning Representations},
  year={2023}
}

@inproceedings{rombach2022high,
  title={High-resolution image synthesis with latent diffusion models},
  author={Rombach, Robin and Blattmann, Andreas and Lorenz, Dominik and Esser, Patrick and Ommer, Bj{\"o}rn},
  booktitle={Proceedings of the IEEE/CVF conference on computer vision and pattern recognition},
  pages={10684--10695},
  year={2022}
}

@inproceedings{esser2024scaling,
  title={Scaling rectified flow transformers for high-resolution image synthesis},
  author={Esser, Patrick and Kulal, Sumith and Blattmann, Andreas and Entezari, Rahim and M{\"u}ller, Jonas and Saini, Harry and Levi, Yam and Lorenz, Dominik and Sauer, Axel and Boesel, Frederic and others},
  booktitle={Forty-first International Conference on Machine Learning},
  year={2024}
}

@inproceedings{chenre,
  title={Re-Imagen: Retrieval-Augmented Text-to-Image Generator},
  author={Chen, Wenhu and Hu, Hexiang and Saharia, Chitwan and Cohen, William W},
  booktitle={The Eleventh International Conference on Learning Representations},
  year={2023}
}

@inproceedings{nichol2022glide,
  title={GLIDE: Towards Photorealistic Image Generation and Editing with Text-Guided Diffusion Models},
  author={Nichol, Alexander Quinn and Dhariwal, Prafulla and Ramesh, Aditya and Shyam, Pranav and Mishkin, Pamela and Mcgrew, Bob and Sutskever, Ilya and Chen, Mark},
  booktitle={International Conference on Machine Learning},
  pages={16784--16804},
  year={2022},
  organization={PMLR}
}

@article{lee2024holistic,
  title={Holistic evaluation of text-to-image models},
  author={Lee, Tony and Yasunaga, Michihiro and Meng, Chenlin and Mai, Yifan and Park, Joon Sung and Gupta, Agrim and Zhang, Yunzhi and Narayanan, Deepak and Teufel, Hannah and Bellagente, Marco and others},
  journal={Advances in Neural Information Processing Systems},
  volume={36},
  year={2024}
}

@article{xu2024imagereward,
  title={Imagereward: Learning and evaluating human preferences for text-to-image generation},
  author={Xu, Jiazheng and Liu, Xiao and Wu, Yuchen and Tong, Yuxuan and Li, Qinkai and Ding, Ming and Tang, Jie and Dong, Yuxiao},
  journal={Advances in Neural Information Processing Systems},
  volume={36},
  year={2024}
}

@inproceedings{zhang2023inversion,
  title={Inversion-based style transfer with diffusion models},
  author={Zhang, Yuxin and Huang, Nisha and Tang, Fan and Huang, Haibin and Ma, Chongyang and Dong, Weiming and Xu, Changsheng},
  booktitle={Proceedings of the IEEE/CVF conference on computer vision and pattern recognition},
  pages={10146--10156},
  year={2023}
}

@inproceedings{wang2023stylediffusion,
  title={Stylediffusion: Controllable disentangled style transfer via diffusion models},
  author={Wang, Zhizhong and Zhao, Lei and Xing, Wei},
  booktitle={Proceedings of the IEEE/CVF International Conference on Computer Vision},
  pages={7677--7689},
  year={2023}
}

@inproceedings{mou2024diffeditor,
  title={Diffeditor: Boosting accuracy and flexibility on diffusion-based image editing},
  author={Mou, Chong and Wang, Xintao and Song, Jiechong and Shan, Ying and Zhang, Jian},
  booktitle={Proceedings of the IEEE/CVF Conference on Computer Vision and Pattern Recognition},
  pages={8488--8497},
  year={2024}
}

@inproceedings{zheng2023layoutdiffusion,
  title={Layoutdiffusion: Controllable diffusion model for layout-to-image generation},
  author={Zheng, Guangcong and Zhou, Xianpan and Li, Xuewei and Qi, Zhongang and Shan, Ying and Li, Xi},
  booktitle={Proceedings of the IEEE/CVF Conference on Computer Vision and Pattern Recognition},
  pages={22490--22499},
  year={2023}
}

@inproceedings{chai2023layoutdm,
  title={Layoutdm: Transformer-based diffusion model for layout generation},
  author={Chai, Shang and Zhuang, Liansheng and Yan, Fengying},
  booktitle={Proceedings of the IEEE/CVF Conference on Computer Vision and Pattern Recognition},
  pages={18349--18358},
  year={2023}
}

@article{chen2024textdiffuser,
  title={Textdiffuser: Diffusion models as text painters},
  author={Chen, Jingye and Huang, Yupan and Lv, Tengchao and Cui, Lei and Chen, Qifeng and Wei, Furu},
  journal={Advances in Neural Information Processing Systems},
  volume={36},
  year={2024}
}

@inproceedings{wu2023uncovering,
  title={Uncovering the disentanglement capability in text-to-image diffusion models},
  author={Wu, Qiucheng and Liu, Yujian and Zhao, Handong and Kale, Ajinkya and Bui, Trung and Yu, Tong and Lin, Zhe and Zhang, Yang and Chang, Shiyu},
  booktitle={Proceedings of the IEEE/CVF conference on computer vision and pattern recognition},
  pages={1900--1910},
  year={2023}
}

@inproceedings{wang2024compositional,
  title={Compositional text-to-image synthesis with attention map control of diffusion models},
  author={Wang, Ruichen and Chen, Zekang and Chen, Chen and Ma, Jian and Lu, Haonan and Lin, Xiaodong},
  booktitle={Proceedings of the AAAI Conference on Artificial Intelligence},
  volume={38},
  pages={5544--5552},
  year={2024}
}

@inproceedings{hoe2024interactdiffusion,
  title={InteractDiffusion: Interaction Control in Text-to-Image Diffusion Models},
  author={Hoe, Jiun Tian and Jiang, Xudong and Chan, Chee Seng and Tan, Yap-Peng and Hu, Weipeng},
  booktitle={Proceedings of the IEEE/CVF Conference on Computer Vision and Pattern Recognition},
  pages={6180--6189},
  year={2024}
}

@inproceedings{jia2024customizing,
  title={Customizing Text-to-Image Generation with Inverted Interaction},
  author={Jia, Xu and Isobe, Takashi and Li, Xiaomin and Wang, Qinghe and Mu, Jing and Zhou, Dong and Lu, Huchuan and Tian, Lu and Sirasao, Ashish and Barsoum, Emad and others},
  booktitle={ACM Multimedia 2024},
  year={2024}
}

@article{huang2024diffstyler,
  title={Diffstyler: Controllable dual diffusion for text-driven image stylization},
  author={Huang, Nisha and Zhang, Yuxin and Tang, Fan and Ma, Chongyang and Huang, Haibin and Dong, Weiming and Xu, Changsheng},
  journal={IEEE Transactions on Neural Networks and Learning Systems},
  year={2024},
  publisher={IEEE}
}

@inproceedings{qi2024deadiff,
  title={DEADiff: An Efficient Stylization Diffusion Model with Disentangled Representations},
  author={Qi, Tianhao and Fang, Shancheng and Wu, Yanze and Xie, Hongtao and Liu, Jiawei and Chen, Lang and He, Qian and Zhang, Yongdong},
  booktitle={Proceedings of the IEEE/CVF Conference on Computer Vision and Pattern Recognition},
  pages={8693--8702},
  year={2024}
}

@inproceedings{chung2022improving,
  title={Improving Diffusion Models for Inverse Problems using Manifold Constraints},
  author={Chung, Hyungjin and Sim, Byeongsu and Ryu, Dohoon and Ye, Jong Chul},
  booktitle={Advances in Neural Information Processing Systems},
  year={2022}
}

@article{ramesh2022hierarchical,
  title={Hierarchical text-conditional image generation with clip latents},
  author={Ramesh, Aditya and Dhariwal, Prafulla and Nichol, Alex and Chu, Casey and Chen, Mark},
  journal={arXiv preprint arXiv:2204.06125},
  volume={1},
  number={2},
  pages={3},
  year={2022}
}

@inproceedings{podellsdxl,
  title={SDXL: Improving Latent Diffusion Models for High-Resolution Image Synthesis},
  author={Podell, Dustin and English, Zion and Lacey, Kyle and Blattmann, Andreas and Dockhorn, Tim and M{\"u}ller, Jonas and Penna, Joe and Rombach, Robin},
  booktitle={The Twelfth International Conference on Learning Representations},
  year={2024}
}

@inproceedings{sohn2023styledrop,
  title={StyleDrop: text-to-image generation in any style},
  author={Sohn, Kihyuk and Ruiz, Nataniel and Lee, Kimin and Chin, Daniel Castro and Blok, Irina and Chang, Huiwen and Barber, Jarred and Jiang, Lu and Entis, Glenn and Li, Yuanzhen and others},
  booktitle={Proceedings of the 37th International Conference on Neural Information Processing Systems},
  pages={66860--66889},
  year={2023}
}

@inproceedings{hertz2024style,
  title={Style aligned image generation via shared attention},
  author={Hertz, Amir and Voynov, Andrey and Fruchter, Shlomi and Cohen-Or, Daniel},
  booktitle={Proceedings of the IEEE/CVF Conference on Computer Vision and Pattern Recognition},
  pages={4775--4785},
  year={2024}
}

@inproceedings{chen2024artadapter,
  title={ArtAdapter: Text-to-Image Style Transfer using Multi-Level Style Encoder and Explicit Adaptation},
  author={Chen, Dar-Yen and Tennent, Hamish and Hsu, Ching-Wen},
  booktitle={Proceedings of the IEEE/CVF Conference on Computer Vision and Pattern Recognition},
  pages={8619--8628},
  year={2024}
}

@article{li2024blip,
  title={Blip-diffusion: Pre-trained subject representation for controllable text-to-image generation and editing},
  author={Li, Dongxu and Li, Junnan and Hoi, Steven},
  journal={Advances in Neural Information Processing Systems},
  volume={36},
  year={2024}
}

@inproceedings{shi2024instantbooth,
  title={Instantbooth: Personalized text-to-image generation without test-time finetuning},
  author={Shi, Jing and Xiong, Wei and Lin, Zhe and Jung, Hyun Joon},
  booktitle={Proceedings of the IEEE/CVF Conference on Computer Vision and Pattern Recognition},
  pages={8543--8552},
  year={2024}
}

@inproceedings{jiang2024videobooth,
  title={Videobooth: Diffusion-based video generation with image prompts},
  author={Jiang, Yuming and Wu, Tianxing and Yang, Shuai and Si, Chenyang and Lin, Dahua and Qiao, Yu and Loy, Chen Change and Liu, Ziwei},
  booktitle={Proceedings of the IEEE/CVF Conference on Computer Vision and Pattern Recognition},
  pages={6689--6700},
  year={2024}
}

@article{xiao2024fastcomposer,
  title={Fastcomposer: Tuning-free multi-subject image generation with localized attention},
  author={Xiao, Guangxuan and Yin, Tianwei and Freeman, William T and Durand, Fr{\'e}do and Han, Song},
  journal={International Journal of Computer Vision},
  pages={1--20},
  year={2024},
  publisher={Springer}
}

@article{giambi2023conditioning,
  title={Conditioning diffusion models via attributes and semantic masks for face generation},
  author={Giambi, Nico and Lisanti, Giuseppe},
  journal={arXiv preprint arXiv:2306.00914},
  year={2023}
}

@inproceedings{peng2024portraitbooth,
  title={Portraitbooth: A versatile portrait model for fast identity-preserved personalization},
  author={Peng, Xu and Zhu, Junwei and Jiang, Boyuan and Tai, Ying and Luo, Donghao and Zhang, Jiangning and Lin, Wei and Jin, Taisong and Wang, Chengjie and Ji, Rongrong},
  booktitle={Proceedings of the IEEE/CVF Conference on Computer Vision and Pattern Recognition},
  pages={27080--27090},
  year={2024}
}

@article{huang2023reversion,
  title={ReVersion: Diffusion-based relation inversion from images},
  author={Huang, Ziqi and Wu, Tianxing and Jiang, Yuming and Chan, Kelvin CK and Liu, Ziwei},
  journal={arXiv preprint arXiv:2303.13495},
  year={2023}
}

@inproceedings{guoanimatediff,
  title={AnimateDiff: Animate Your Personalized Text-to-Image Diffusion Models without Specific Tuning},
  author={Guo, Yuwei and Yang, Ceyuan and Rao, Anyi and Liang, Zhengyang and Wang, Yaohui and Qiao, Yu and Agrawala, Maneesh and Lin, Dahua and Dai, Bo},
  booktitle={The Twelfth International Conference on Learning Representations},
  year={2024}
}

@article{ye2023ip,
  title={Ip-adapter: Text compatible image prompt adapter for text-to-image diffusion models},
  author={Ye, Hu and Zhang, Jun and Liu, Sibo and Han, Xiao and Yang, Wei},
  journal={arXiv preprint arXiv:2308.06721},
  year={2023}
}

@inproceedings{li2023gligen,
  title={Gligen: Open-set grounded text-to-image generation},
  author={Li, Yuheng and Liu, Haotian and Wu, Qingyang and Mu, Fangzhou and Yang, Jianwei and Gao, Jianfeng and Li, Chunyuan and Lee, Yong Jae},
  booktitle={Proceedings of the IEEE/CVF Conference on Computer Vision and Pattern Recognition},
  pages={22511--22521},
  year={2023}
}

@article{cheng2023layoutdiffuse,
  title={Layoutdiffuse: Adapting foundational diffusion models for layout-to-image generation},
  author={Cheng, Jiaxin and Liang, Xiao and Shi, Xingjian and He, Tong and Xiao, Tianjun and Li, Mu},
  journal={arXiv preprint arXiv:2302.08908},
  year={2023}
}

@inproceedings{niecompositional,
  title={Compositional Text-to-Image Generation with Dense Blob Representations},
  author={Nie, Weili and Liu, Sifei and Mardani, Morteza and Liu, Chao and Eckart, Benjamin and Vahdat, Arash},
  booktitle={Forty-first International Conference on Machine Learning},
  year={2024}
}

@inproceedings{zhou2024migc,
  title={Migc: Multi-instance generation controller for text-to-image synthesis},
  author={Zhou, Dewei and Li, You and Ma, Fan and Zhang, Xiaoting and Yang, Yi},
  booktitle={Proceedings of the IEEE/CVF Conference on Computer Vision and Pattern Recognition},
  pages={6818--6828},
  year={2024}
}

@inproceedings{huang2023composer,
  title={Composer: creative and controllable image synthesis with composable conditions},
  author={Huang, Lianghua and Chen, Di and Liu, Yu and Shen, Yujun and Zhao, Deli and Zhou, Jingren},
  booktitle={Proceedings of the 40th International Conference on Machine Learning},
  pages={13753--13773},
  year={2023}
}

@article{smith2023continual,
  title={Continual diffusion: Continual customization of text-to-image diffusion with c-lora},
  author={Smith, James Seale and Hsu, Yen-Chang and Zhang, Lingyu and Hua, Ting and Kira, Zsolt and Shen, Yilin and Jin, Hongxia},
  journal={arXiv preprint arXiv:2304.06027},
  year={2023}
}

@article{gu2024mix,
  title={Mix-of-show: Decentralized low-rank adaptation for multi-concept customization of diffusion models},
  author={Gu, Yuchao and Wang, Xintao and Wu, Jay Zhangjie and Shi, Yujun and Chen, Yunpeng and Fan, Zihan and Xiao, Wuyou and Zhao, Rui and Chang, Shuning and Wu, Weijia and others},
  journal={Advances in Neural Information Processing Systems},
  volume={36},
  year={2024}
}

@inproceedings{kumari2023multi,
  title={Multi-concept customization of text-to-image diffusion},
  author={Kumari, Nupur and Zhang, Bingliang and Zhang, Richard and Shechtman, Eli and Zhu, Jun-Yan},
  booktitle={Proceedings of the IEEE/CVF Conference on Computer Vision and Pattern Recognition},
  pages={1931--1941},
  year={2023}
}

@article{li2024dreamscene,
  title={DreamScene: 3D Gaussian-based Text-to-3D Scene Generation via Formation Pattern Sampling},
  author={Li, Haoran and Shi, Haolin and Zhang, Wenli and Wu, Wenjun and Liao, Yong and Wang, Lin and Lee, Lik-hang and Zhou, Pengyuan},
  journal={arXiv preprint arXiv:2404.03575},
  year={2024}
}

@inproceedings{wang2024instancediffusion,
  title={Instancediffusion: Instance-level control for image generation},
  author={Wang, Xudong and Darrell, Trevor and Rambhatla, Sai Saketh and Girdhar, Rohit and Misra, Ishan},
  booktitle={Proceedings of the IEEE/CVF Conference on Computer Vision and Pattern Recognition},
  pages={6232--6242},
  year={2024}
}

@article{ma2023fgahoi,
  title={Fgahoi: Fine-grained anchors for human-object interaction detection},
  author={Ma, Shuailei and Wang, Yuefeng and Wang, Shanze and Wei, Ying},
  journal={IEEE Transactions on Pattern Analysis and Machine Intelligence},
  year={2023},
  publisher={IEEE}
}

@inproceedings{du2023reduce,
  title={Reduce, reuse, recycle: Compositional generation with energy-based diffusion models and mcmc},
  author={Du, Yilun and Durkan, Conor and Strudel, Robin and Tenenbaum, Joshua B and Dieleman, Sander and Fergus, Rob and Sohl-Dickstein, Jascha and Doucet, Arnaud and Grathwohl, Will Sussman},
  booktitle={International conference on machine learning},
  pages={8489--8510},
  year={2023},
  organization={PMLR}
}

@inproceedings{liu2022compositional,
  title={Compositional visual generation with composable diffusion models},
  author={Liu, Nan and Li, Shuang and Du, Yilun and Torralba, Antonio and Tenenbaum, Joshua B},
  booktitle={European Conference on Computer Vision},
  pages={423--439},
  year={2022},
  organization={Springer}
}

@inproceedings{duposition,
  title={Position: Compositional Generative Modeling: A Single Model is Not All You Need},
  author={Du, Yilun and Kaelbling, Leslie Pack},
  booktitle={Forty-first International Conference on Machine Learning},
  year={2024}
}

@article{du2020compositional,
  title={Compositional visual generation with energy based models},
  author={Du, Yilun and Li, Shuang and Mordatch, Igor},
  journal={Advances in Neural Information Processing Systems},
  volume={33},
  pages={6637--6647},
  year={2020}
}

@article{garipov2023compositional,
  title={Compositional sculpting of iterative generative processes},
  author={Garipov, Timur and De Peuter, Sebastiaan and Yang, Ge and Garg, Vikas and Kaski, Samuel and Jaakkola, Tommi},
  journal={Advances in neural information processing systems},
  volume={36},
  pages={12665--12702},
  year={2023}
}

@book{sarkka2019applied,
  title={Applied stochastic differential equations},
  author={S{\"a}rkk{\"a}, Simo and Solin, Arno},
  volume={10},
  year={2019},
  publisher={Cambridge University Press}
}

@article{bradley2025mechanisms,
  title={Mechanisms of Projective Composition of Diffusion Models},
  author={Bradley, Arwen and Nakkiran, Preetum and Berthelot, David and Thornton, James and Susskind, Joshua M},
  journal={arXiv preprint arXiv:2502.04549},
  year={2025}
}

@article{thornton2025composition,
  title={Composition and Control with Distilled Energy Diffusion Models and Sequential Monte Carlo},
  author={Thornton, James and B{\'e}thune, Louis and Zhang, Ruixiang and Bradley, Arwen and Nakkiran, Preetum and Zhai, Shuangfei},
  journal={arXiv preprint arXiv:2502.12786},
  year={2025}
}

@article{skreta2025feynman,
  title={Feynman-kac correctors in diffusion: Annealing, guidance, and product of experts},
  author={Skreta, Marta and Akhound-Sadegh, Tara and Ohanesian, Viktor and Bondesan, Roberto and Aspuru-Guzik, Al{\'a}n and Doucet, Arnaud and Brekelmans, Rob and Tong, Alexander and Neklyudov, Kirill},
  journal={arXiv preprint arXiv:2503.02819},
  year={2025}
}

@article{kingma2013auto,
  title={Auto-encoding variational bayes},
  author={Kingma, Diederik P and Welling, Max},
  journal={arXiv preprint arXiv:1312.6114},
  year={2013}
}

@inproceedings{biggs2024diffusion,
  title={Diffusion soup: Model merging for text-to-image diffusion models},
  author={Biggs, Benjamin and Seshadri, Arjun and Zou, Yang and Jain, Achin and Golatkar, Aditya and Xie, Yusheng and Achille, Alessandro and Swaminathan, Ashwin and Soatto, Stefano},
  booktitle={European Conference on Computer Vision},
  pages={257--274},
  year={2024},
  organization={Springer}
}

@inproceedings{skretasuperposition,
  title={The Superposition of Diffusion Models Using the It{\^o} Density Estimator},
  author={Skreta, Marta and Atanackovic, Lazar and Bose, Joey and Tong, Alexander and Neklyudov, Kirill},
  booktitle={The Thirteenth International Conference on Learning Representations},
  year={2025},
}

@article{he2025rne,
  title={RNE: a plug-and-play framework for diffusion density estimation and inference-time control},
  author={He, Jiajun and Hern{\'a}ndez-Lobato, Jos{\'e} Miguel and Du, Yuanqi and Vargas, Francisco},
  journal={arXiv preprint arXiv:2506.05668},
  year={2025}
}

@inproceedings{ohdawin,
  title={DaWin: Training-free Dynamic Weight Interpolation for Robust Adaptation},
  author={Oh, Changdae and Li, Yixuan and Song, Kyungwoo and Yun, Sangdoo and Han, Dongyoon},
  booktitle={The Thirteenth International Conference on Learning Representations},
  year={2025}
}

@inproceedings{wangensembling,
  title={Ensembling Diffusion Models via Adaptive Feature Aggregation},
  author={Wang, Cong and Tian, Kuan and Guan, Yonghang and Shen, Fei and Jiang, Zhiwei and Gu, Qing and Zhang, Jun},
  booktitle={The Thirteenth International Conference on Learning Representations},
  year={2025}
}

@inproceedings{hessel2021clipscore,
  title={CLIPScore: A Reference-free Evaluation Metric for Image Captioning},
  author={Hessel, Jack and Holtzman, Ari and Forbes, Maxwell and Le Bras, Ronan and Choi, Yejin},
  booktitle={Proceedings of the 2021 Conference on Empirical Methods in Natural Language Processing},
  pages={7514--7528},
  year={2021}
}

@inproceedings{kim2021lipschitz,
  title={The lipschitz constant of self-attention},
  author={Kim, Hyunjik and Papamakarios, George and Mnih, Andriy},
  booktitle={International Conference on Machine Learning},
  pages={5562--5571},
  year={2021},
  organization={PMLR}
}

@inproceedings{castin2024smooth,
  title={How smooth is attention?},
  author={Castin, Val{\'e}rie and Ablin, Pierre and Peyr{\'e}, Gabriel},
  booktitle={Proceedings of the 41st International Conference on Machine Learning},
  pages={5817--5840},
  year={2024}
}

@inproceedings{qilipsformer,
  title={LipsFormer: Introducing Lipschitz Continuity to Vision Transformers},
  author={Qi, Xianbiao and Wang, Jianan and Chen, Yihao and Shi, Yukai and Zhang, Lei},
  booktitle={The Eleventh International Conference on Learning Representations},
  year={2023}
}

@book{vershynin2018high,
  title={High-dimensional probability: An introduction with applications in data science},
  author={Vershynin, Roman},
  volume={47},
  year={2018},
  publisher={Cambridge university press}
}

@article{laurent2000adaptive,
  title={Adaptive estimation of a quadratic functional by model selection},
  author={Laurent, Beatrice and Massart, Pascal},
  journal={Annals of statistics},
  pages={1302--1338},
  year={2000},
  publisher={JSTOR}
}
\bibliographystyle{icml2026}

%%%%%%%%%%%%%%%%%%%%%%%%%%%%%%%%%%%%%%%%%%%%%%%%%%%%%%%%%%%%%%%%%%%%%%%%%%%%%%%
%%%%%%%%%%%%%%%%%%%%%%%%%%%%%%%%%%%%%%%%%%%%%%%%%%%%%%%%%%%%%%%%%%%%%%%%%%%%%%%
% APPENDIX
%%%%%%%%%%%%%%%%%%%%%%%%%%%%%%%%%%%%%%%%%%%%%%%%%%%%%%%%%%%%%%%%%%%%%%%%%%%%%%%
%%%%%%%%%%%%%%%%%%%%%%%%%%%%%%%%%%%%%%%%%%%%%%%%%%%%%%%%%%%%%%%%%%%%%%%%%%%%%%%
\newpage
\appendix
\onecolumn
\section{Concentration of Measure Theorem for High-Dimensional Independent Normal Distribution}\label{concentration}
\begin{lemma}\label{lemma1}
Let \( X = (X_1, X_2, \ldots, X_n) \) be an \( n \)-dimensional random vector where each component \( X_i \) is independently and identically distributed as \( \mathcal{N}(0, \sigma^2 ) \). Then, as the dimension \( n \) increases, the norm \( \|X\| \) of \( X \) is concentrated around \( \sqrt{n} \sigma \). Specifically, for any small \( \epsilon \geq 0 \) and sufficiently large \( n \),
\begin{equation}
P\left( \left| \|X\| - \sqrt{n} \sigma \right| \leq \epsilon \sqrt{n} \sigma \right) \geq 1 - 2 e^{ - \frac{n\epsilon^2}{1 + 2\epsilon}}.
\end{equation}
This means \( X \) will concentrate near the hypersphere of radius \( \sqrt{n}\sigma \) with high probability.
\end{lemma}
\textit{Proof}.
First, we consider the squared norm of a random vector \( X:  \| X \|^2 = X_1^2 + X_2^2 + \ldots + X_n^2\).
Since \( X_i \sim \mathcal{N}(0, \sigma^2 ) \), it follows that \( Y = \| X \|^2 / \sigma^2 \sim \chi^2(n) \).

Using the Chernoff bound of the chi-square distribution:
\begin{equation}
P(Y\geq(1 + \delta)n)\leq e^{-\frac{n}{2}\left[\delta-\ln(1 + \delta)\right]},
\end{equation}
\begin{equation}
P(Y\leq(1 - \delta)n)\leq e^{-\frac{n}{2}\left[-\delta-\ln(1 - \delta)\right]},
\end{equation}
where $\delta\geq 0$ is a small value. According to $\ln(1+\delta)\leq \frac{(2+\delta)\delta}{2(1+\delta)}$ and $\ln(1-\delta)\leq -\delta - \frac{\delta^2}{2} $, we can deduce that:
\begin{equation}
\begin{aligned}
P(\| X \|^2\geq(1 + \delta)n\sigma^2)&=P(Y\geq(1 + \delta)n)\\
&\leq e^ {- \frac{n\delta^2}{4(1 + \delta)}},
\end{aligned}
\end{equation}
\begin{equation}
\begin{aligned}
P(\| X \|^2\leq(1 - \delta)n\sigma^2)&=P(Y\leq(1 - \delta)n)\\
&\leq e^ {- \frac{n\delta^2}{4}}\\
&\leq e^ {- \frac{n\delta^2}{4(1 + \delta)}}.
\end{aligned}
\end{equation}
Given the $\sqrt{1+\delta}\leq1+\frac{\delta}{2}$, we have:
\begin{equation}
\begin{aligned}
P(\| X \|\geq (1 + \frac{\delta}{2})\sqrt{n}\sigma)&\leq P(\| X \|\geq \sqrt{(1 + \delta)}\sqrt{n}\sigma)\\
&\leq e^ {- \frac{n\delta^2}{4(1 + \delta)}},
\end{aligned}
\end{equation}
which indicates that:
\begin{equation}\label{shangjie}
\begin{aligned}
P(\| X \|\geq (1 + \delta)\sqrt{n}\sigma)&\leq e^ {- \frac{4n\delta^2}{4(1 + 2\delta)}}\\
&\leq e^ {- \frac{n\delta^2}{4(1 + \delta)}},
\end{aligned}
\end{equation}
Also, given the $\sqrt{1-\delta} \geq 1 - \delta$, we have:
\begin{equation}\label{xiajie}
\begin{aligned}
P(\| X \|\leq (1 - \delta)\sqrt{n}\sigma)&\leq P(\| X \|\leq \sqrt{(1 - \delta)}\sqrt{n}\sigma)\\
&\leq e^ {- \frac{n\delta^2}{4(1 + \delta)}}.
\end{aligned}
\end{equation}
Let $\epsilon=\delta/2$, and then organize equations (\ref{shangjie}) and (\ref{xiajie}):
\begin{equation}
P\left( \left| \|X\| - \sqrt{n} \sigma \right| \leq \epsilon \sqrt{n} \sigma \right) \geq 1 - 2 e^{ - \frac{n\epsilon^2}{1 + 2\epsilon}},
\end{equation}
which concludes the proof.
\section{Proof of Proposition \ref{p_normupper}}\label{proof_normupper}

\noindent \textbf{Proposition \ref{p_normupper}.}
Let \( \mathbf z'_t \) denote the aggregated variable at time step \( t \). For the sampling step from \( t \) to \( t-1 \), two diffusion models \(p_{\theta_1}\) and \(p_{\theta_2}\) sample \( \mathbf{z}^{\theta_1}_{t-1} \) and \( \mathbf{z}^{\theta_2}_{t-1} \) respectively from (\ref{cfg}). Then, with probability at least $1-\gamma$, the absolute difference in norms and the angles are bounded by:
\begin{equation}
\Big|\|\mathbf z^{\theta_1}_{t-1}\|-\|\mathbf z^{\theta_2}_{t-1}\|\Big|\leq L_t+L_{\theta_2}L_\tau+2\sigma_t \sqrt{2 \ln \frac{4}{\gamma}},
\tag{\ref{e_normupper}}
\end{equation}
\begin{equation}
\begin{aligned}
\cos \varphi &\geq 1-\frac{\delta^2}{\|\mathbf z^{\theta_1}_{t-1}\|^2+\|\mathbf z^{\theta_2}_{t-1}\|^2},\\
\delta&\leq L_t+L_{\theta_2}L_\tau + \sigma_t \sqrt{2 \left( n + 2\sqrt{n \ln \frac{1}{\gamma}} + 2\ln \frac{1}{\gamma} \right)}.
\end{aligned}
\tag{\ref{cosin}}
\end{equation}
where $\varphi$ is the angle between $\mathbf z^{\theta_1}_{t-1}$ and $\mathbf z^{\theta_2}_{t-1}$, and $L_{\theta_2}$ is the Lipschitz constant.

\textit{Proof}.

According to \eqref{cfg}, we have:
\begin{equation}
\begin{aligned}
    \mathbf{z}^{\theta_1}_{t-1}=\bm\mu_{\theta_1}(\mathbf{z}'_{t}, t, y_1)+ \sigma_t \boldsymbol{\epsilon}_1,\\
    \mathbf{z}^{\theta_2}_{t-1}=\bm\mu_{\theta_2}(\mathbf{z}'_{t}, t, y_2) + \sigma_t \boldsymbol{\epsilon}_2.
\end{aligned}
\end{equation}

Then, the difference between the norms is:
\begin{equation}\label{delta}
    \begin{aligned}
        \Big|\|\mathbf z^{\theta_1}_{t-1}\|-\|\mathbf z^{\theta_2}_{t-1}\|\Big|&=\Big| \|\bm\mu_{\theta_1}(\mathbf{z}'_{t}, t, y_1)+ \sigma_t \boldsymbol{\epsilon}_1\| - \|\bm\mu_{\theta_2}(\mathbf{z}'_{t}, t, y_2) + \sigma_t \boldsymbol{\epsilon}_2\| \Big|\\
        &=\Big| \|\bm\mu_{\theta_1}(\mathbf{z}'_{t}, t, y_1)+ \sigma_t \boldsymbol{\epsilon}_1\| - \|\bm\mu_{\theta_2}(\mathbf{z}'_{t}, t, y_2) + \sigma_t \boldsymbol{\epsilon}_1\| \Big.\\
        &\quad+\Big. \|\bm\mu_{\theta_2}(\mathbf{z}'_{t}, t, y_2) + \sigma_t \boldsymbol{\epsilon}_1\| - \|\bm\mu_{\theta_2}(\mathbf{z}'_{t}, t, y_2) + \sigma_t \boldsymbol{\epsilon}_2\| \Big|\\
        &\leq \Big| \|\bm\mu_{\theta_1}(\mathbf{z}'_{t}, t, y_1)+ \sigma_t \boldsymbol{\epsilon}_1\| - \|\bm\mu_{\theta_2}(\mathbf{z}'_{t}, t, y_2) + \sigma_t \boldsymbol{\epsilon}_1\|\Big| 
        \\ &\quad+\Big| \|\bm\mu_{\theta_2}(\mathbf{z}'_{t}, t, y_2) + \sigma_t \boldsymbol{\epsilon}_1\| - \|\bm\mu_{\theta_2}(\mathbf{z}'_{t}, t, y_2) + \sigma_t \boldsymbol{\epsilon}_2\| \Big|\\
        &\leq\|\bm\mu_{\theta_1}(\mathbf{z}'_{t}, t, y_1)-\bm\mu_{\theta_2}(\mathbf{z}'_{t}, t, y_2)\| 
        +\Big| \|\bm\mu_{\theta_2}(\mathbf{z}'_{t}, t, y_2) + \sigma_t \boldsymbol{\epsilon}_1\| - \|\bm\mu_{\theta_2}(\mathbf{z}'_{t}, t, y_2) + \sigma_t \boldsymbol{\epsilon}_2\| \Big|.\\
    \end{aligned}
\end{equation}

For the first term of (\ref{delta}):
\begin{equation}
    \begin{aligned}
        &\|\bm\mu_{\theta_1}(\mathbf{z}'_{t}, t, y_1)-\bm\mu_{\theta_2}(\mathbf{z}'_{t}, t, y_2)\|\\
        =&\|\bm\mu_{\theta_1}(\mathbf{z}'_{t}, t, y_1)- \bm\mu_{\theta_2}(\mathbf{z}'_{t}, t, y_1)+\bm\mu_{\theta_2}(\mathbf{z}'_{t}, t, y_1)-\bm\mu_{\theta_2}(\mathbf{z}'_{t}, t, y_2)\|\\
        \leq &\|\bm\mu_{\theta_1}(\mathbf{z}'_{t}, t, y_1)- \bm\mu_{\theta_2}(\mathbf{z}'_{t}, t, y_1)\|+\|\bm\mu_{\theta_2}(\mathbf{z}'_{t}, t, y_1)-\bm\mu_{\theta_2}(\mathbf{z}'_{t}, t, y_2)\|\\
        \leq& L_t+\|\bm\mu_{\theta_2}(\mathbf{z}'_{t}, t, y_1)-\bm\mu_{\theta_2}(\mathbf{z}'_{t}, t, y_2)\|.
    \end{aligned}
\end{equation}

Due to the presence of dot-product self-attention and layer normalization, the U-Net architecture with Transformer modules in diffusion models cannot satisfy global Lipschitz continuity \citep{kim2021lipschitz, castin2024smooth, qilipsformer}. However, note that in $\|\bm\mu_{\theta_2}(\mathbf{z}'_{t}, t, y_1)-\bm\mu_{\theta_2}(\mathbf{z}'_{t}, t, y_2)\|$, the conditions $y_1$ and $y_2$ belong to the same fine-grained task setting. Therefore, within this setting, the U-Net architecture admits a local Lipschitz bound $L_{\theta_2}$ with respect to the conditioning variable. Hence:
\begin{equation}
    \begin{aligned}
        &\|\bm\mu_{\theta_1}(\mathbf{z}'_{t}, t, y_1)-\bm\mu_{\theta_2}(\mathbf{z}'_{t}, t, y_2)\|\\
        \leq &L_t+\|\bm\mu_{\theta_2}(\mathbf{z}'_{t}, t, y_1)-\bm\mu_{\theta_2}(\mathbf{z}'_{t}, t, y_2)\|\\
        \leq& L_t + L_{\theta_2}\|y_1-y_2\|\\
        \leq& L_t+L_{\theta_2}L_\tau
    \end{aligned}
\end{equation}

For the second term in \eqref{delta}, it intuitively measures the distance between two samples generated by $p_{\theta_2}$, which admits a controllable upper bound. More formally, define the norm function $f: \mathbb{R}^n \to \mathbb{R}$ with respect to noise $\boldsymbol{\epsilon}$:
\begin{equation}
f(\boldsymbol{\epsilon}) = \|\boldsymbol{\mu}_{\theta_2} + \sigma_t \boldsymbol{\epsilon}\|,
\end{equation}
the Lipschitz constant derivation:
\begin{equation}
|f(\mathbf{x}) - f(\mathbf{y})| = \left| \|\boldsymbol{\mu} + \sigma_t \mathbf{x}\| - \|\boldsymbol{\mu} + \sigma_t \mathbf{y}\| \right| \le \| (\boldsymbol{\mu} + \sigma_t \mathbf{x}) - (\boldsymbol{\mu} + \sigma_t \mathbf{y}) \| = \sigma_t \|\mathbf{x} - \mathbf{y}\|.
\end{equation}
Thus, $f(\cdot)$ is $\sigma_t$-Lipschitz.
For $\boldsymbol{\epsilon} \sim \mathcal{N}(\mathbf{0}, \mathbf{I})$, applying the Gaussian Concentration Inequality for Lipschitz functions (Theorem 5.1.3 in \cite{vershynin2018high}):
\begin{equation}
P\left( |f(\boldsymbol{\epsilon}) - \mathbb{E}[f]| \ge u \right) \le 2 \exp\left( -\frac{u^2}{2\sigma_t^2} \right)
\end{equation}
Let $\gamma = |f(\boldsymbol{\epsilon}_1) - f(\boldsymbol{\epsilon}_2)|$, then:
\begin{equation}
\gamma \le |f(\boldsymbol{\epsilon}_1) - \mathbb{E}[f]| + |f(\boldsymbol{\epsilon}_2) - \mathbb{E}[f]|.
\end{equation}

To bound $\gamma < 2u$ with probability $1-\gamma$, we bound the failure probability $P(\gamma \ge 2u)$:
\begin{equation}
\begin{aligned}
P(\gamma \ge 2u) &\le P(|f(\boldsymbol{\epsilon}_1) - \mathbb{E}[f]| \ge u) + P(|f(\boldsymbol{\epsilon}_2) - \mathbb{E}[f]| \ge u) \\
&\le 2 \exp\left( -\frac{u^2}{2\sigma_t^2} \right) + 2 \exp\left( -\frac{u^2}{2\sigma_t^2} \right) \\
&= 4 \exp\left( -\frac{u^2}{2\sigma_t^2} \right).
\end{aligned}
\end{equation}
Set the failure probability to $\gamma$ and solve for the threshold $2u$:
\begin{equation}
\gamma = 4 \exp\left( -\frac{u^2}{2\sigma_t^2} \right) \implies \ln\frac{\gamma}{4} = -\frac{u^2}{2\sigma_t^2} \implies u = \sigma_t \sqrt{2 \ln \frac{4}{\gamma}}.
\end{equation}
Substituting $u$ back into the bound $2u$:
\begin{equation}\label{diff_sample}
\Big| \|\bm\mu_{\theta_2}(\mathbf{z}'_{t}, t, y_2) + \sigma_t \boldsymbol{\epsilon}_1\| - \|\bm\mu_{\theta_2}(\mathbf{z}'_{t}, t, y_2) + \sigma_t \boldsymbol{\epsilon}_2\| \Big| \le 2\sigma_t \sqrt{2 \ln \frac{4}{\gamma}} \quad (\text{w.p. } \ge 1 - \gamma).
\end{equation}
Therefore,
\begin{equation}
\Big|\|\mathbf z^{\theta_1}_{t-1}\|-\|\mathbf z^{\theta_2}_{t-1}\|\Big|\leq L_t+L_{\theta_2}L_\tau+2\sigma_t \sqrt{2 \ln \frac{4}{\gamma}} \quad (\text{w.p. } \ge 1 - \gamma).
\end{equation}

Denote $\varphi$ is the angle between $\mathbf z^{\theta_1}_{t-1}$ and $\mathbf z^{\theta_2}_{t-1}$, then
\begin{equation}
\begin{aligned}
    \cos \varphi &= \frac{\|\mathbf z^{\theta_1}_{t-1}\|^2+\|\mathbf z^{\theta_2}_{t-1}\|^2-\|\mathbf z^{\theta_1}_{t-1}-\mathbf z^{\theta_2}_{t-1}\|^2}{2\|\mathbf z^{\theta_1}_{t-1}\|\|\mathbf z^{\theta_2}_{t-1}\|}\\
    &\geq  \frac{\|\mathbf z^{\theta_1}_{t-1}\|^2+\|\mathbf z^{\theta_2}_{t-1}\|^2-\|\mathbf z^{\theta_1}_{t-1}-\mathbf z^{\theta_2}_{t-1}\|^2}{\|\mathbf z^{\theta_1}_{t-1}\|^2+\|\mathbf z^{\theta_2}_{t-1}\|^2}\\
    & = 1-\frac{\|\mathbf z^{\theta_1}_{t-1}-\mathbf z^{\theta_2}_{t-1}\|^2}{\|\mathbf z^{\theta_1}_{t-1}\|^2+\|\mathbf z^{\theta_2}_{t-1}\|^2},
\end{aligned}
\end{equation}
let $\delta=\|\mathbf z^{\theta_1}_{t-1}-\mathbf z^{\theta_2}_{t-1}\|$, then:
\begin{equation}
\begin{aligned}
\delta&=\|\bm\mu_{\theta_1}(\mathbf{z}'_{t}, t, y_1)-\bm\mu_{\theta_2}(\mathbf{z}'_{t}, t, y_2)+\sigma_t(\boldsymbol{\epsilon}_1-\boldsymbol{\epsilon}_2)\|\\
&\leq L_t+L_{\theta_2}L_\tau + \|\sigma_t(\boldsymbol{\epsilon}_1-\boldsymbol{\epsilon}_2)\|.
\end{aligned}
\end{equation}
To analyze the term $\|\sigma_t(\boldsymbol{\epsilon}_1-\boldsymbol{\epsilon}_2)\|$, we first define $Z = \left\| \frac{\Delta \boldsymbol{\epsilon}}{\sqrt{2}} \right\|^2 = \frac{1}{2} \| \boldsymbol{\epsilon}_1 - \boldsymbol{\epsilon}_2 \|^2$, which follows a Chi-squared distribution $\chi^2(n)$. Then, using the Laurent-Massart Bound for $\chi^2(n)$ \cite{laurent2000adaptive}, for any $u > 0$:
\begin{equation}
P\left( Z \ge n + 2\sqrt{du} + 2u \right) \le \exp(-u).
\end{equation}
Setting $u = \ln(1/\gamma)$, we have with probability at least $1-\gamma$:
\begin{equation}
\frac{1}{2\sigma_t^2} \| \sigma_t(\boldsymbol{\epsilon}_1 - \boldsymbol{\epsilon}_2) \|^2 = Z \le n + 2\sqrt{n \ln \frac{1}{\gamma}} + 2\ln \frac{1}{\gamma}.
\end{equation}
Thus:
\begin{equation}\label{diff_noise}
\| \sigma_t(\boldsymbol{\epsilon}_1 - \boldsymbol{\epsilon}_2) \| \le \sigma_t \sqrt{2 \left( n + 2\sqrt{n \ln \frac{1}{\gamma}} + 2\ln \frac{1}{\gamma} \right)}.
\end{equation}
Therefore:
\begin{equation}
\begin{aligned}
    \cos \varphi &\geq 1-\frac{\|\mathbf z^{\theta_1}_{t-1}-\mathbf z^{\theta_2}_{t-1}\|^2}{\|\mathbf z^{\theta_1}_{t-1}\|^2+\|\mathbf z^{\theta_2}_{t-1}\|^2}\\
    & \geq 1-\frac{\delta^2}{\|\mathbf z^{\theta_1}_{t-1}\|^2+\|\mathbf z^{\theta_2}_{t-1}\|^2},\\
\end{aligned}
\end{equation}
where $\delta\leq L_t+L_{\theta_2}L_\tau + \sigma_t \sqrt{2 \left( n + 2\sqrt{n \ln \frac{1}{\gamma}} + 2\ln \frac{1}{\gamma} \right)} \quad (\text{w.p. } \ge 1 - \gamma)$.

which concludes the proof.
\section{Proof of Proposition \ref{p1}}\label{proof_p1}
\noindent \textbf{Proposition \ref{p1}.}
For the diffusion model \( p_{\theta_1} \) defined by (\ref{cfg}) and any new intermediate variable \(\mathbf{z}'_{t-1}\) from (\ref{agg_x}), let:
    \begin{equation}\tag{\ref{op}}
        \mathbf{\tilde z}_{t-1} = \mathbf{z}'_{t-1} - \eta^{\theta_1}_{t-1} \frac{\mathbf{z}'_{t-1} - \mu_{\theta_1}(\mathbf{z}^{\theta_1}_t, t, y_1)}{\|\mathbf{z}'_{t-1} - \mu_{\theta_1}(\mathbf{z}^{\theta_1}_t, t, y_1)\|},
    \end{equation}
where $\eta^{\theta_1}_{t-1}$ is a small optimization step size. There exists $\eta^{\theta_1}_{t-1}$ such that $\mathbf{\tilde z}_{t-1}\in D^{\theta_1}_{t-1,y_1}$. Moreover, an approximate lower bound on the probability that $\mathbf{\tilde z}_{t-1}\in D^{\theta_2}_{t-1,y_2}$ is given by:
\begin{equation}\tag{\ref{lowbound}}
P\left(\mathbf{\tilde z}_{t-1}\in D^{\theta_2}_{t-1,y_2}\right)\geq 1 - 2 \exp\left(
   -\frac{n\left(\epsilon_{t-1}^{\theta_2} - \tfrac{d}{\sigma_t\sqrt{n}}\right)^{2}}
          {1 + 2\left(\epsilon_{t-1}^{\theta_2} - \tfrac{d}{\sigma_t\sqrt{n}}\right)}
\right),
\end{equation}
where $d = \phi_w(\varphi) \delta + \eta^{\theta_1}_{t-1}$ and $\phi_w(\varphi)
= \sin\!\bigl((1-w)\varphi/2\bigr)/\sin(\varphi/2)$.

\textit{Proof}.

(1) The existence of $\eta^{\theta_1}_{t-1}$ 
such that $\mathbf{\tilde z}_{t-1}\in D^{\theta_1}_{t-1,y_1}$:

We begin by defining the spherical shell as the high-probability set of $p_{\theta_i}(\mathbf{z}^{\theta_i}_{t-1} \mid \mathbf{z}^{\theta_i}_t, y_i)$:
\begin{equation}
    \mathcal A^{\theta_i}_{t-1,y_i}(\epsilon_{t-1}^{\theta_i})=\left\{ \left| \|\mathbf z^{\theta_i}_{t-1}-\mu_{\theta_i}(\mathbf{z}^{\theta_i}_t, t, y_i)\| - \sqrt{n} \sigma_t \right| \leq \epsilon_{t-1}^{\theta_i} \sqrt{n} \sigma_t \right\},
\end{equation}
where $\epsilon_{t-1}^{\theta_i}=\sup \left\{ \epsilon \geq 0 : 
\mathcal{A}^{\theta_i}_{t-1,y_i}(\epsilon) \subseteq 
D^{\theta_i}_{t-1,y_i}(\tau_{t-1}^{\theta_i}) \right\}$.

Note that:
\begin{equation}
    \begin{aligned}
    \nabla_{\mathbf{z}'_{t-1}} p_{\theta_1}(\mathbf{z}'_{t-1} | \mathbf{z}^{\theta_1}_{t}, y_1)
    &=\frac{1}{(2\pi\sigma_t^2)^{n/2}} \nabla_{\mathbf{z}'_{t-1}} \left( e^{ -\frac{\|\mathbf{z}'_{t-1} - \mu_\theta(\mathbf{z}^{\theta_1}_{t}, t, y_1)\|^2}{2\sigma_t^2}} \right)\\
    &=-\frac{1}{(2\pi\sigma_t^2)^{n/2}} e^{-\frac{\|\mathbf{z}'_{t-1} - \mu_\theta(\mathbf{z}^{\theta_1}_{t}, t, y_1)\|^2}{2\sigma_t^2}} \left( \frac{\mathbf{z}'_{t-1} - \mu_\theta(\mathbf{z}^{\theta_1}_{t}, t, y_1)}{\sigma_t^2} \right).
    \end{aligned}
\end{equation}

Therefore,
\begin{equation}
\begin{aligned}
    \mathbf{\tilde z}_{t-1} &= \mathbf{z}'_{t-1} - \eta_{t-1}^{\theta_1} \frac{\mathbf{z}'_{t-1} - \mu_\theta(\mathbf{z}^{\theta_1}_{t}, t, y_1)}{\|\mathbf{z}'_{t-1} - \mu_\theta(\mathbf{z}^{\theta_1}_{t}, t, y_1)\|}\\
    &= \mathbf{z}'_{t-1} + \eta_{t-1}^{\theta_1} \frac{\nabla_{\mathbf{z}'_{t-1}} p_{\theta_1}(\mathbf{z}'_{t-1} | \mathbf{z}^{\theta_1}_{t},y_1)}{\|\nabla_{\mathbf{z'}_{t-1}} p_{\theta_1}(\mathbf{z}'_{t-1} | \mathbf{z}^{\theta_1}_{t}, y_1)\|}.
\end{aligned}
\end{equation}

This indicates that $\mathbf{\tilde z}_{t-1}$ is obtained from $\mathbf{z}'_{t-1}$ by moving $\eta_{t-1}^{\theta_1}$ units in the radial 
direction opposite to $p_{\theta_1}(\mathbf{z}'_{t-1} \mid \mathbf{z}^{\theta_1}_{t}, y_1)$. 
Then, it suffices to ensure that:
\begin{equation}
\left|\, \|\mathbf{z}'_{t-1}-\mu_\theta(\mathbf{z}^{\theta_1}_{t}, t, y_1)\| 
- \eta_{t-1}^{\theta_1} - \sqrt{n}\,\sigma_t \,\right| 
< \epsilon_{t-1}^{\theta_1}\sqrt{n}\,\sigma_t,
\end{equation}

from which we obtain:
\begin{equation}
\begin{aligned}
&||z'_{t-1} - \mu_{\theta}(z_t^{\theta_1}, t, y_1)|| - \sqrt{n}\sigma_t - \epsilon_{t-1}^{\theta_1}\sqrt{n}\sigma_t \\&< \eta_{t-1}^{\theta_1} <\\& ||z'_{t-1} - \mu_{\theta}(z_t^{\theta_1}, t, y_1)|| - \sqrt{n}\sigma_t + \epsilon_{t-1}^{\theta_1}\sqrt{n}\sigma_t,
\end{aligned}
\end{equation}
thus, we have $\mathbf{\tilde z}_{t-1}\in \mathcal{A}^{\theta_1}_{t-1,y_1} \subseteq D^{\theta_1}_{t-1,y_1}$, which completes the proof of existence.

(2) The approximate lower bound on the probability that $\mathbf{\tilde z}_{t-1}\in D^{\theta_2}_{t-1,y_2}$:

By the triangle inequality, for any $z^{\theta_2}_{t-1} \in \mathcal A^{\theta_2}_{t-1,y_2}$, we have:
\begin{equation}\label{triin}
\left\lVert \tilde{\mathbf z}_{t-1}-\mathbf z^{\theta_2}_{t-1}\right\rVert
\le
\left\lVert \mathbf z'_{t-1}-\mathbf z^{\theta_2}_{t-1}\right\rVert
+\eta^{\theta_1}_{t-1}. 
\end{equation}

From proposition \ref{p_normupper}, we assume that $\left\lVert \mathbf z^{\theta_1}_{t-1}\right\rVert \approx \left\lVert \mathbf z^{\theta_2}_{t-1}\right\rVert \approx r_{t-1}$, then, the chord lengths satisfy:
\begin{equation}
\left\lVert \mathbf z'_{t-1}-\mathbf z^{\theta_2}_{t-1}\right\rVert
= 2 r_{t-1} \sin\Bigl(\tfrac{(1-w)\varphi}{2}\Bigr),
\qquad
\delta = \left\lVert \mathbf z^{\theta_1}_{t-1}-\mathbf z^{\theta_2}_{t-1}\right\rVert
= 2 r_{t-1}\sin\Bigl(\tfrac{\varphi}{2}\Bigr).
\end{equation}

Dividing the two identities yields:
\begin{equation}\label{deltabound}
\left\lVert \mathbf z'_{t-1}-\mathbf z^{\theta_2}_{t-1}\right\rVert
= \phi_w(\varphi)\,\delta,
\quad
\phi_w(\varphi)=
\frac{\sin\bigl(\tfrac{(1-w)\varphi}{2}\bigr)}{\sin\bigl(\tfrac{\varphi}{2}\bigr)}.
\end{equation}

Combining (\ref{triin}) and (\ref{deltabound}), denote $d = \phi_w(\varphi)\,\delta +\eta^{\theta_1}_{t-1}$ and We can obtain the bound:
\begin{equation}
\operatorname{dist}\bigl(\tilde{\mathbf z}_{t-1},\,\mathcal A^{\theta_2}_{t-1,y_2}\bigr)
\le d.
\end{equation}

So, we can also refer to $d$ as the maximum distance to $p_{\theta_2}$. This implies that, if:
\begin{equation}
\bigl|\, \| \mathbf z^{\theta_2}_{t-1} - \mu_{\theta_2} \| - \sigma_t\sqrt{n} \,\bigr|
\le \bigl(\epsilon_{t-1}^{\theta_2} - d/\sigma_t\sqrt{n}\bigr) \sigma_t\sqrt{n},
\end{equation}
then $B \left(\mathbf z^{\theta_2}_{t-1},d\right) \subseteq
\mathcal{A}^{\theta_2}_{t-1,y_2}$. Therefore:
\begin{equation}
\begin{aligned}
P\left(\mathbf{\tilde z}_{t-1}\in D^{\theta_2}_{t-1,y_2}\right)
&\ge P\left(\mathbf{\tilde z}_{t-1}\in \mathcal A^{\theta_2}_{t-1,y_2}\right)\\
&\ge P\left(
\bigl|\, \| \mathbf z^{\theta_2}_{t-1} - \mu_{\theta_2} \| - \sigma_t\sqrt{n} \,\bigr|
\le \bigl(\epsilon_{t-1}^{\theta_2} - \tfrac{d}{\sigma_t\sqrt{n}}\bigr) \sigma_t\sqrt{n}
\right)\\
&\ge 1 - 2 \exp\left(
   -\frac{n\left(\epsilon_{t-1}^{\theta_2} - \tfrac{d}{\sigma_t\sqrt{n}}\right)^{2}}
          {1 + 2\left(\epsilon_{t-1}^{\theta_2} - \tfrac{d}{\sigma_t\sqrt{n}}\right)}
\right).
\end{aligned}
\end{equation}

The first line follows from 
$\mathbf{\tilde z}_{t-1} \in \mathcal{A}^{\theta_2}_{t-1,y_2} \subseteq D^{\theta_2}_{t-1,y_2}$, 
and the third line follows from Lemma \ref{lemma1}, which concludes the proof.
\section{Geometric Properties of AMDM}\label{geo_amdm}
The following proposition shows that the points obtained by applying the deviation optimization in \eqref{op} within the AMDM algorithm also lie on a local spherical manifold; therefore, spherical interpolation remains applicable.

\begin{proposition}\label{amdm_prop}
Let $\mathbf{z}'_t$ denote the aggregated variable at time step $t$. For the AMDM step from $t$ to $t-1$, the deviation optimization is first performed via \eqref{op} using diffusion models $p_{\theta_1}$ and $p_{\theta_2}$, followed by sampling according to \eqref{cfg} to obtain $\mathbf{z}^{\theta_1}_{t-1}$ and $\mathbf{z}^{\theta_2}_{t-1}$, respectively. Then, with probability at least $1-\gamma$, the absolute difference in norms and the angles are bounded by:
\begin{equation}
\Big|||\mathbf z_{t-1}^{\theta_1}|| - ||\mathbf z_{t-1}^{\theta_2}||\Big| \le L_t + L_{\theta_2}L_{\tau} + L_{\mathbf z}^{\theta_1} \eta_t^{\theta_1} + L_{\mathbf z}^{\theta_2} \eta_t^{\theta_2} + 2\sigma_t \sqrt{2 \ln \frac{4}{\gamma}}.
\end{equation}
\begin{equation}
\begin{gathered}
\cos \varphi \ge 1 - \frac{\delta^2}{||\mathbf z_{t-1}^{\theta_1}||^2 + ||\mathbf z_{t-1}^{\theta_2}||^2},\\
\delta \le  L_t+L_{\theta_2}L_\tau + L_{\mathbf z}^{\theta_1} \eta_t^{\theta_1} +L_{\mathbf z}^{\theta_2} \eta_t^{\theta_2}+ \sigma_t \sqrt{2 \left( n + 2\sqrt{n \ln \frac{1}{\gamma}} + 2\ln \frac{1}{\gamma} \right)}.
\end{gathered}
\end{equation}
\end{proposition}

\textit{Proof}.

According to the definition of AMDM, deviation optimization is introduced in the step from $t$ to $t-1$. Based on \eqref{cfg} and \eqref{op}, the sampled variable $\mathbf{z}_{t-1}^{\theta_i}$ after deviation optimization can be expressed as:
\begin{equation}
\begin{gathered}
\mathbf{z}_{t-1}^{\theta_i} = \bm\mu_{\theta_i}(\tilde{\mathbf{z}}_t^{\theta_i}, t, y_i) + \sigma_t \boldsymbol{\epsilon}_i,\quad \tilde{\mathbf{z}}_t^{\theta_i} = \mathbf{z}_t^{\prime} + \mathbf{d}_i,\quad \mathbf{d}_i=- \eta^{\theta_i}_{t} \frac{\mathbf{z}'_{t} - \bm\mu_{\theta_i}(\mathbf{z}^{\theta_i}_{t+1}, t+1, y_i)}{\|\mathbf{z}'_{t} - \bm\mu_{\theta_i}(\mathbf{z}^{\theta_i}_{t+1}, t+1, y_i)\|}.
\end{gathered}
\end{equation}
Then, the absolute difference in norms is:
\begin{equation}\label{new_dnorm}
\begin{aligned}
\Big|\|\mathbf z_{t-1}^{\theta_1}\| - \|\mathbf z_{t-1}^{\theta_2}\|\Big|  &= \Big|||\bm\mu_{\theta_1}(\tilde{\mathbf{z}}_t^{\theta_1}, t, y_1) + \sigma_t \boldsymbol{\epsilon}_1|| - ||\bm\mu_{\theta_2}(\tilde{\mathbf{z}}_t^{\theta_2}, t, y_2) + \sigma_t \boldsymbol{\epsilon}_2||\Big| \\
&\le ||\bm\mu_{\theta_1}(\tilde{\mathbf{z}}_t^{\theta_1}, t, y_1) - \bm\mu_{\theta_2}(\tilde{\mathbf{z}}_t^{\theta_2}, t, y_2)|| + \Big| ||\bm\mu_{\theta_2}(\tilde{\mathbf{z}}_t^{\theta_2}, t, y_2) + \sigma_t \boldsymbol{\epsilon}_1|| - ||\bm\mu_{\theta_2}(\tilde{\mathbf{z}}_t^{\theta_2}, t, y_2) + \sigma_t \boldsymbol{\epsilon}_2|| \Big|.
\end{aligned}
\end{equation}
For the first term of \eqref{new_dnorm}:
\begin{equation}\label{new_mu_mu}
\begin{aligned}
&||\bm\mu_{\theta_1}(\tilde{\mathbf{z}}_t^{\theta_1}, t, y_1) - \bm\mu_{\theta_2}(\tilde{\mathbf{z}}_t^{\theta_2}, t, y_2)||\\
=&||\bm\mu_{\theta_1}(\tilde{\mathbf{z}}_t^{\theta_1}, t, y_1) - \bm\mu_{\theta_1}(\mathbf{z}'_{t}, t, y_1) + \bm\mu_{\theta_1}(\mathbf{z}'_{t}, t, y_1) - \bm\mu_{\theta_2}(\mathbf{z}'_{t}, t, y_2) +\bm\mu_{\theta_2}(\mathbf{z}'_{t}, t, y_2) - \bm\mu_{\theta_2}(\tilde{\mathbf{z}}_t^{\theta_2}, t, y_2)||\\
\leq &||\bm\mu_{\theta_1}(\tilde{\mathbf{z}}_t^{\theta_1}, t, y_1) - \bm\mu_{\theta_1}(\mathbf{z}'_{t}, t, y_1) || + ||\bm\mu_{\theta_2}(\mathbf{z}'_{t}, t, y_2) - \bm\mu_{\theta_2}(\tilde{\mathbf{z}}_t^{\theta_2}, t, y_2)||+|| \bm\mu_{\theta_1}(\mathbf{z}'_{t}, t, y_1) - \bm\mu_{\theta_2}(\mathbf{z}'_{t}, t, y_2) ||\\
\leq & ||\bm\mu_{\theta_1}(\tilde{\mathbf{z}}_t^{\theta_1}, t, y_1) - \bm\mu_{\theta_1}(\mathbf{z}'_{t}, t, y_1) || + ||\bm\mu_{\theta_2}(\mathbf{z}'_{t}, t, y_2) - \bm\mu_{\theta_2}(\tilde{\mathbf{z}}_t^{\theta_2}, t, y_2)||+L_t+L_{\theta_2}L_\tau.
\end{aligned}
\end{equation}
Practically, the denoising network $\mu_{\theta_i}(\cdot)$ is composed of standard neural network operators
(e.g., Linear layers, LayerNorm, SiLU) with finite weights, and is differentiable almost everywhere.
Therefore, within a small neighborhood induced by the optimization step size $\eta^{\theta_i}_{t}$, we assume that the input Jacobian remains bounded. Hence, there exists a local constant $L_z^{\theta_i}$ such that $\bigl\|\bm\mu_{\theta_i}(\tilde {\mathbf z})-\bm\mu_{\theta_i}(\mathbf z)\bigr\|
\;\le\;L_{\mathbf z}^{\theta_i}\,\bigl\|\tilde {\mathbf z}-\mathbf z\bigr\|$. Then, the first term of \eqref{new_mu_mu} is:
\begin{equation}
\begin{aligned}
||\bm\mu_{\theta_1}(\tilde{\mathbf{z}}_t^{\theta_1}, t, y_1) - \bm\mu_{\theta_1}(\mathbf{z}'_{t}, t, y_1) ||&\leq L_{\mathbf z}^{\theta_1}\,\bigl\|\tilde{\mathbf{z}}_t^{\theta_1}-\mathbf{z}'_{t}\bigr\|\\
&=L_{\mathbf z}^{\theta_1} \|\mathbf d_1\|\\
&=L_{\mathbf z}^{\theta_1} \eta_t^{\theta_1}.
\end{aligned}
\end{equation}
Similarily, $||\bm\mu_{\theta_2}(\tilde{\mathbf{z}}_t^{\theta_2}, t, y_2) - \bm\mu_{\theta_2}(\mathbf{z}'_{t}, t, y_1) ||\leq L_{\mathbf z}^{\theta_2} \eta_t^{\theta_2}$.
Hence, \eqref{new_mu_mu} can be simplified:
\begin{equation}\label{re_new_mu_mu}
||\bm\mu_{\theta_1}(\tilde{\mathbf{z}}_t^{\theta_1}, t, y_1) - \bm\mu_{\theta_2}(\tilde{\mathbf{z}}_t^{\theta_2}, t, y_2)||
\leq L_{\mathbf z}^{\theta_1} \eta_t^{\theta_1} +L_{\mathbf z}^{\theta_2} \eta_t^{\theta_2}  +L_t+L_{\theta_2}L_\tau.
\end{equation}
For the second term of \eqref{new_dnorm}, by applying the result in (29), we obtain that:
\begin{equation}
\Big| ||\bm\mu_{\theta_2}(\tilde{\mathbf{z}}_t^{\theta_2}, t, y_2) + \sigma_t \boldsymbol{\epsilon}_1|| - ||\bm\mu_{\theta_2}(\tilde{\mathbf{z}}_t^{\theta_2}, t, y_2) + \sigma_t \boldsymbol{\epsilon}_2|| \Big| \le 2\sigma_t \sqrt{2 \ln \frac{4}{\gamma}} \quad (\text{w.p. } \ge 1 - \gamma).
\end{equation}
For $\delta$:
\begin{equation}
\begin{aligned}
\delta&=\|\mathbf z^{\theta_1}_{t-1}-\mathbf z^{\theta_2}_{t-1}\|\\
&=||\bm\mu_{\theta_1}(\tilde{\mathbf{z}}_t^{\theta_1}, t, y_1) - \bm\mu_{\theta_2}(\tilde{\mathbf{z}}_t^{\theta_2}, t, y_2)+\sigma_t(\boldsymbol{\epsilon}_1-\boldsymbol{\epsilon}_2)||\\
&\leq ||\bm\mu_{\theta_1}(\tilde{\mathbf{z}}_t^{\theta_1}, t, y_1) - \bm\mu_{\theta_2}(\tilde{\mathbf{z}}_t^{\theta_2}, t, y_2)|| + ||\sigma_t(\boldsymbol{\epsilon}_1-\boldsymbol{\epsilon}_2)||.
\end{aligned}
\end{equation}
Accordding to \eqref{re_new_mu_mu} and \eqref{diff_noise}, then:
\begin{equation}
\delta\leq L_{\mathbf z}^{\theta_1} \eta_t^{\theta_1} +L_{\mathbf z}^{\theta_2} \eta_t^{\theta_2}  +L_t+L_{\theta_2}L_\tau + \sigma_t \sqrt{2 \left( n + 2\sqrt{n \ln \frac{1}{\gamma}} + 2\ln \frac{1}{\gamma} \right)} \quad (\text{w.p. } \ge 1 - \gamma)
\end{equation}
which concludes the proof.

Proposition \ref{amdm_prop} shows that the AMDM algorithm shares similar geometric properties with those described in Proposition \ref{p_normupper}, in the sense that the generated samples also lie on the same local spherical manifold. Moreover, the introduction of the deviation optimization strategy leads to a more pronounced improvement in sampling quality. Relevant numerical results are provided in Appendix \ref{Sta_anal}.

\section{Statistical analysis}\label{Sta_anal}
\subsection{Gaussianity in latent space} \label{Gaussianity}
\cite{chung2022improving} have shown that $\mathcal M_t$ is concentrated on an $(n-1)$-dimensional manifold, 
which approximates an $n$-dimensional hypersphere as $t$ becomes large. Motivated by this property and proposition \ref{p_normupper}, the AMDM algorithm performs spherical aggregation in order to reduce manifold deviation. 
Next, we focus on the statistical characteristics of the distribution $p(\mathbf z_t)$, in particular its first and second-order moments, to validate the appropriateness of the hyperspherical approximation in the large $t$ regime. 

The continuous form of (\ref{forward_t}) can be rewritten as:
\begin{equation}\label{vpsde}
\mathrm d \mathbf z_t = -\frac{1}{2}g^2_t \mathbf z_t \mathrm dt + g_t \mathrm d\mathbf w_t,
\end{equation}

We can use equations (5.50) and (5.51) \citep{sarkka2019applied} to derive the relationship between the mean and variance of (\ref{vpsde}) as they evolve over time:
\begin{equation}
\frac{\mathrm d \mathbf m}{\mathrm d t}= \mathbb{E}\left[-\frac{1}{2}g_t^2 \mathbf z_t\right]=-\frac{1}{2}g_t^2\mathbf m,
\end{equation}
\begin{equation}
\begin{aligned}
\frac{\mathrm d \mathbf P}{\mathrm d t} &= \mathbb{E}\left[-\frac{1}{2}g_t^2 \mathbf z_t (\mathbf z_t - \mathbf m)^T\right] + \mathbb{E}\left[-\frac{1}{2}g_t^2 (\mathbf z_t - \mathbf m) \mathbf z_t^T \right] + g_t^2 \bm I  \\
& = -g_t^2 \mathbf P + g_t^2 \bm I.
\end{aligned}
\end{equation}

\begin{figure}[!ht]
  \centering
  \begin{subfigure}[b]{0.49\textwidth}
    \includegraphics[width=\linewidth]{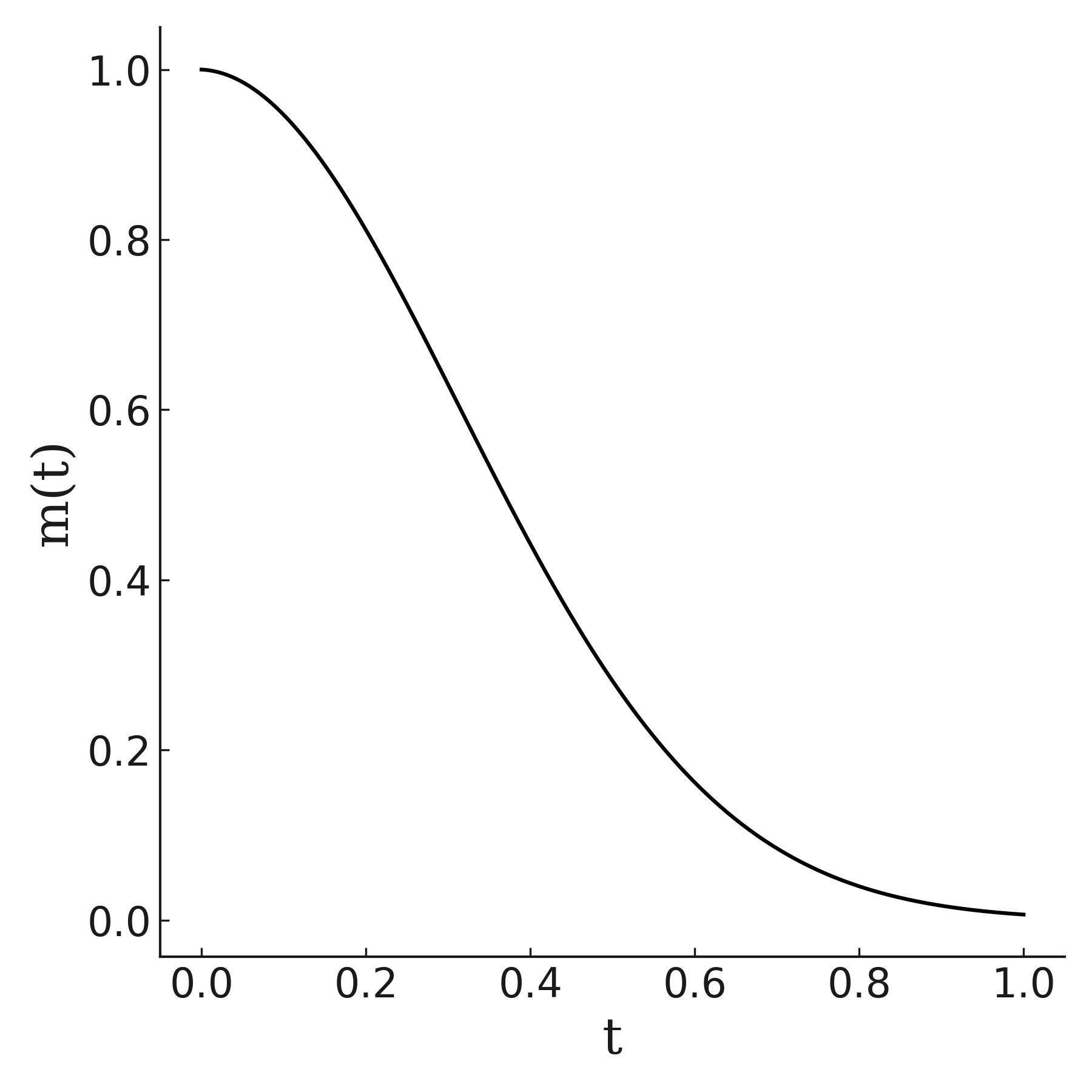}
    \caption{Mean over time.}
    \label{junzhi}
  \end{subfigure}
  \hfill
  \begin{subfigure}[b]{0.49\textwidth}
    \includegraphics[width=\linewidth]{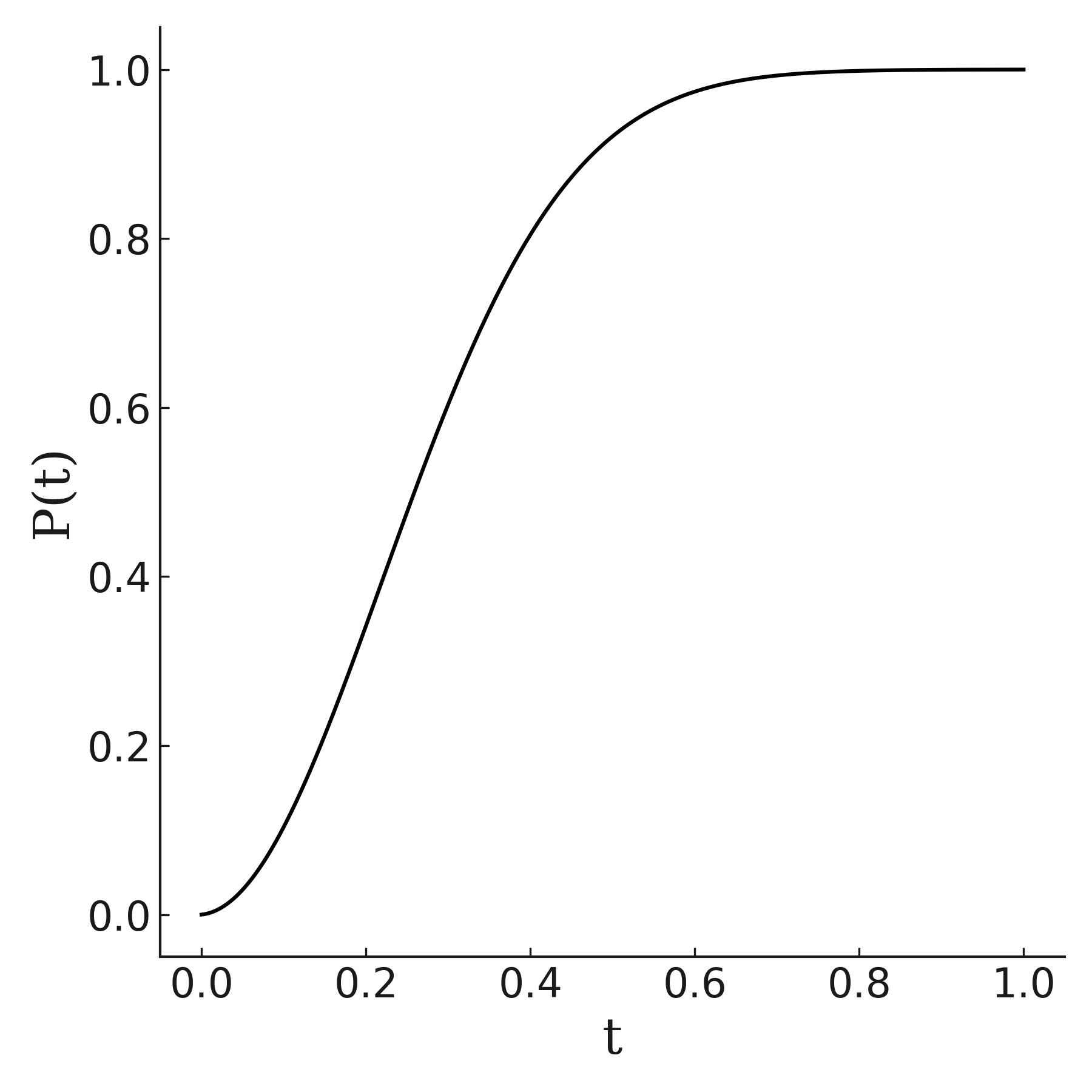}
    \caption{Variance over time.}
    \label{fangcha}
  \end{subfigure}
  \caption{Statistical properties of $p(\mathbf{z}_t)$ over time.}
  \label{gaosibianhua}
  % \vskip -0.1in
\end{figure}

The solutions are:
\begin{equation}
\mathbf m(t) = \mathbf m(0) e^{-\frac{1}{2}\int_0^t g_s^2 \mathrm d s}
\end{equation}
\begin{equation}
\mathbf P(t) = \bm I + \left(\mathbf P(0) - \bm I \right)e^{-\int_0^t g_s^2 \mathrm d s},
\end{equation}
where $g_t = \sqrt{\beta_t}$, and we set $\beta_t$ to follow the linear schedule in SD \citep{rombach2022high}. For $\mathbf m(0)$ and $\mathbf P(0)$, since the data $\mathbf{z}_0^{(i)} \in [-1,1]$, it follows that $\mathbf m(0)^{(i)} \in [-1,1]$ and $\mathbf P(0)^{(i)} \in [0,1]$. Therefore, we consider the extreme case with $\mathbf m(0) = \bm 1$ and $\mathbf P(0) = \mathbf 0$.  
The resulting mean and variance of $p(\mathbf z_t)$ as functions of time are shown in Figure \ref{gaosibianhua}.

From Figure \ref{gaosibianhua}, it can be observed that when $t > 0.6$, the mean rapidly converges to 0 while the variance rapidly converges to 1, approaching a standard Gaussian distribution. When the Gaussianity of $\mathcal M_t$ is well-behaved, implying that the global geometry exhibits desirable spherical characteristics, it is both reasonable and natural to regard two vectors with a small angular separation and equal magnitudes as lying on a local sphere.

\subsection{Numerical Analysis of the Experiments for Proposition \ref{p_normupper}}\label{assumpnorm}
\begin{figure}[!ht]
\begin{center}
\centerline{\includegraphics[width=0.5\columnwidth]{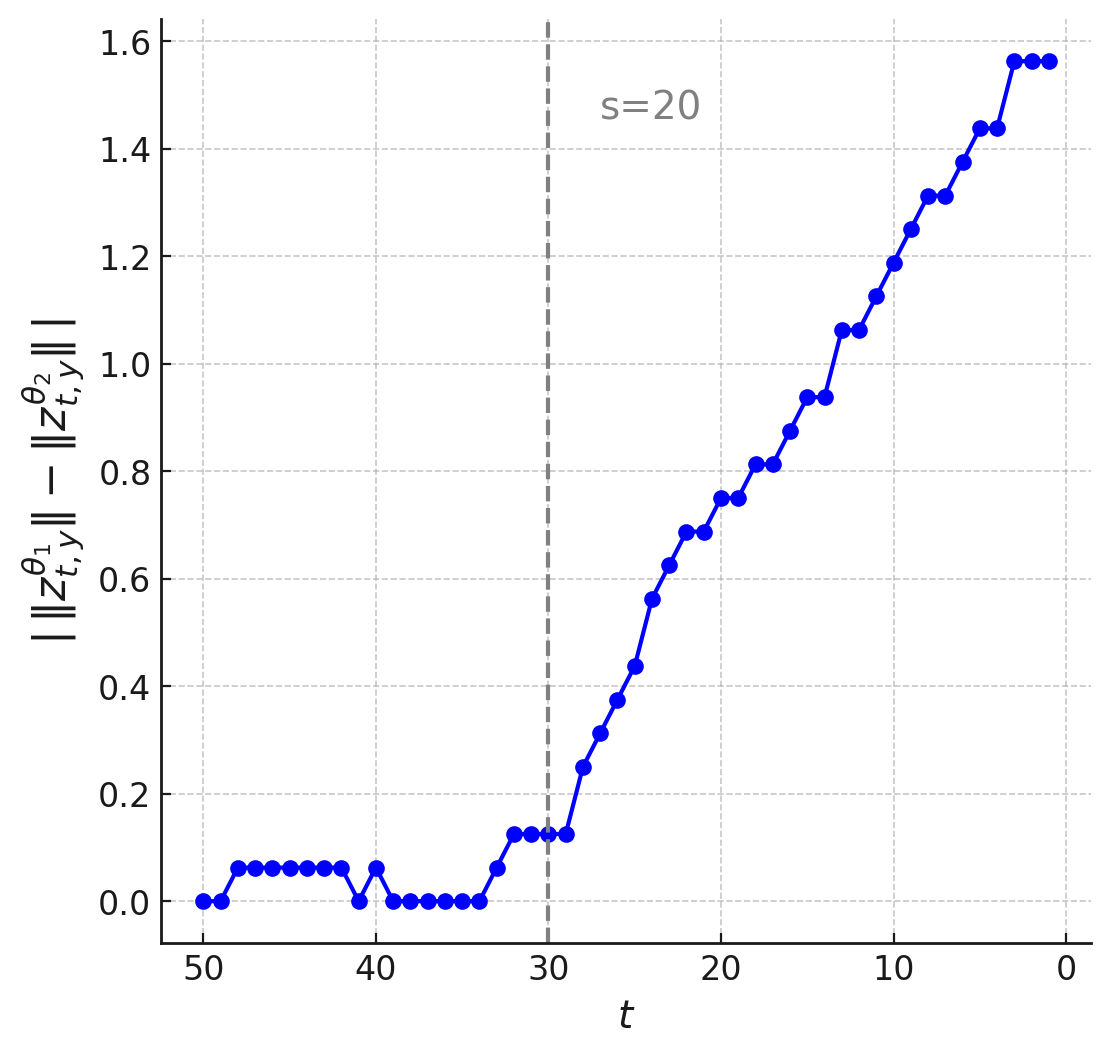}}
\caption{Difference of norm varying with time.}
\label{z_norm_diff}
\end{center}
\end{figure}

To assess the validity of Proposition \ref{p_normupper}, we log intermediate variables from InteractDiffusion(+MIGC) and report them in Table \ref{stats} and Figure \ref{z_norm_diff}. During the first \( s=20 \) aggregation steps (\( t=50 \) to \( t=30 \)), \(\left| \lVert \mathbf z_{t}^{\theta_1} \rVert - \lVert \mathbf z_{t}^{\theta_2} \rVert \right| \approx 0 \) and $\varphi$ is small, which directly supports the analysis of Proposition \ref{p_normupper}. This behavior is expected for the following reasons. At initialization (\( t=50 \)), both trajectories are sampled from the standard Gaussian and thus have numerically almost identical magnitudes; the subsequent single-step denoising from \( t=50\!\to\!49 \), although performed by different models, introduces only a small variations, so the magnitudes remain nearly equal. Moreover, the small variations in the angle \( \varphi \) and in the difference norm \( \delta=\lVert \mathbf z_{t}^{\theta_1} - \mathbf z_{t}^{\theta_2} \rVert \) further indicate that the per-step generative error is well controlled. Once aggregation takes effect, \( \mathbf z_{t-1}^{\theta_1} \) and \( \mathbf z_{t-1}^{\theta_2} \) start from the common aggregation point \( \mathbf z'_{t} \), undergo small deviation optimization, and are then sampled separately; because the aggregation point is shared, the deviations are limited, and each sampling step introduces only a small error (conditional proximity and functional proximity), the magnitudes remain almost equal. Taken together, these observations justify the equal-norm assumption throughout the aggregation process.

\begin{figure}[!ht]
  \centering
  \begin{subfigure}[b]{0.49\textwidth}
    \includegraphics[width=\linewidth]{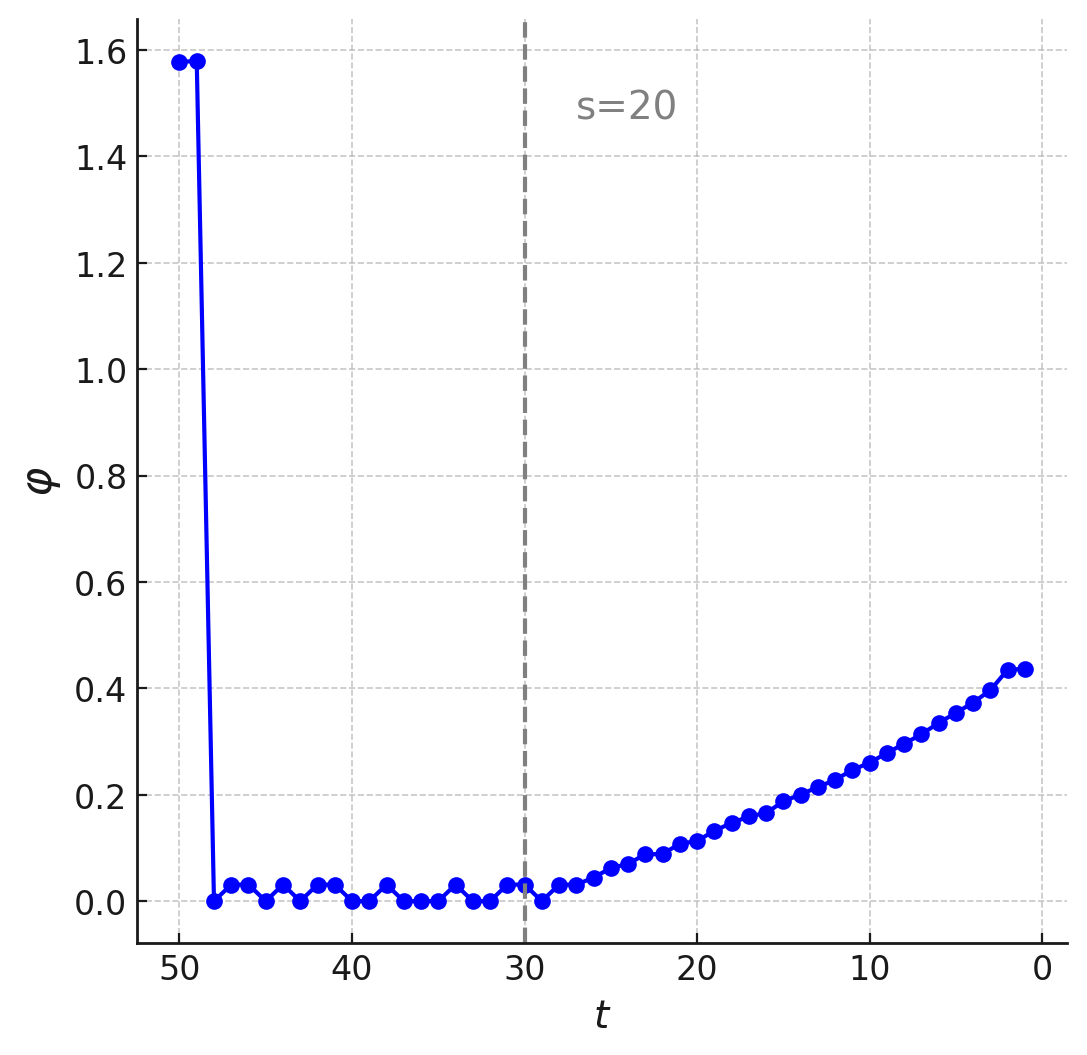}
    \caption{Angle $\varphi$ varying with time.}
    \label{varphi}
  \end{subfigure}
  \hfill
  \begin{subfigure}[b]{0.49\textwidth}
    \includegraphics[width=\linewidth]{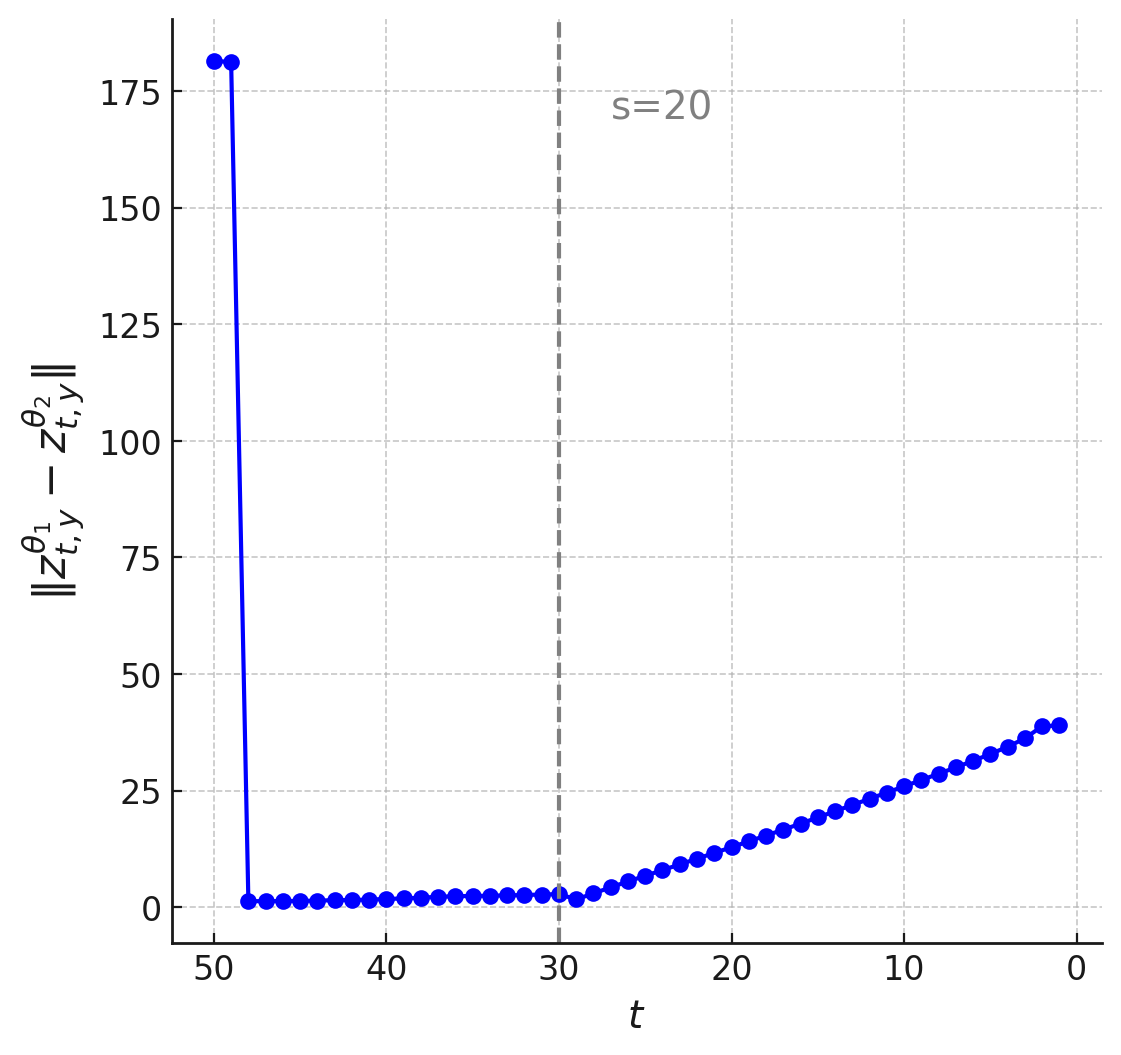}
    \caption{Difference norm varying with time.}
    \label{difference_z}
  \end{subfigure}
  \caption{Variation of angle and the difference norm.}
  % \vskip -0.5 in
\end{figure}

\begin{table}[!ht]
\centering
\caption{Values of different intermediate variables over time in InteractDiffusion (+MIGC)}
\small
\begin{tabular}{cccc}
\toprule
$t$ & $\varphi$ & $\left\Vert \mathbf z^{\theta_1}_{t}-\mathbf z^{\theta_2}_{t}\right\Vert$ & $\left|\left\Vert \mathbf z^{\theta_1}_{t}\right\Vert-\left\Vert \mathbf z^{\theta_2}_{t}\right\Vert\right|$ \\
\midrule
50 & 1.5781 & 181.3750 & 0.0000 \\
49 & 1.5791 & 181.2500 & 0.0000 \\
48 & 0.0000 & 1.3086 & 0.0625 \\
47 & 0.0312 & 1.3125 & 0.0625 \\
46 & 0.0312 & 1.3281 & 0.0625 \\
45 & 0.0000 & 1.3652 & 0.0625 \\
44 & 0.0312 & 1.4219 & 0.0625 \\
43 & 0.0000 & 1.5244 & 0.0625 \\
42 & 0.0312 & 1.5625 & 0.0625 \\
41 & 0.0312 & 1.6426 & 0.0000 \\
40 & 0.0000 & 1.7812 & 0.0625 \\
39 & 0.0000 & 1.8926 & 0.0000 \\
38 & 0.0312 & 2.0586 & 0.0000 \\
37 & 0.0000 & 2.2402 & 0.0000 \\
36 & 0.0000 & 2.3730 & 0.0000 \\
35 & 0.0000 & 2.4258 & 0.0000 \\
34 & 0.0312 & 2.4707 & 0.0000 \\
33 & 0.0000 & 2.5508 & 0.0625 \\
32 & 0.0000 & 2.6309 & 0.1250 \\
31 & 0.0312 & 2.7324 & 0.1250 \\
30 & 0.0312 & 2.8184 & 0.1250 \\
\arrayrulecolor{gray!60}\midrule
\arrayrulecolor{black} % 用完改回黑色，避免影响后续线条
29 & 0.0000 & 1.6904 & 0.1250 \\
28 & 0.0312 & 2.9727 & 0.2500 \\
27 & 0.0312 & 4.2734 & 0.3125 \\
26 & 0.0442 & 5.5273 & 0.3750 \\
25 & 0.0625 & 6.7500 & 0.4375 \\
24 & 0.0699 & 7.9648 & 0.5625 \\
23 & 0.0884 & 9.1797 & 0.6250 \\
22 & 0.0884 & 10.4141 & 0.6875 \\
21 & 0.1083 & 11.6406 & 0.6875 \\
20 & 0.1127 & 12.8750 & 0.7500 \\
19 & 0.1327 & 14.1250 & 0.7500 \\
18 & 0.1467 & 15.3750 & 0.8125 \\
17 & 0.1595 & 16.6406 & 0.8125 \\
16 & 0.1655 & 17.9219 & 0.8750 \\
15 & 0.1877 & 19.2500 & 0.9375 \\
14 & 0.2004 & 20.5938 & 0.9375 \\
13 & 0.2146 & 21.9219 & 1.0625 \\
12 & 0.2280 & 23.2500 & 1.0625 \\
11 & 0.2467 & 24.5781 & 1.1250 \\
10 & 0.2603 & 25.9219 & 1.1875 \\
9  & 0.2786 & 27.2656 & 1.2500 \\
8  & 0.2959 & 28.6094 & 1.3125 \\
7  & 0.3137 & 29.9844 & 1.3125 \\
6  & 0.3352 & 31.3906 & 1.3750 \\
5  & 0.3540 & 32.8438 & 1.4375 \\
4  & 0.3733 & 34.4062 & 1.4375 \\
3  & 0.3967 & 36.1875 & 1.5625 \\
2  & 0.4353 & 38.8125 & 1.5625 \\
1  & 0.4365 & 38.9688 & 1.5625 \\
\bottomrule
\end{tabular}
\label{stats}
\end{table}

\subsection{Maximum distance term}\label{maxdistance}
In Proposition \ref{p1}, we derived the approximate lower bound of the probability:
\begin{equation}\tag{\ref{lowbound}}
P\left(\mathbf{\tilde z}_{t-1}\in D^{\theta_2}_{t-1,y_2}\right)\geq 1 - 2 \exp\left(
   -\frac{n\left(\epsilon_{t-1}^{\theta_2} - \tfrac{d}{\sigma_t\sqrt{n}}\right)^{2}}
          {1 + 2\left(\epsilon_{t-1}^{\theta_2} - \tfrac{d}{\sigma_t\sqrt{n}}\right)}
\right),
\end{equation}
where the key quantity is the maximum distance to the model, given by $d = \phi_w(\varphi)\,\delta + \eta^{\theta_1}_{t-1}$. The numerical value of \(d\) directly determines whether the optimized variable \(\mathbf z^{\theta_1}_{t}\) is likely to lie on the data manifold of model \(\theta_2\), thereby enabling fine-grained generation. Theoretically, the smaller the value of \(d\), the higher the likelihood. To this end, we separately analyze $\phi_w(\varphi)=\frac{\sin\!\bigl((1-w)\varphi/2\bigr)}{\sin(\varphi/2)}$ and $\delta=\left\lVert \mathbf z^{\theta_1}_{t-1}-\mathbf z^{\theta_2}_{t-1}\right\rVert$.

For $\phi_w(\varphi)$, we first examine the variation of the angle $\varphi$ over time. The detailed experimental results can be referred to in the Table \ref{stats} and Fig \ref{varphi}. At the initial step $t=50$, since random points in high-dimensional Gaussian space are almost orthogonal, the angle satisfies $\varphi \approx \frac{\pi}{2}$. At $t=49$, a single sampling step of the model introduces only a negligible perturbation, so $\varphi$ remains essentially unchanged. Subsequently, following an argument similar to the analysis of difference of norm in Appendix \ref{assumpnorm}: since $\mathbf z_{t-1}^{\theta_1}$ and $\mathbf z_{t-1}^{\theta_2}$ are both derived from the common aggregation point $\mathbf z'_{t}$, and then undergo deviation optimization and independent sampling, these operations only introduce small perturbations, implying that $\varphi$ also stays small. Expanding $\phi_w(\varphi)$ in a Taylor series gives $\phi_w(\varphi) = (1-w) + O(\varphi^2)$. 
Since $\varphi$ is very small, the higher-order terms can be neglected, yielding $\phi_w(\varphi) \approx (1-w)$. In summary, at the initial step,
$\phi_w(\varphi) \approx \phi_w\!\left(\tfrac{\pi}{2}\right) = \sqrt{2}\,\sin\!\left(\tfrac{(1-w)\pi}{4}\right) \leq 1$, whereas after aggregation,
$\phi_w(\varphi) \approx (1-w) \leq 1$.
Thus, we conclude that $\phi_w(\varphi) \leq 1$.

For $\delta$, the specific experimental values are provided in the Table \ref{stats} and Figure \ref{difference_z}. Based on a similar analysis as above, it can be inferred that at the initial steps $t=50$ and $t=49$, the vectors are generated by random sampling, leading to a large magnitude of their difference. In subsequent steps, however, since both vectors originate from the same aggregation point of the previous step and the following operations introduce only small perturbations, the magnitude of their difference becomes significantly smaller.

In summary, only after the first aggregation optimization may the variable $\mathbf{\tilde z}_{T-1}$ deviate from $D^{\theta_2}_{T-1,y_2}$. In the subsequent aggregation steps, however, $\mathbf{\tilde z}_{t,y_2}$ remains, with high probability, within $D^{\theta_2}_{t,y_2}$, thereby fully incorporating the characteristics of different models and enabling fine-grained generation. This observation is also intuitive: since the initial sampling is performed from two random points that are relatively far apart, it is difficult for the first aggregation to lie simultaneously on the manifolds of both models at step $T-1$.

In addition, we conducted 10 additional tests with different random seeds. The results show that the angle $\varphi$ and the difference of norm $\left|\|\mathbf{z}_{t}^{\theta_1}\|-\|\mathbf{z}_{t}^{\theta_2}\|\right|$ remain nearly identical across experiments, while the difference norm $\left\lVert \mathbf z^{\theta_1}_{t-1}-\mathbf z^{\theta_2}_{t-1}\right\rVert$ fluctuates only within $\pm 1$. Hence, the above analysis can be regarded as robust.

\section{Comparison with Compositional Methods} \label{comp}
Compositional generation aims to synthesize complex new samples that simultaneously satisfy multiple attributes or concepts by combining information from multiple different distributions. In the context of diffusion models, this is often achieved by combining the score functions $\nabla_x \log p_t(x)$ of the noise processes from different models to simulate a new reverse stochastic differential equation (SDE).

\citet{du2023reduce} proposed an energy-based framework for compositional generation, where the scores from multiple models are summed to approximate the score of the joint distribution, that is:
\[
\nabla \log p_{\text{PoE}}(x) \approx \sum_{i=1}^N \nabla \log p_i(x),
\]
and the corresponding target distribution is the Product of Experts (PoE):
\[
p_{\text{PoE}}(x) \propto \prod_{i=1}^N p_i(x).
\]
Sampling is performed using MCMC methods such as Langevin dynamics or HMC. However, these approaches are heuristic in nature, lack theoretical guarantees, and incur high computational costs.

\citet{skreta2025feynman} introduce the \textit{Feynman-Kac Corrector (FKC)} framework, which provides a principled approach to sampling from modified target distributions:
\begin{equation*}
\begin{aligned}
&\textbf{Annealed:} \quad
p_{t, \beta}^{\text{anneal}}(x) = \frac{1}{Z_t(\beta)} q_t(x)^\beta ,\\
&\textbf{Product:} \quad
p_t^{\text{prod}}(x) = \frac{1}{Z_t} q_t^1(x) q_t^2(x) ,\\
&\textbf{Geometric Avg:} \quad
p_{t, \beta}^{\text{geo}}(x) = \frac{1}{Z_t(\beta)} q_t^1(x)^{1 - \beta} q_t^2(x)^{\beta}.
\end{aligned}
\end{equation*}
They derive weighted SDEs:
\[
dx_t = (-f_t(x_t) + \sigma_t^2 \nabla \log p_t(x_t))dt + \sigma_t dW_t,
\]
and apply Sequential Monte Carlo (SMC) to correct sample trajectories. This framework generalizes classifier-free guidance (CFG) and improves inference-time controllability.

\citet{thornton2025composition} introduce a novel framework that distills pretrained diffusion models into energy-parameterized models, where the score function is expressed as the gradient of a learned energy $s(x, t) = -\nabla_x E_\theta(x, t)$. To address training instability common in energy-based models, they propose a conservative projection loss that distills the score function from a pretrained teacher model:
\[
\min_\theta \mathbb{E}_{p_t} \left[ \|\nabla E_\theta(x, t) + s_{\text{teacher}}(x, t)\|^2 \right].
\]
This energy formulation enables the definition of new target distributions for sampling. Specifically, in compositional generation tasks, they define a composed distribution as the product of multiple submodels:
\[
p(x) \propto \prod_i \exp(-E_i(x)) = \exp\left(-\sum_i E_i(x)\right),
\]
which implies that the composed score is equivalent to the sum of individual scores:
\[
\nabla \log p(x) = -\sum_i \nabla E_i(x) = \sum_i s_i(x).
\]

This resembles the linear score composition used in prior works such as projective composition, but, unlike those methods, which plug the summed scores directly into the reverse diffusion equation.

\citet{bradley2025mechanisms} provide a theoretical foundation for \textit{projective composition}. Assuming a \textit{Factorized Conditional} structure, they show:
\[
\nabla \log p(x) = \sum_i \nabla_x \log p_i(x) - (k-1)\nabla_x \log p_b(x).
\]

where $p_b(x)$ is the background distribution (as in the unconditional model). This combination corresponds to a Bayesian fusion distribution:
\[
p(x) \propto \frac{\prod_i p_i(x)}{p_b(x)^{k-1}}.
\]

This enables direct reverse diffusion sampling without correction under appropriate assumptions. Furthermore, they generalize to feature space combinations, where orthogonal transformations \( z = A(x) \) can preserve compositionality. 

\citet{skretasuperposition} leverage the Itô density estimator to dynamically weight the score functions of different models, yielding a new generative vector field. Given $M$ diffusion models with score functions $\nabla \log q^i_t(x)$, SUPERDIFF defines:
\[u_\tau(x) = \sum_{i=1}^M \kappa^i_\tau(x) \, \nabla \log q^i_t(x).\]
In the OR mode (mixture of densities), the weights are defined by a softmax over the estimated log-densities:
\[\kappa^i_\tau(x) = \frac{\exp\big(T \cdot \log q^i_t(x) + \ell\big)}{\sum_j \exp\big(T \cdot \log q^j_t(x) + \ell\big)},\]
allowing the combined model to generate samples consistent with any of the component models. In the AND mode (intersection of densities), the weights are obtained by solving a linear system such that all log-densities evolve consistently: 
\[\sum_{i=1}^M \kappa^i_\tau(x) \, \frac{d}{d\tau}\log q^i_t(x) \approx \text{constant across } i,\]
enforcing agreement among models and producing samples that satisfy all constraints simultaneously. Finally, the sampling process follows the reverse-time SDE: 
\[dx_\tau = \big(-f_{1-\tau}(x_\tau) + g^2_{1-\tau} u_\tau(x_\tau)\big) d\tau + g_{1-\tau} dW_\tau,\]
which yields samples representing either the union (OR) or the intersection (AND) of the underlying model distributions.

% \begin{table}[ht]
% \caption{Comparison between AMDM and Composition Methods}
% \vspace{0.1 in}
% \centering
% \resizebox{\textwidth}{!}{
% \begin{tabular}{l c c}
% \toprule
% \textbf{Description} & \textbf{AMDM} & \textbf{Composition Methods} \\
% \midrule
% Symbol & $p^{\theta_1}(x\mid y)$, $p^{\theta_2}(x\mid y)$ or w/o $y$ & $p^{\theta_1}(x\mid y_1)$, $p^{\theta_2}(x\mid y_2)$ or w/o $y_1, y_2$ \\

% Distribution & Both standard Gaussian distribution & Different data distributions \\

% Condition Input & Same text & Different texts \\

% Output & Within the domain, still a standard Gaussian distribution & 
% Outside the domain, a new distribution such as $p_1 + p_2$ or $p_1 p_2$ \\

% Task & Fine-grained generation & Different distribution composition \\
% \bottomrule
% \end{tabular}
% }
% \vspace{0.1 in}
% \label{compositiondiff}
% \end{table}

To further compare the differences in task adaptation, we also provide compositional methods with the same fine-grained input and conduct compositional experiments on both MIGC and InteractDiffusion. The results are compared with the AMDM method and presented in Table \ref{compcoco-mig}.

\begin{table}[ht]
\centering
\caption{Composition of MIGC and InteractDiffusion on the COCO-MIG Benchmark and HOI Detection Score}
\vspace{0.1 in}
\label{compcoco-mig}
\resizebox{\textwidth}{!}{
\begin{tabular}{lcccc}
\toprule
\textbf{Method} & \textbf{Instance Success Rate Avg} $\uparrow$ & \textbf{mIoU Score Avg} $\uparrow$ & \textbf{Default (Full)} $\uparrow$ & \textbf{Time (s)} $\downarrow$ \\
\midrule
\citet{du2023reduce}         & 45.14 & 40.03 & 19.23 & 42.6 \\
\citet{bradley2025mechanisms} & 47.82 & 42.71 & 22.23 & 12.9 \\
\citet{skreta2025feynman}   & 46.35 & 41.25 & 21.74 & 13.6\\
\citet{skretasuperposition}   & 46.56 & 41.81 & 23.15 & 15.5\\
\midrule
InteractDiffusion(+MIGC) & \textbf{54.78} & \textbf{47.74} & \underline{26.04} & \textbf{8.8} (single model 7.5s) \\
MIGC(+InteractDiffusion) & \underline{49.78 } & \underline{43.69} & \textbf{31.40} & \underline{9.4} \\
\bottomrule
\end{tabular}
}
\end{table}

From equation (\ref{cfg}), we observe that the linear combination of different $\mathbf z_t$ from various models corresponds to the linear combination of their predicted noise, which is proportional to the score. Therefore, the linear combination in AMDM can be interpreted as a linear combination of scores, which is actually similar to the compositional methods. As shown in our aggregation ablation study of AMDM in Figure \ref{ablation2}, linear combination performs worse than spherical aggregation. This is because, in fine-grained generation, the data lies on a spherical manifold, and spherical aggregation better preserves generation quality. Although Du's method \citep{du2023reduce} applies Langevin dynamics to improve sample quality after linear aggregation and achieves good results, it comes with substantial computational overhead. Similarly, SUPERDIFF \citep{skretasuperposition} reduces manifold deviation from a methodological perspective and achieves favorable performance. However, in fine-grained generation scenarios, its "AND" mode requires solving a system of linear equations, which introduces additional overhead. Moreover, Bradley's work \citep{bradley2025mechanisms} theoretically proves that linear score combination only yields high-quality results when the compositional conditions lie in orthogonal feature spaces. However, fine-grained generation typically involves nearly identical conditions, which clearly violates this assumption, providing theoretical support for the observed experimental results.

Finally, we conduct a simple compositional generation experiment (2D Composition Mixture) to illustrate the limitations of AMDM in compositional generation. The experimental results are shown in Table \ref{non-gaussian}. The reason for the failure of the AMDM algorithm is also quite simple: AMDM is only applicable under the approximate high-dimensional Gaussian distribution; for general distributions, spherical aggregation and deviation optimization fail. It is clear that AMDM differs significantly from other combination methods in terms of application scenarios.

\begin{table}[ht]
\centering
\caption{Comparison of AMDM and Du 2023 under a non-Gaussian distribution}
\vspace{0.1in}
\label{non-gaussian}
\begin{tabular}{lccc}
\toprule
\textbf{Method} & \textbf{ln(MMD)} $\downarrow$ & \textbf{LL} $\uparrow$ & \textbf{Var} $\downarrow$ \\
\midrule
AMDM    & -3.51 & -2.43 & 0.032 \\
\citet{du2023reduce} & \textbf{-4.48} & \textbf{1.30} & \textbf{0.007} \\
\bottomrule
\end{tabular}
\end{table}

\section{Experimental Details}\label{ed}
\subsection{Evaluation Metrics}
\textbf{COCO-MIG Benchmark} \citep{zhou2024migc} assesses the enhancement in attribute metrics. The COCO-MIG Benchmark is based on the COCO-position layout, where each instance is assigned a color attribute, requiring the generated instances to satisfy both position and color constraints. The process includes sampling layouts from COCO and categorizing layouts into five levels (L2-L6) based on the number of instances. Then, colors are assigned to each instance, global prompts are constructed, and a test file containing 800 entries is generated. The COCO-MIG metrics primarily include Instance Success Rate and mIoU Score. The Instance Success Rate measures the probability of each instance being generated correctly, while mIoU Score calculates the average of the maximum IoU for all instances; if the color attribute is incorrect, the IoU value is set to 0.
Since the MIG-Benchmark does not contain interactive prompts, we set the "action" input in the InteractDiffusion model to "and".

\textbf{FGAHOI} \citep{ma2023fgahoi} can measure interaction controllability. We use the FGAHOI model based on the Swin-Tiny architecture, as an HOI detector to evaluate the model's ability to control interactions. HOI Detection Scores are categorized into two types: Default and Known Object. The Default setting is more challenging, as it requires distinguishing between irrelevant images.

\textbf{CLIP} \citep{hessel2021clipscore} is the CLIPScore of the generated images with captions of the image prompts.

\subsection{Parameter Settings}

\textbf{InteractDiffusion and MIGC.} The total sampling steps \( T \) are set to 50, the aggregation step \( s \) is set to 20, the weighting factor \( w \) is set to 0.5 and the optimization steps \( \eta^{\theta_1}_t \) and \( \eta^{\theta_2}_t \) are both set to 0.3. We use InteractDiffusion v1.0, and MIGC is modified to use the DDIM sampling method to align with the same diffusion process.

\textbf{InteractDiffusion and IP-Adapter.} The total sampling steps \( T \) are set to 10, the aggregation step \( s \) is set to 5, the weighting factor \( w \) is set to 0.5, ip scale is set to 0.8 and the optimization steps \( \eta^{\theta_1}_t \) and \( \eta^{\theta_3}_t \) are both set to 0.3. We utilized IP-Adapter based on SD1.5 while keeping InteractDiffusion unchanged.

\textbf{InteractDiffusion, MIGC and IP-Adapter.} The pretrained models for the three architectures remain consistent with those mentioned above. The total sampling step \(T\) is set to 10, an aggregation step \(s\) set to 5, and weight factors \(w_1\) and \(w_2\) set to 0.5 and 0.65. The optimization steps \(\eta^{\theta_1}_t\), \(\eta^{\theta_2}_t\), and \(\eta^{\theta_3}_t\) are all simply set to 0.3, respectively.

\subsection{Additional Experiments}
In order to further evaluate the style features, we additionally provide CLIP scores also on COCO validation set for reference. Table \ref{tab:clipt} shows a significant improvement in the aggregated style control from 0.533 to 0.581, nearly reaching the performance of IP-Adapter, also demonstrating the effectiveness of AMDM. 

\begin{table}[ht]
\centering
\caption{CLIP scores for InteractDiffusion(+IP-Adapter).}
\begin{tabular}{l c}
\toprule
\textbf{Method} & \textbf{CLIP $\uparrow$} \\
\midrule
InteractDiffusion & 0.533 \\
InteractDiffusion(+IP-Adapter) & 0.581 \\
\textcolor{gray}{IP-Adapter} & \textcolor{gray}{0.588} \\
\bottomrule
\end{tabular}
\label{tab:clipt}
\end{table}

In the main experiments, we have used different diffusion versions to validate the generality, which are also within the same diffusion model ecosystem: the IP-Adapter model is based on SD1.5, while InteractDiffusion and MIGC are based on SD1.4. Additionally, we conducted experiments using SDXL-based InteractDiffusion and IP-Adapter. SDXL and SD 1.4/1.5 use different VAE encoders, meaning that they are in the different diffusion model ecosystem. Building on this, we additionally conduct experiments on models from the SDXL ecosystem to further validate the generality of the AMDM algorithm. We also used CLIP to evaluate style features. As shown in Table \ref{tab:cliptxl}, the style control significantly improves from 0.545 to 0.597, nearly reaching the performance of IP-Adapter. Because AMDM operates exclusively on the intermediate latent variables \(\mathbf z_t\) of the diffusion process rather than on the parameterization of noise prediction, it is intrinsically decoupled from both \(\epsilon\)-prediction and \(v\)-prediction. Consequently, as long as the latent \(\mathbf z_t\) at the same time step \(t\) is accessible, the method can be seamlessly paired with different prediction parameterizations and samplers without modifying model weights or training objectives, demonstrating broad applicability and portability.

\begin{table}[ht]
\centering
\caption{CLIP scores for InteractDiffusion(+IP-Adapter) based on SDXL.}
\begin{tabular}{l c}
\toprule
\textbf{Method} & \textbf{CLIP $\uparrow$} \\
\midrule
InteractDiffusion & 0.545 \\
InteractDiffusion(+IP-Adapter) & 0.597 \\
\textcolor{gray}{IP-Adapter} & \textcolor{gray}{0.613} \\
\bottomrule
\end{tabular}
\label{tab:cliptxl}
\end{table}

Finally, we conduct experiments using InteractDiffusion based on SD1.4 and MIGC based on EDM sampling. Compared with DDPM, EDM adopts a different noise schedule, and therefore the two models 
belong to different diffusion model ecosystems. We use this experiment to validate the analysis in Section \ref{analysis}, 
which states that aggregation occurs within the same diffusion model ecosystem. As shown in Table \ref{interactmigctableedm}, when the two models are not derived from the same SDE, the AMDM algorithm fails: the generated images exhibit poor quality and deviate from the data manifold, since the two data domains do not lie on the same $\mathcal{M}_t$, making aggregation inapplicable. 
This is because, as analyzed in Section \ref{analysis}, aggregation operations can only occur between models within the same diffusion model ecosystem. This constraint arises from mathematical properties and also constitutes a fundamental principle that other compositional methods must adhere to. In contrast, works such as \citep{biggs2024diffusion, ohdawin, wangensembling} impose an additional restriction, requiring identical denoising networks. From this perspective, AMDM does not impose specific constraints on the design of the denoising network. Given that many recent works build upon the same pretrained models for secondary development, these requirements are relatively easy to satisfy in practice, which endows AMDM with broad applicability and extends its potential to a wider range of application scenarios.

\begin{table}[!ht]
    \caption{Quantitative results in different diffusion model ecosystems on the COCO-MIG benchmark and CLIP Score in InteractDiffusion(+MIGC).}
    \centering
    \resizebox{\textwidth}{!}{
    \begin{tabular}{c|c c c c c c |c c c c c c | c c}
        \toprule
        \multirow{2}{*}{\textbf{Method}} & \multicolumn{6}{c|}{\textbf{Instance Success Rate (\%) $\uparrow$ }} & \multicolumn{6}{c|}{\textbf{mIoU Score (\%) $\uparrow$}} & \multicolumn{2}{c}{\textbf{CLIP Score $\uparrow$}} \\ 
         \cmidrule(lr){2-7} \cmidrule(lr){8-13} \cmidrule(lr){14-15}
         & $L_2$ & $L_3$ & $L_4$ & $L_5$ & $L_6$ & Avg & $L_2$ & $L_3$ & $L_4$ & $L_5$ & $L_6$ & Avg & Global & Local \\
        \midrule
        InteractDiffusion  & 37.50 & 35.62 & 35.31 & 30.62 & 34.16 & \cellcolor[gray]{0.9}34.06 & 32.98 & 31.63 & 30.82 & 28.29 & 30.40 & \cellcolor[gray]{0.9}30.40 & 31.09 & 27.56 \\
        InteractDiffusion(+MIGC) & 20.03 & 18.67 & 18.45 & 15.62 & 16.29 & \cellcolor[gray]{0.9}18.13 &17.64 & 16.38 & 15.60 & 13.25 &14.98 & \cellcolor[gray]{0.9}15.03 & 14.77 & 13.91\\
        \midrule
        \midrule
        \textcolor{gray}{MIGC} &
        \textcolor{gray}{67.70} & \textcolor{gray}{59.61} & \textcolor{gray}{58.09} & \textcolor{gray}{56.16} & \textcolor{gray}{56.88} & \textcolor{gray}{58.43} & \textcolor{gray}{59.39} & \textcolor{gray}{52.73} & \textcolor{gray}{51.45} & \textcolor{gray}{49.52} & \textcolor{gray}{49.89} & \textcolor{gray}{51.48} &\textcolor{gray}{33.01}&\textcolor{gray}{28.95}\\
        \bottomrule
    \end{tabular}
    }
    \label{interactmigctableedm}
\end{table}

\section{Visual Results of Table \ref{ab_aggregation_stages}}\label{visualab}

To better demonstrate that diffusion models initially focus on generating coarse-grained features such as position, attributes, and style, while later stages emphasize quality and consistency, we further visualize  the experiment where $s=5$.

\begin{figure}[!ht]
    \centering
    \includegraphics[width=1\textwidth]{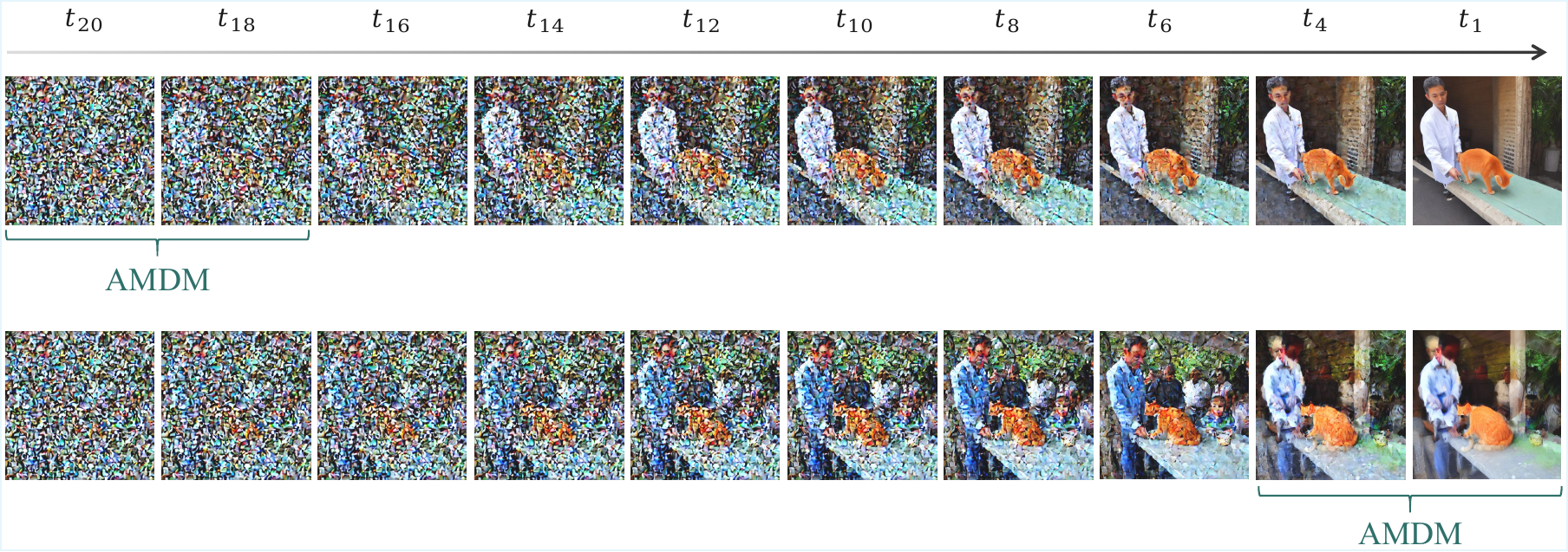}
    \caption{Different stages of the AMDM when aggregation steps are \(s = 5\).}
    \label{vab2}
\end{figure}

From Figure \ref{vab2}, it becomes even clearer that when aggregation occurs in the later stages, especially once the diffusion model begins focusing on detail generation, applying the aggregation algorithm at that point induces manifold deviation, causing a sharp deterioration in generation quality.

This further supports our conclusion that diffusion models initially focus on generating coarse-grained features such as position, attributes, and style, while later stages emphasize quality and consistency.
\section{Limitations and Future Work}\label{limi}
We propose a novel aggregation algorithm AMDM. This algorithm is specifically designed for fine-grained generation tasks within the same diffusion model ecosystem.
The theoretical assumptions in this work (functional proximity, conditional proximity, and local Lipschitz continuity) are primarily tailored to fine-grained tasks, and this is empirically supported by our observations. These assumptions form the key distinction of AMDM from existing compositional diffusion approaches. Although it do not provide global guarantees, and we leave the development of more relaxed theoretical foundations to future work.
In addition, the selection of the hyperparameter $\eta$ in the AMDM algorithm is largely heuristic. In future work, we will investigate dynamic adaptive selection strategies that are grounded in data-driven criteria or theoretical analysis to enhance the method’s robustness and generalization.

\section{Broad Impacts}
We advocate against injecting harmful features into high-quality models to prevent the generation of socially harmful images.

\section{Additional Visual Results}
\begin{figure}[ht]
    \centering
    \includegraphics[width=1\textwidth]{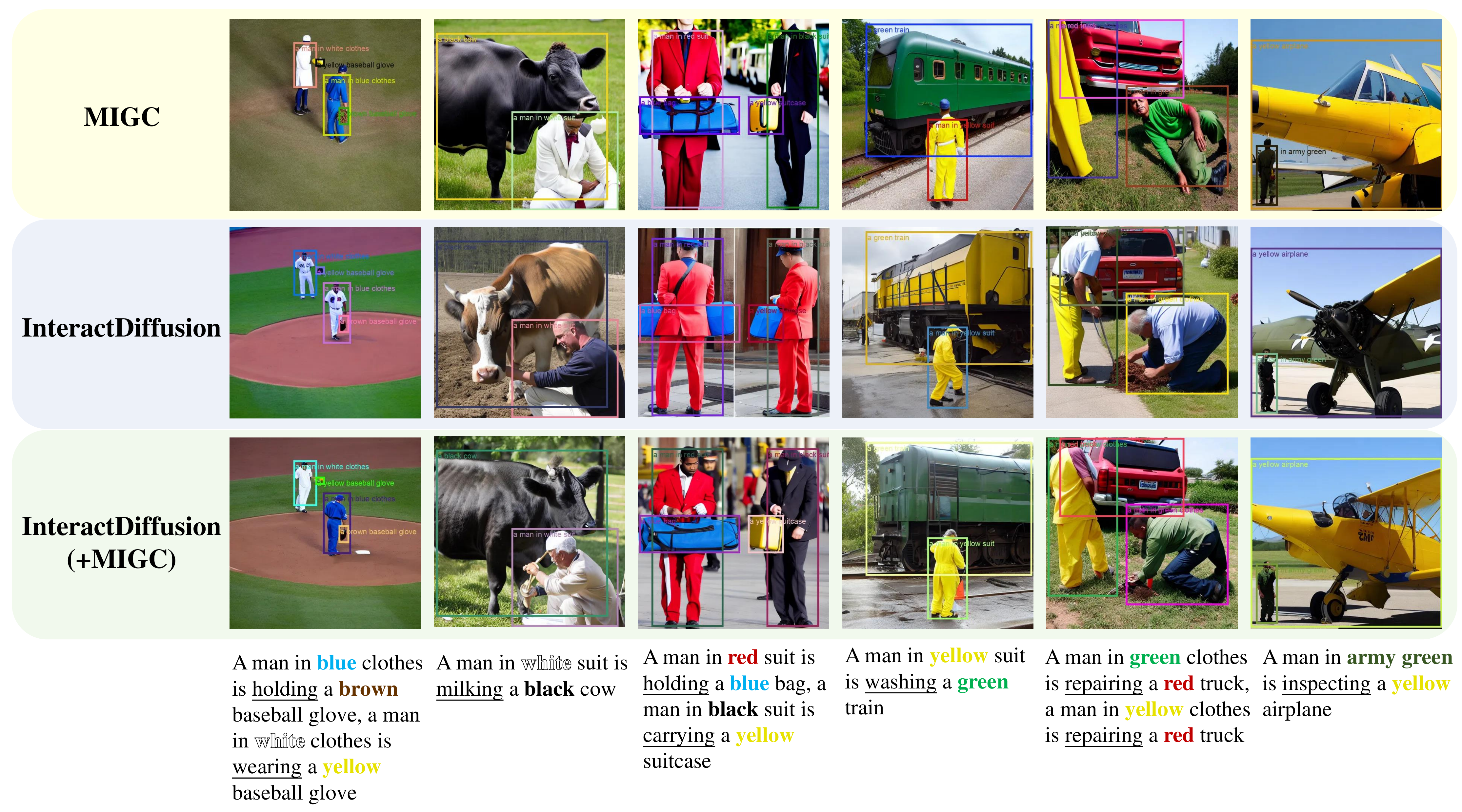} % 调整图片宽度
    \caption{Additional visual results of aggregating MIGC into InteractDiffusion applying the AMDM algorithm.}
    \label{ap_interactmigc}
\end{figure}

\begin{figure}[ht]
    \centering
    \includegraphics[width=1\textwidth]{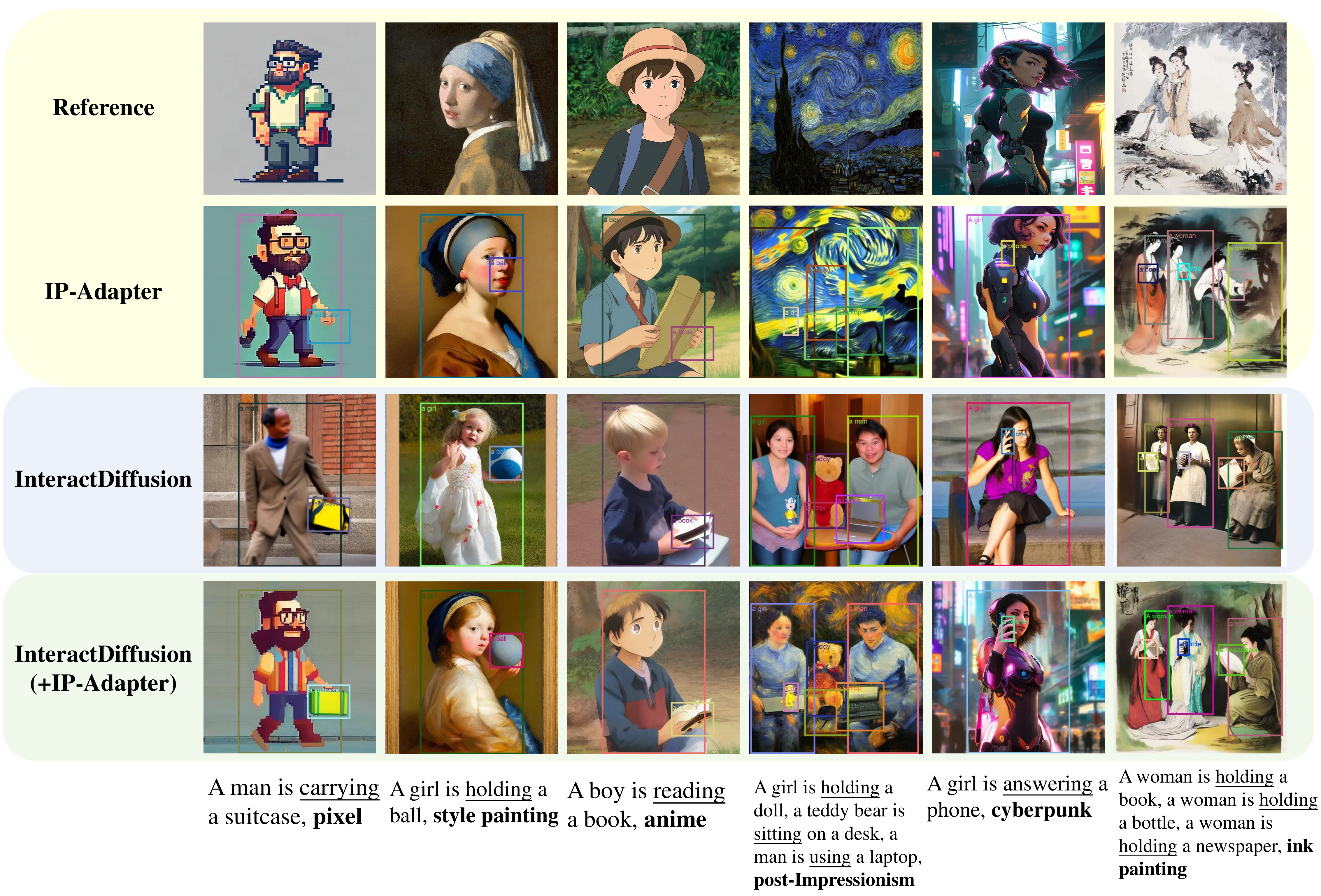} % 调整图片宽度
    \caption{Additional visual results of aggregating IP-Adapter into InteractDiffusion applying AMDM algorithm.}
    \label{ap_interactip}
\end{figure}

\begin{figure}[ht]
    \centering
    \includegraphics[width=1\textwidth]{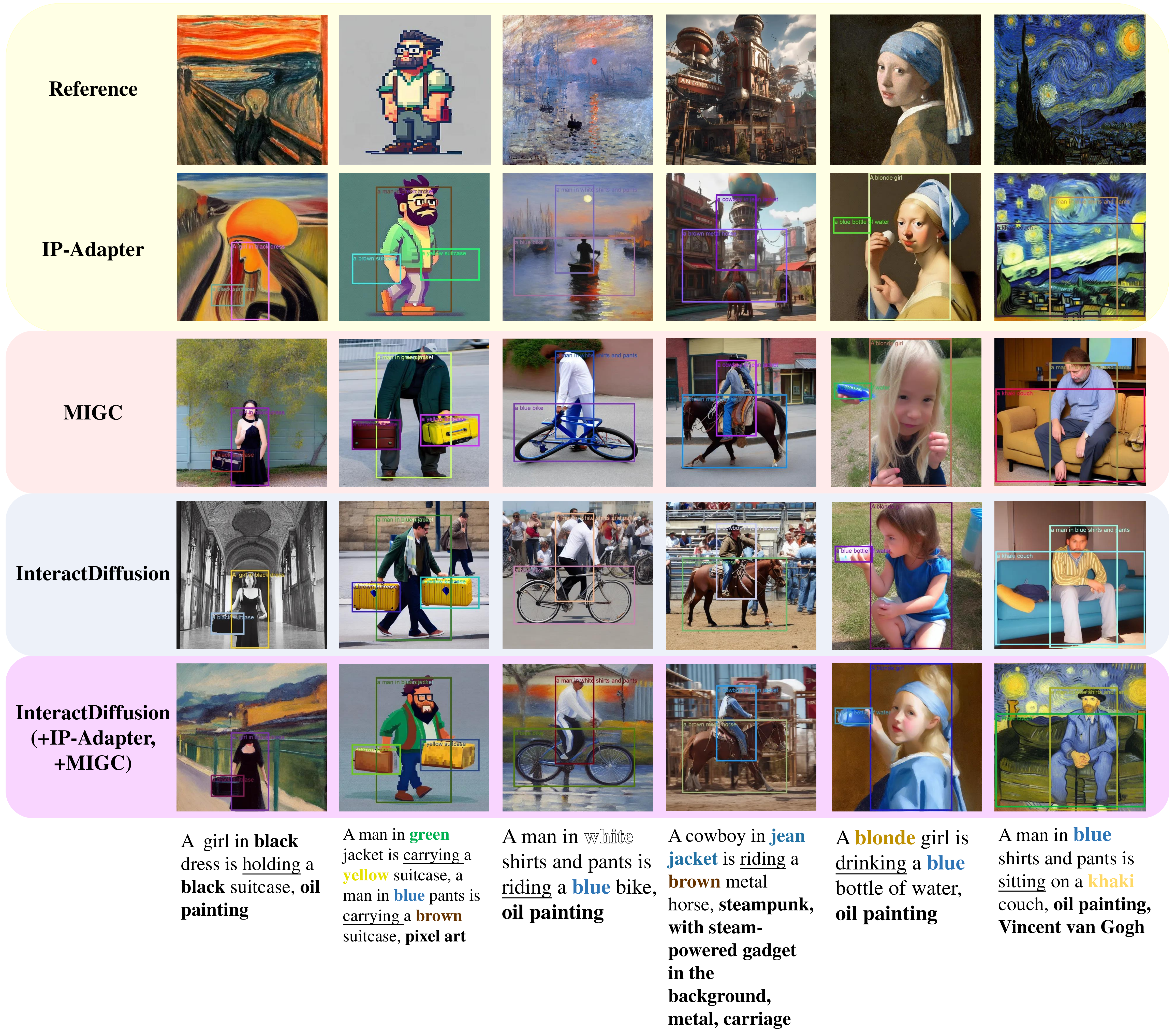} % 调整图片宽度
    \caption{Additional visual results of aggregating MIGC and IP-Adapter into InteractDiffusion applying the AMDM algorithm.}
    \label{ap_interactmigcip}
\end{figure}
\clearpage
\section{Use of LLMs}
We have carefully reviewed the paper for spelling, grammar, punctuation issues using LLMs.

\end{document}